\documentclass{article}

\PassOptionsToPackage{numbers,compress}{natbib}
\usepackage[eandd, preprint]{neurips_2026}

\usepackage[utf8]{inputenc} 
\DeclareUnicodeCharacter{2261}{\ensuremath{\equiv}}
\usepackage[T1]{fontenc}    
\usepackage{hyperref}       
\usepackage{url}            
\usepackage{booktabs}       
\usepackage{amsfonts}       
\usepackage{nicefrac}       
\usepackage{microtype}      
\usepackage{xcolor}         

\usepackage{amsmath}
\usepackage{amssymb}
\usepackage{amsthm}
\usepackage{mathtools}
\usepackage{algorithm}
\usepackage{multirow}
\usepackage{makecell}
\usepackage{array}
\usepackage{rotating}
\usepackage{hyperref}

\usepackage{graphicx}
\usepackage{subcaption}

\usepackage{enumitem}

\usepackage{hyperref}
\usepackage[capitalize,noabbrev]{cleveref}

\usepackage{float}              
\usepackage{placeins} 
\usepackage{tikz}
\usetikzlibrary{arrows.meta, positioning, fit, backgrounds, shapes.geometric}

\usepackage{algpseudocode}
\usepackage[breakable]{tcolorbox}
\tcbset{
  promptstyle/.style={
    breakable,
    colback=gray!5,
    colframe=gray!50!black,
    boxsep=3pt, top=5pt, bottom=5pt, left=6pt, right=6pt,
    fontupper=\footnotesize\ttfamily,
    fonttitle=\bfseries\footnotesize,
    arc=2pt,
  }
}

\title{MathConstraint: Automated Generation of Verified Combinatorial Reasoning Instances for LLMs
}

\author{%
  Viresh Pati$^{1\dagger}$, Zhengyu Li$^1$, Piyush Jha$^1$, Rahul Garg$^1$, Yatharth Sejpal $^2$, Vijay Ganesh$^{1\dagger}$ \\
  $^1$Georgia Institute of Technology, USA \quad $^2$KNOWIDEA Technologies\\
  $^\dagger$\texttt{\{vireshpati, vganesh\}@gatech.edu}
}

\begin{document}

\maketitle
\begin{abstract}

We introduce \textsc{MathConstraint}, a hard, adaptive benchmark for evaluating the combinatorial reasoning capabilities of LLMs. We combine constraint satisfaction problems with rigorous solver-based verification and design an adaptive generator to create instances that remain challenging as the LLMs improve in their reasoning capabilities. Unlike existing benchmarks that quickly saturate on fixed datasets or use LLM-as-a-judge for checking solutions, \textsc{MathConstraint} uses parameterized problem types that enable scalable generation of arbitrarily difficult and automatically verifiable instances. We release \textsc{MathConstraint-Easy} ($266$ instances), on which frontier models achieve between $72.6\%$ (\textsc{gemini-3.1-flash-lite}) and $87.6\%$ (\textsc{gpt-5.5}) accuracy, and \textsc{MathConstraint} ($329$ instances) on which the same models drop to between $18.5\%$ (\textsc{claude-4.6-sonnet}) and $66.9\%$ (\textsc{gpt-5.5}) accuracy, demonstrating the resilience of our benchmark generator against rapid progress in LLM reasoning capabilities. We evaluate 12 frontier and open-weight models with and without access to a sandboxed Python environment that includes generic SAT/SMT solvers. Tool access roughly doubles frontier accuracy on \textsc{MathConstraint} (mean $+28$pp; up to $+52$pp for \textsc{claude-4.6-sonnet}). Further, halving the tool-call budget from $8$ to $4$ rounds erases up to $37$ points --- a sensitivity that most single-budget benchmarks miss. We release the generator, dataset, and evaluation harness as a robust environment for studying combinatorial reasoning and tool-use behavior under adversarially-tunable difficulty.

\end{abstract}

\section{Introduction}
\label{sec:intro}


Large language models are improving quickly on mathematical and algorithmic tasks, but evaluating that progress reliably is becoming harder. Fixed benchmarks saturate, public leaderboards age quickly, and contamination becomes increasingly difficult to rule out as models ingest larger fractions of the open web and common benchmark corpora. These problems are especially acute for reasoning tasks, where a benchmark can stop being useful long before it stops being cited. What is needed is not merely another static test set, but a benchmark generator: evaluation infrastructure that can keep producing fresh, hard, automatically gradable instances as model capabilities change.

Combinatorial reasoning is a particularly attractive setting for such infrastructure. Many combinatorial problems admit a natural difficulty dial through instance size or structural parameters; the answer space is too large to memorize; and correctness is brittle, since a single violated constraint invalidates an otherwise plausible solution. At the same time, these tasks are well matched to exact solvers, which makes it possible to grade outputs against solver-verified ground truth rather than relying on heuristics or LLM-as-a-judge. This combination of scalable hardness and rigorous verification makes combinatorial reasoning a natural stress test for LLMs.

Recent NP-hard and SAT-oriented evaluations demonstrate that combinatorial reasoning remains challenging for current models, but many are fixed collections of instances or narrow slices of problem families~\citep{fan2024nphardeval, yang2025nondeterministic, wei2025satbench, lin2025zebralogic}. Verified mathematical benchmarks provide strong grading guarantees, but often emphasize proof or olympiad-style mathematics rather than broad, parameterized combinatorial search and reasoning~\citep{zheng2021minif2f, tsoukalas2024putnambench, balunovic2025mathconstruct}. Finally, adaptive benchmarks mitigate saturation and contamination, yet typically do not pair continuous difficulty control with solver-backed verification over a broad pool of combinatorial families~\citep{white2024livebench, balunovic2025matharena, xu2026mathduels}. 

We introduce \textsc{MathConstraint} to combine these ingredients in a single evaluation framework. \textsc{MathConstraint} is an adaptive benchmark generator for combinatorial reasoning. Starting from parameterized problem families drawn from constraint programming and graph construction domains, it automatically writes natural-language problem instances, uses a solver to determine the correct answer, and, when a concrete solution exists, stores a solver-verified example solution. Because difficulty is controlled through generation rather than through a fixed test set, the same framework can be rerun at cranked parameters to produce harder benchmarks without changing the prompt format or the verification contract. This makes \textsc{MathConstraint} better viewed as evaluation infrastructure than as a one-off benchmark release.

This setup also lets us study tool use in a way that is tightly coupled to reasoning quality. We evaluate models both without tools and with access to a sandboxed Python environment that includes generic SAT/SMT solving support. In this regime, performance depends not only on whether a model can reason about a combinatorial instance, but also on whether it can encode the problem correctly, decide when to delegate to a solver, and manage a limited sequence of tool calls. Our experiments show that solver access substantially changes the leaderboard, and that measured accuracy is highly sensitive to the available tool budget. Together, these results suggest that budgeted tool use is not an implementation detail of evaluation, but a representative capability in its own right. We introduce \textsc{sim@}$k$, a replay metric that scores logged traces under smaller tool budgets, and show that measured accuracy is highly sensitive to that budget. Strong performance requires not just abstract reasoning, but effective orchestration of external computation under constraints.

Our contributions are:
\begin{enumerate}[leftmargin=*,itemsep=0.3em]
    \item We introduce \textsc{MathConstraint}, an adaptive benchmark framework that generates parameterized constraint-programming and graph/SAT-style problem types, with solver-certified satisfiability labels and verifier-checked candidate witnesses.
    \item We instantiate this framework as two evaluation datasets, \textsc{MathConstraint-Easy} ($266$ instances, 25 types) and \textsc{MathConstraint} ($329$ instances, 39 types), and demonstrate adaptive resistance to saturation. Frontier no-tools accuracy falls from $72.6$--$87.6\%$ on \textsc{MathConstraint-Easy} to $18.5$--$66.9\%$ on \textsc{MathConstraint}.
     \item We evaluate $12$ frontier and open-weight models across no-tools and tool-enabled settings. On \textsc{MathConstraint}, only $3/12$ models exceed $50\%$ accuracy without tools, while tool access raises mean accuracy by $28$ percentage points and improves \textsc{claude-4.6-sonnet} by $52$ points.
    \item We introduce \textsc{sim@}$k$, a tool-budget replay metric, and highlight tool-use efficiency as a first-class dimension of combinatorial reasoning abilities: halving the budget from $8$ to $4$ rounds erases up to $37$\,pp of accuracy.
    \item We release an open source library for evaluating LLMs with agentic tool-use capabilities on combinatorial reasoning tasks. Our evaluation code is available at this \href{https://github.com/vireshpati/Math-Constraint}{link}. Our datasets, \href{https://huggingface.co/datasets/MathConstraint/MathConstraint}{\textsc{MathConstraint}} and \href{https://huggingface.co/datasets/MathConstraint/MathConstraint-Easy}{\textsc{MathConstraint-Easy}} are available on HuggingFace.
\end{enumerate}

\section{Related Work}
In this section, we discuss important prior approaches to evaluating the combinatorial and mathematical reasoning abilities of LLMs.

\paragraph{Combinatorial and NP-hard reasoning benchmarks.}
Combinatorial search is a natural stress test for LLMs. Hardness rises continuously with problem size, the answer space is too large to memorize, and a single missed constraint makes a candidate wrong. A recent line of work grounds this evaluation in computational complexity, anchored by \textsc{NPHardEval}~\citep{fan2024nphardeval} and \textsc{NPPC}~\citep{yang2025nondeterministic}; \textsc{NP-Engine}~\citep{li2025np} pushes the same idea toward optimization reasoning. A second line stays close to constraint-programming and SAT. \textsc{ZebraLogic}~\citep{lin2025zebralogic} indexes hardness by Z3 conflict count and reports a sharp accuracy collapse with scale, a pattern echoed by \textsc{SATBench}~\citep{wei2025satbench}, by the natural-language combinatorial-optimization suite of \citet{jiang2026reasoning}, and by \textsc{Enigmata}~\citep{chen2025enigmata}. Graph and logic-puzzle variants such as \textsc{GraphArena}~\citep{tang2024grapharena} and \textsc{HardcoreLogic}~\citep{liang2025hardcorelogic} extend the lens to structured objects and long-tail rule shifts. \textsc{CO-Bench}~\citep{sun2026co} sits a level above by scoring agents that design solvers, while \textsc{MathConstruct}~\citep{balunovic2025mathconstruct} applies programmatic verifiers to constructive Olympiad problems that are themselves often NP-hard.

\paragraph{Verified mathematical benchmarks}
As LLMs are deployed on longer and more open-ended mathematics, automatic verification with formal contracts becomes essential for grading at scale \citep{ju2026ai}. Verified benchmarks such as \textsc{miniF2F}~\citep{zheng2021minif2f} and \textsc{PutnamBench}~\citep{tsoukalas2024putnambench} provide olympiad- and Putnam-level targets that have driven a substantial body of prover work in Lean4 \citep{delean}. A complementary thread verifies the object a model produces rather than its proof, via Python checkers in \textsc{MathConstruct}~\citep{balunovic2025mathconstruct} or SAT-solver ground truth in \textsc{SATBench}~\citep{wei2025satbench}.

\paragraph{Adaptive and contamination-resistant benchmarks.}
Frontier capabilities now advance on a timescale that outpaces benchmark construction. Any fixed test set is on borrowed time, as leaderboards saturate, and contamination from training data becomes hard to rule out. \textsc{LiveBench}~\citep{white2024livebench} and \textsc{MathArena}~\citep{balunovic2025matharena} respond by drawing continuously from newly released competitions, the latter producing direct contamination evidence on AIME 2024. \textsc{GSM-Symbolic}~\citep{mirzadeh2024gsm} shows that even surface symbolic re-skinning of GSM8K~\citep{cobbe2021training} can move accuracy by tens of points. More recently, \textsc{MathDuels}~\citep{xu2026mathduels} lets difficulty co-evolve with participants by casting evaluation as self-play between authors and solvers, \textsc{FrontierCS}~\citep{mang2025frontiercs} curates open-ended tasks from expert programmers, and the \textsc{ARC-AGI} series~\citep{chollet2024arc, chollet2025arc} pursues memorization-resistant puzzles by design. Together, these efforts mark a shift away from one-shot benchmarks and toward evaluation infrastructure that is expected to keep producing harder problems as models improve and inspire our benchmark in this way.

\paragraph{Tool-use evaluation benchmarks.}
Tool use is now a frontier capability in its own right. Deciding when to delegate to a solver, search engine, or interpreter is at least as load-bearing as the underlying reasoning. \textsc{BFCL}~\citep{patil2025berkeley} anchors a line that scores whether tool calls are well-formed, largely orthogonal to the reasoning content of the call. Mathematical benchmarks have begun reporting paired runs with and without a Python sandbox. Both \textsc{MathConstruct}~\citep{balunovic2025mathconstruct} and \textsc{MathArena}~\citep{balunovic2025matharena} observe that code access roughly doubles accuracy at a several-fold cost increase. Most mathematical tool-use benchmarks employ a single tool call budget (\textsc{MathConstruct} uses a $4$ round budget). We identify this as a category gap and analyze tool use under varied synthetic budgets in \cref{exp:tools}.

\section{MathConstraint}
\label{sec:methodology}

\textsc{MathConstraint} is a generator-verifier loop for natural-language combinatorial decision problems. A released dataset is a finite slice of this loop. Each instance carries a prompt, a backend encoding, a solver-certified polarity, optional witness data, difficulty metadata, and a verifier contract. This section describes the generator first, then the evaluation interface used in our experiments.

\subsection{Problem Families}

\textsc{MathConstraint} contains $39$ parameterized problem types. Each type is a decision problem: the model must either produce a witness satisfying all constraints, or correctly report that no witness exists. The registry spans combinatorial designs (BIBD, Hadamard matrices, Latin squares), graph threshold problems (maximum clique, vertex cover, $k$-colorability), number-theoretic and sequence constructions (Golomb rulers, van der Waerden colorings), placement and scheduling problems ($n$-queens, social golfers), and extremal graph-construction tasks (Ramsey graphs, triangle-free graphs).

Encodings come from two solver-facing backends. The \texttt{pycsp3-models}\footnote{\url{https://github.com/xcsp3team/PyCSP3-models}} corpus supplies constraint-programming formulations solved with general-purpose CP/SAT backends~\citep{lecoutre2020pycsp3}. The \texttt{pysms} library\footnote{\url{https://github.com/markirch/sat-modulo-symmetries}} supplies SAT encodings with symmetry-breaking predicates for extremal graph families~\citep{kirchweger2024sat}. The backends are not exposed to the model in the no-tools condition; they define the ground-truth and verification semantics.

A detailed description of the problem set can be found in \cref{app:problems} and \cref{app:problems-cat}.

\subsection{Automated Generation}

Generation is profile-driven. A profile names types, parameter domains, per-type counts, prompt templates, and a seed. For each sampled parameter tuple, the generator instantiates the backend encoding, solves it with the reference solver, renders the prompt, and writes a self-contained record.

\usetikzlibrary{calc}
\begin{figure}[htbp]
  \centering
  \begin{tikzpicture}[x=1mm, y=1mm,
      data/.style    = {draw=blue!35!black!60, line width=0.4pt,
                        fill=blue!7,  rounded corners=1.5pt,
                        align=left, inner xsep=5pt, inner ysep=4pt,
                        font=\footnotesize},
      proc/.style    = {draw=black!60, line width=0.4pt,
                        fill=white,   rounded corners=1.5pt,
                        align=center, inner xsep=5pt, inner ysep=4pt,
                        minimum height=9mm, font=\footnotesize},
      encoder/.style = {proc},
      solver/.style  = {draw=orange!50!black!50, line width=0.4pt,
                        fill=orange!12, rounded corners=1.5pt,
                        align=center, inner xsep=5pt, inner ysep=4pt,
                        minimum height=9mm, font=\footnotesize},
      artifact/.style = {draw=blue!35!black!60, line width=0.4pt,
                        fill=blue!4, rounded corners=1.5pt,
                        align=left, inner sep=6pt, font=\footnotesize},
      flow/.style    = {-{Stealth[length=4pt,width=4pt]}, semithick},
      pic/.style     = {x=1.4mm, y=1.4mm},
    ]


    \node[data, anchor=west, minimum width=42mm, minimum height=28mm,
          inner ysep=4pt, inner xsep=4pt] (profile) at (0, 0) {};

    \node[anchor=north, font=\footnotesize\bfseries]
      at ($(profile.north) + (0, -2)$) {Generation profile};

    \begin{scope}[shift={($(profile.north) + (-13, -12)$)}]
      \draw[black!55, line width=0.3pt] (0,0) grid[step=1.2] (4.8, 4.8);
      \fill[blue!40!black!75] (0.6, 4.2) circle (0.38);
      \fill[blue!40!black!75] (3.0, 3.0) circle (0.38);
      \fill[blue!40!black!75] (1.8, 1.8) circle (0.38);
      \fill[blue!40!black!75] (4.2, 0.6) circle (0.38);
    \end{scope}
    \begin{scope}[shift={($(profile.north) + (-3, -12)$)}]
      \coordinate (a) at (0,    2.4);
      \coordinate (b) at (1.8,  4.4);
      \coordinate (c) at (3.8,  2.4);
      \coordinate (d) at (3,    0.3);
      \coordinate (e) at (0.8,  0.3);
      \draw[black!55, line width=0.4pt]
        (a)--(b) (b)--(c) (c)--(d) (d)--(e) (e)--(a) (a)--(c);
      \foreach \p in {a,b,c,d,e}
        \fill[blue!40!black!75] (\p) circle (0.42);
    \end{scope}
    \begin{scope}[shift={($(profile.north) + (8, -10)$)}]
      \draw[black!55, line width=0.4pt] (0, 0) -- (5, 0);
      \foreach \x in {0, 0.625, 1.25, 1.875, 2.5, 3.125, 3.75, 4.375, 5} {
        \draw[black!55, line width=0.4pt] (\x, -0.5) -- (\x, 0.5);
      }
      \fill[blue!40!black!75] (0,    0) circle (0.45);
      \fill[blue!40!black!75] (1.25, 0) circle (0.45);
      \fill[blue!40!black!75] (3.125,0) circle (0.45);
      \fill[blue!40!black!75] (5,    0) circle (0.45);
    \end{scope}

    \node[anchor=north, font=\scriptsize, align=left]
      at ($(profile.north) + (0, -17)$) {%
      $\{T_i\}$ types,\ $\Theta_i$ domains,\\
      $n_i$ counts,\ $s$ seed,\ NL templates};


    \node[encoder, anchor=west, minimum width=38mm] (enc_csp)
      at ([xshift=12mm, yshift=8mm]profile.east) {%
      \texttt{pycsp3} encoder\\
      {\sffamily\scriptsize N-queens, Golomb, Sudoku, ...}
    };
    \node[encoder, anchor=west, minimum width=38mm] (enc_sms)
      at ([xshift=12mm, yshift=-8mm]profile.east) {%
      \texttt{pysms} encoder\\
      {\sffamily\scriptsize Ramsey, chromatic, $k$-coloring, ...}
    };

    \node[solver, anchor=west, minimum width=24mm, minimum height=18mm]
      (solver) at ([xshift=8mm]$(enc_csp.east)!0.5!(enc_sms.east)$) {};

    \begin{scope}[shift={($(solver.center) + (0, 2.5)$)}]
      \def\rinner{1.5}   
      \def\router{2.4}   
      \foreach \i in {0,...,7} {
        \pgfmathsetmacro{\angA}{45*\i - 8}
        \pgfmathsetmacro{\angB}{45*\i + 8}
        \pgfmathsetmacro{\angC}{45*\i + 14}
        \pgfmathsetmacro{\angD}{45*\i + 31}
        \fill[orange!50!black!75]
          (\angA:\rinner) -- (\angA:\router) -- (\angB:\router)
          -- (\angB:\rinner) -- (\angC:\rinner) -- (\angC:\router)
          -- cycle;
      }
      \fill[orange!50!black!75] (0,0) circle (\rinner);
      \fill[orange!12] (0,0) circle (0.55);
    \end{scope}

    \node[anchor=north, font=\footnotesize]
      at ($(solver.center) + (0, -2)$) {Solvers};

    \draw[flow] (profile.east) -- ++(5,0) |- (enc_csp.west);
    \draw[flow] (profile.east) -- ++(5,0) |- (enc_sms.west);
    \draw[flow] (enc_csp.east) -- ([yshift=2mm]solver.west);
    \draw[flow] (enc_sms.east) -- ([yshift=-2mm]solver.west);


    \coordinate (pipe_center) at ($(profile.west)!0.5!(solver.east)$);
    \coordinate (pipe_bottom) at ($(profile.south)!0.5!(enc_sms.south)$);
    \node[artifact, anchor=north, text width=110mm] (inst)
      at ([yshift=-12mm]pipe_center |- pipe_bottom) {%
      \textbf{Instance}\\[2pt]
      \begin{tabular}{@{}l@{\hspace{0.8em}}l@{\hspace{2em}}l@{\hspace{0.8em}}l@{}}
        \textbf{prompt}    & \multicolumn{3}{@{}l@{}}{\texttt{"Place 4 queens on a 4$\times$4 board such that no two attack..."}} \\
        \textbf{verdict}   & \texttt{SAT}     & \textbf{solution} & \texttt{[1, 3, 0, 2]} \\
        \textbf{solver}    & \texttt{810\,ms} & \textbf{size}     & \texttt{$|V|{=}4,\ |C|{=}3$} \\
      \end{tabular}
    };

    \draw[flow] (solver.south) |- ($(inst.north) + (0,4)$) -- (inst.north);

  \end{tikzpicture}

  \caption{Generation pipeline. A profile samples from a large
  registry of problem types and a parameter domain. Each
  instance is encoded via \texttt{pycsp3} or \texttt{pysms} and
  solved by the reference back-end, producing an output bundling
  the natural-language prompt with solver-verified ground truth.}
  \label{fig:generation}
\end{figure}
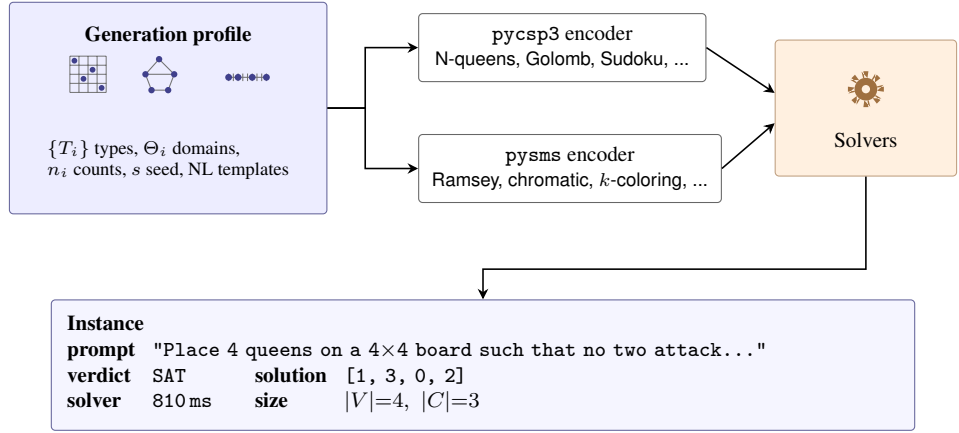

\paragraph{Polarity.} Polarity is determined by the reference solver after parameter sampling. We do not balance SAT and UNSAT by construction; the split is induced by the parameter domains of each problem type. \textsc{MathConstraint} is $69.3\%$ SAT, while \textsc{MathConstraint-Easy} is $77.4\%$ SAT. UNSAT instances are first-class: the model must identify infeasibility rather than provide a witness, and the verifier applies the same polarity contract to both tool and no-tool runs.

\paragraph{Partial assignments.} SAT instances may include a partial assignment that fixes a subset of decision-variable positions to values from a reference solution. These fixed values are appended to the prompt and enforced by the verifier. Partial assignments create a scaffolding axis orthogonal to instance size. They can reduce search while preserving the same formal constraint system.

\paragraph{Difficulty.} Each instance records reference solver wall time. We also store structural proxies such as variable counts, constraint counts, edge counts, and a search-space estimate when available. Difficulty is tunable before generation by changing profile ranges, and after generation by filtering on recorded solver metadata. 

\begin{table}[htbp]
\caption{Generation-time difficulty of the released datasets. Solver-time columns report reference solver wall time in seconds.}
\label{tab:dataset-difficulty}
\centering
\small
\renewcommand{\arraystretch}{0.96}
\setlength{\tabcolsep}{3pt}
\begin{tabular}{@{}lccccccc@{}}
\toprule
Component & $n$ & Types & SAT & p50 & p90 & p99 & Max \\
\midrule
\textsc{MathConstraint-Easy} & 266 & 25 & 206 (77.4\%) & 0.20 & 1.79 & 11.8 & 56.5 \\
\textsc{MathConstraint} & 329 & 39 & 228 (69.3\%) & 2.52 & 75.6 & 653 & 1372 \\
\bottomrule
\end{tabular}
\renewcommand{\arraystretch}{1}
\end{table}

\subsection{Verification and Tool Interface}
\label{sec:methodology:evaluation}

Verification is handled by formal contracts. A response is parsed as a satisfiability claim plus an optional solution. UNSAT claims are compared to the solver-certified polarity. SAT claims are graded up to isomorphism. The candidate is injected as hard unit constraints on top of the original encoding and re-solved, so any witness in the same canonical equivalence class is accepted. This grades the submitted object, not a textual derivation, and naturally handles multiple correct solutions.

We evaluate each model in two conditions. In \textsc{no\_tools}, the model returns only a final JSON answer. In \textsc{tools}, the model may interleave reasoning with calls to \texttt{execute\_python}, a sandboxed Python interpreter with generic SAT/SMT libraries, and must finish with \texttt{submit\_answer}. The tool budget is capped at eight rounds. If a model reaches the cap without an explicit submission, the evaluator force-submits the final natural-language response through a strict parser; we report this behavior separately in \cref{tab:force-submit-fallback}. Metric definitions are given in \cref{app:metrics}.

\usetikzlibrary{calc}
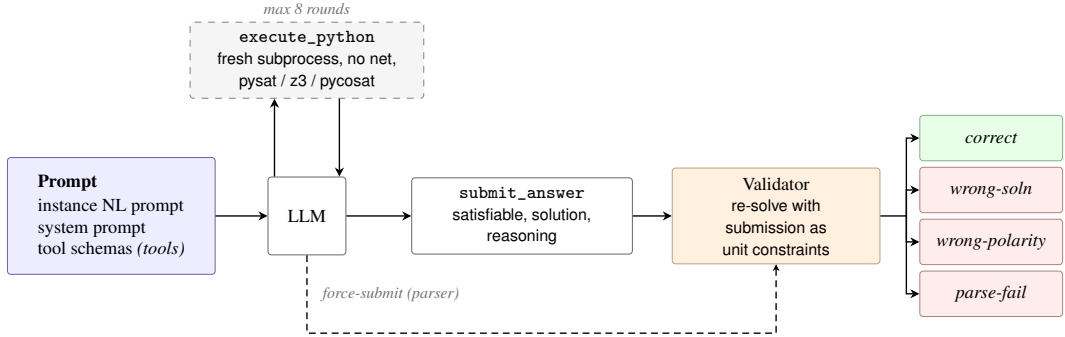
\begin{figure}[htbp]
  \centering
  \resizebox{\linewidth}{!}{%
  \begin{tikzpicture}[x=1mm, y=1mm,
      data/.style    = {draw=blue!35!black!60, line width=0.4pt,
                        fill=blue!7,  rounded corners=1.5pt,
                        align=left, inner xsep=5pt, inner ysep=4pt,
                        font=\footnotesize},
      proc/.style    = {draw=black!60, line width=0.4pt,
                        fill=white,   rounded corners=1.5pt,
                        align=center, inner xsep=5pt, inner ysep=4pt,
                        minimum height=9mm, font=\footnotesize},
      solver/.style  = {draw=orange!50!black!50, line width=0.4pt,
                        fill=orange!12, rounded corners=1.5pt,
                        align=center, inner xsep=5pt, inner ysep=4pt,
                        minimum height=9mm, font=\footnotesize},
      sandbox/.style = {draw=black!50, line width=0.4pt, dashed,
                        fill=black!4, rounded corners=1.5pt,
                        align=center, inner xsep=5pt, inner ysep=4pt,
                        minimum height=9mm, font=\footnotesize},
      v_ok/.style    = {draw=green!40!black!50, line width=0.4pt,
                        fill=green!10, rounded corners=1.5pt,
                        align=center, inner sep=4pt, minimum height=7mm,
                        minimum width=22mm, font=\footnotesize\itshape},
      v_bad/.style   = {draw=red!40!black!50, line width=0.4pt,
                        fill=red!7,  rounded corners=1.5pt,
                        align=center, inner sep=4pt, minimum height=7mm,
                        minimum width=22mm, font=\footnotesize\itshape},
      flow/.style    = {-{Stealth[length=4pt,width=4pt]}, semithick},
      loopflow/.style= {-{Stealth[length=3pt,width=3pt]}, semithick,
                        densely dashed},
      notelabel/.style = {font=\scriptsize\itshape, text=black!55,
                          inner sep=1pt},
    ]

    \node[data, anchor=west, minimum width=32mm, minimum height=18mm] (prompt)
      at (0, 0) {%
      \textbf{Prompt}\\[1pt]
      instance NL prompt\\
      system prompt\\
      tool schemas \emph{(tools)}
    };

    \node[proc, anchor=west, minimum width=12mm, minimum height=12mm] (llm)
      at ([xshift=8mm]prompt.east) {LLM};

    \node[sandbox, anchor=south, minimum width=36mm] (exec)
      at ([yshift=12mm]llm.north) {%
      \texttt{execute\_python}\\
      {\sffamily\scriptsize fresh subprocess, no net,}\\
      {\sffamily\scriptsize pysat / z3 / pycosat}
    };

    \node[proc, anchor=west, minimum width=34mm] (submit)
      at ([xshift=10mm]llm.east) {%
      \texttt{submit\_answer}\\
      {\sffamily\scriptsize satisfiable, solution,}\\
      {\sffamily\scriptsize reasoning}
    };

    \node[solver, anchor=west, minimum width=32mm] (validator)
      at ([xshift=6mm]submit.east) {%
      Validator\\
      {\sffamily\scriptsize re-solve with}\\
      {\sffamily\scriptsize submission as}\\
      {\sffamily\scriptsize unit constraints}
    };

    \node[v_ok,  anchor=west, yshift=12mm]   (v_ok)
      at ([xshift=6mm]validator.east) {correct};
    \node[v_bad, anchor=west, yshift=4mm]    (v_wrong)
      at ([xshift=6mm]validator.east) {wrong-soln};
    \node[v_bad, anchor=west, yshift=-4mm]   (v_pol)
      at ([xshift=6mm]validator.east) {wrong-polarity};
    \node[v_bad, anchor=west, yshift=-12mm]  (v_parse)
      at ([xshift=6mm]validator.east) {parse-fail};

    \draw[flow] (prompt.east) -- (llm.west);
    \draw[flow] ([xshift=-5mm]llm.north) -- ([xshift=-5mm]exec.south);
    \draw[flow] ([xshift=5mm]exec.south) -- ([xshift=5mm]llm.north);
    \node[notelabel] at ([yshift=2mm]exec.north) {max 8 rounds};

    \draw[flow] (llm.east) -- (submit.west);
    \draw[flow] (submit.east) -- (validator.west);

    \draw[loopflow]
      (llm.south) -- ([yshift=-12mm]llm.south)
      node[notelabel, midway, anchor=west, xshift=2mm] {force-submit (parser)}
      -| (validator.south);

    \coordinate (vstub) at ([xshift=4mm]validator.east);
    \draw[flow] (validator.east) -- (vstub) |- (v_ok.west);
    \draw[flow] (vstub) |- (v_wrong.west);
    \draw[flow] (vstub) |- (v_pol.west);
    \draw[flow] (vstub) |- (v_parse.west);

  \end{tikzpicture}%
  }

  \caption{Evaluation pipeline. The model interleaves
  \texttt{execute\_python} (sandboxed, max $8$ rounds) and
  \texttt{submit\_answer}; on budget exhaustion the parser
  force-submits the last NL response. The validator re-solves with
  the submission injected as constraints, distinguishing four
  outcomes.}
  \label{fig:evaluation}
\end{figure}

\subsection{Adaptivity}
\label{sec:methodology:adaptivity}

\textsc{MathConstraint}'s adaptivity is a crucial property of the framework. The generator-verifier-filter loop (\cref{fig:adaptivity}) can be rerun by changing parameter ranges, changing the frontier-model admission filter, or adding new families from \texttt{pycsp3-models} (which contains over 400 problem models) and \texttt{pysms} (for graph problems). Like \textsc{MathArena}, we treat benchmark refresh as an explicit release cycle rather than a one-time dataset construction, and adopt a similar refresh cadence. However, each refresh is manufactured from our own parameterized, solver-backed registry rather than harvested from new external competitions~\citep{balunovic2025matharena}. \textsc{MathConstraint-Easy} and \textsc{MathConstraint} are the first two slices from this loop. \textsc{MathConstraint-Easy} is retained as a lower-difficulty slice, while the headline \textsc{MathConstraint} release reruns the same machinery with harder profile regimes and admits instances that remain discriminative under frontier-model evaluation.

\usetikzlibrary{calc, shapes.geometric}
\begin{figure}[htbp]
  \centering
  \resizebox{0.42\linewidth}{!}{%
  \begin{tikzpicture}[x=1mm, y=1mm,
      data/.style    = {draw=blue!35!black!60, line width=0.6pt,
                        fill=blue!10, rounded corners=1.5pt,
                        align=center, inner xsep=3pt, inner ysep=2pt,
                        font=\scriptsize\bfseries,
                        minimum height=8mm, minimum width=18mm},
      proc/.style    = {draw=black!60, line width=0.4pt,
                        fill=white,   rounded corners=1.5pt,
                        align=center, inner xsep=3pt, inner ysep=2pt,
                        font=\scriptsize, minimum height=8mm,
                        minimum width=18mm},
      generator/.style = {draw=orange!50!black!50, line width=0.4pt,
                        fill=orange!12, rounded corners=1.5pt,
                        align=center, inner xsep=3pt, inner ysep=2pt,
                        font=\scriptsize, minimum height=8mm,
                        minimum width=18mm},
      optinput/.style = {draw=black!50, line width=0.4pt, dashed,
                        fill=white, rounded corners=1.5pt,
                        align=center, inner xsep=2pt, inner ysep=1pt,
                        font=\scriptsize, minimum height=5mm,
                        minimum width=14mm},
      gate/.style    = {draw=black!65, line width=0.4pt,
                        fill=yellow!8, diamond, aspect=2,
                        align=center, inner xsep=1pt, inner ysep=1pt,
                        font=\scriptsize, minimum width=18mm,
                        minimum height=10mm},
      drop/.style    = {draw=red!40!black!50, line width=0.4pt,
                        fill=red!7, rounded corners=1.5pt,
                        align=center, inner xsep=3pt, inner ysep=1pt,
                        font=\scriptsize\itshape, minimum height=5mm,
                        minimum width=11mm},
      flow/.style    = {-{Stealth[length=4pt,width=4pt]}, semithick},
      droparrow/.style = {-{Stealth[length=3pt,width=3pt]}, semithick,
                        draw=black!45},
      loopin/.style  = {-{Stealth[length=3pt,width=3pt]}, semithick,
                        densely dashed, draw=black!55},
      selfloop/.style = {-{Stealth[length=3pt,width=3pt]}, semithick,
                        draw=black!55},
      gatelabel/.style = {font=\scriptsize, text=black!70, inner sep=1pt},
      annot/.style   = {font=\scriptsize\itshape, text=black!65,
                        inner sep=1pt},
    ]

    \coordinate (cc) at (0, 0);

    \node[generator] (gen)  at ($(cc) + (0,  12)$) {Generator};
    \node[proc]      (eval) at ($(cc) + (24,  0)$) {%
      Evaluation\\
      {\scriptsize\,$+$ verification}};
    \node[gate]      (g)    at ($(cc) + (0, -12)$) {%
      $\geq 1$ frontier\\ failed?};
    \node[data]      (mc)   at ($(cc) + (-24, 0)$) {%
      \textsc{Math-}\\\textsc{Constraint}};

    \draw[flow] (gen.east)   to[bend left=15] (eval.north);
    \draw[flow] (eval.south) to[bend left=15] (g.east);
    \draw[flow] (g.west)     to[bend left=15] node[gatelabel, midway, below=1pt]
                {yes} (mc.south);
    \draw[flow] (mc.north)   to[bend left=15] (gen.west);

    \draw[selfloop] (gen.north) ++(-2.5,0)
      to[out=120, in=60, looseness=8] node[annot, above=1pt]
      {cranked params}
      ([xshift=2.5mm]gen.north);

    \node[drop, anchor=north] (dropb) at ([yshift=-6mm]g.south) {drop};
    \draw[droparrow] (g.south) -- node[gatelabel, right] {no} (dropb.north);

    \node[optinput, anchor=south east] (newtypes)
      at ([xshift=-1mm, yshift=4mm]gen.north west) {%
      new types\\
      {\scriptsize\itshape (optional)}};
    \draw[loopin] (newtypes.south east) -- (gen.north west);

  \end{tikzpicture}%
  }

  \caption{Adaptivity loop. The generator produces candidates;
  each is evaluated and verified, and admitted into
  \textsc{MathConstraint} only if at least one frontier model
  fails. \textsc{MathConstraint} feeds back into the generator,
  which cranks parameters across cycles (self-loop) and may
  optionally introduce new problem types. The benchmark refreshes
  itself as models improve.}
  \label{fig:adaptivity}
\end{figure}
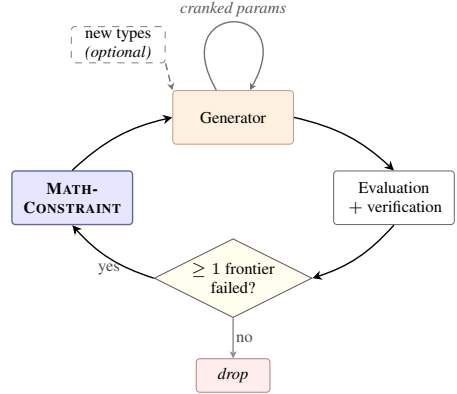

\section{Evaluation}
\label{sec:eval}

\paragraph{Setup.}
We evaluate twelve models. Ten are frontier or near-frontier systems: \textsc{gpt-5.5}~\citep{openai2026gpt55systemcard}, \textsc{claude-opus-4.7}~\citep{anthropic2026opus47systemcard}, \textsc{claude-4.6-sonnet}~\citep{anthropic2026sonnet46systemcard}, \textsc{gemini-3.1-pro} and \textsc{gemini-3.1-flash-lite}~\citep{google2026gemini31prosystemcard}, \textsc{grok-4.20}~\citep{xai2025grok4}, \textsc{deepseek-v4-pro} and \textsc{deepseek-v4-flash}~\citep{deepseekai2026deepseekv4}, \textsc{qwen3.6-plus}~\citep{qwen36plus}, and \textsc{kimi-k2.6}~\citep{kimiteam2025kimik2openagentic}. The remaining two are open-weight baselines: \textsc{gpt-oss-120b}~\citep{agarwal2025gpt} and \textsc{llama-3.3-70b-instruct}~\citep{grattafiori2024llama}. Models are accessed through OpenRouter at temperature $0$; model configuration, pricing, and implementation details are in \cref{app:models,app:impl}.

\subsection{Main Results}
\label{exp:main}

\paragraph{The adaptive refresh: \textsc{Easy} to \textsc{MathConstraint}.}
\cref{tab:mathconstraint-no-tools,tab:easy-combined} show that rerunning the generator with harder profiles restores discrimination. Among frontier and near-frontier systems, mean no-tools accuracy falls from $80.5\%$ on \textsc{MathConstraint-Easy} to $42.0\%$ on \textsc{MathConstraint}; across all twelve models, the number above $70\%$ drops from $10$ to $0$, and the number above $50\%$ drops from $11$ to $3$. The headline \textsc{MathConstraint} benchmark remains challenging and unsaturated: \textsc{gpt-5.5} solves $220/329$ ($66.9\%$), \textsc{grok-4.20} solves $191/329$ ($58.1\%$), and the cohort spans down to \textsc{claude-4.6-sonnet} at $18.5\%$.

\paragraph{The witness gap.}
Across no-tools runs, SAT-polarity accuracy systematically exceeds end-to-end verifier accuracy. The gap (\cref{tab:witness-gap}) is largest for \textsc{llama-3.3-70b-instruct} ($+54.4$\,pp), \textsc{claude-4.6-sonnet} ($+53.5$\,pp), and \textsc{gemini-3.1-flash-lite} ($+51.4$\,pp). Trace audits show two contributing patterns: models declare SAT and submit invalid witnesses, or fail to find a witness for a SAT instance and resolve UNSAT. Deciding existence is easier than constructing an object that survives exact verification. The witness gap is the dominant no-tools failure mode and the lever that tools pull on \cref{exp:tools}.

\paragraph{Free-UNSAT baseline.}
$30.7\%$ of \textsc{MathConstraint} is UNSAT, so a model that always answers UNSAT scores $30.7\%$ overall and $100\%$ on the UNSAT subset. By contrast, \textsc{claude-4.6-sonnet} ($16.8\%$ UNSAT-conditional accuracy) and \textsc{llama-3.3-70b-instruct} ($24.8\%$) perform far below the UNSAT-subset ceiling. They actively guess wrong on UNSAT rather than free-riding on polarity

\begin{table}[htbp]
\caption{No-tools performance on the full 329-instance \textsc{MathConstraint} release. Accuracy and SAT Acc. are defined in \cref{app:metrics}.}
\label{tab:mathconstraint-no-tools}
\centering
\small
\renewcommand{\arraystretch}{0.96}
\setlength{\tabcolsep}{3pt}
\begin{tabular}{@{}lccc@{}}
\toprule
Model & Correct & Accuracy (\%) & SAT Acc. (\%) \\
\midrule
\textsc{gpt-5.5} & \textbf{220} & \textbf{66.9} & \textbf{79.0} \\
\textsc{claude-opus-4.7} & 140 & 42.6 & 56.8 \\
\textsc{claude-4.6-sonnet} & 61 & 18.5 & 72.0 \\
\textsc{gemini-3.1-pro} & 170 & 51.7 & 67.2 \\
\textsc{gemini-3.1-flash-lite} & 82 & 24.9 & \underline{76.3} \\
\textsc{grok-4.20} & \underline{191} & \underline{58.1} & 67.8 \\
\midrule
\textsc{deepseek-v4-pro} & 130 & 39.5 & 49.2 \\
\textsc{deepseek-v4-flash} & 112 & 34.0 & 61.4 \\
\textsc{qwen3.6-plus} & 156 & 47.4 & 72.3 \\
\textsc{kimi-k2.6} & 118 & 35.9 & 42.9 \\
\midrule
\textsc{gpt-oss-120b} & 130 & 39.5 & 47.7 \\
\textsc{llama-3.3-70b-instruct} & 28 & 8.5 & 62.9 \\
\bottomrule
\end{tabular}
\renewcommand{\arraystretch}{1}
\end{table}

\paragraph{Backend asymmetry.}
Stratifying by backend reveals a structural gap. On \textsc{MathConstraint} no-tools, mean accuracy across all twelve models is $61.1\%$ on \texttt{pysms} instances versus $29.0\%$ on \texttt{pycsp}; the six-model frontier cohort shows a similar split, $66.5\%$ versus $33.6\%$. Extremal-graph \texttt{pysms} instances admit many witnesses with structural symmetry, while \texttt{pycsp} encodings often constrain a unique numeric object. Tools narrow the gap \cref{exp:tools}. Full per-model stratification can be found in \cref{tab:stratification-mathconstraint-no-tools}.

\begin{table}[htbp]
\caption{Performance on \textsc{MathConstraint-Easy} ($n=266$). Accuracy is verifier acceptance; SAT Acc.\ is polarity accuracy (\cref{app:metrics}). The same frontier models that span $18.5$--$66.9\%$ on \textsc{MathConstraint} (\cref{tab:mathconstraint-no-tools}) cluster between $72.6\%$ and $87.6\%$ here without tools and approach saturation with tools.}
\label{tab:easy-combined}
\centering
\small
\renewcommand{\arraystretch}{0.96}
\setlength{\tabcolsep}{3pt}
\begin{tabular}{@{}lcccc@{}}
\toprule
 & \multicolumn{2}{c}{no\_tools} & \multicolumn{2}{c}{tools} \\
\cmidrule(lr){2-3}\cmidrule(lr){4-5}
Model & Acc.\ (\%) & SAT Acc.\ (\%) & Acc.\ (\%) & SAT Acc.\ (\%) \\
\midrule
\textsc{gpt-5.5} & \textbf{87.6} & 91.4 & \textbf{91.7} & \textbf{96.6} \\
\textsc{claude-opus-4.7} & \underline{85.3} & 88.7 & 90.2 & 94.7 \\
\textsc{claude-4.6-sonnet} & 75.2 & \underline{94.0} & 89.8 & 94.7 \\
\textsc{gemini-3.1-pro} & 84.2 & 89.8 & \underline{91.4} & \underline{95.5} \\
\textsc{gemini-3.1-flash-lite} & 72.6 & \textbf{94.4} & 89.5 & 94.0 \\
\textsc{grok-4.20} & 85.0 & 87.2 & 87.2 & 92.5 \\
\midrule
\textsc{deepseek-v4-pro} & 82.7 & 85.7 & 71.8 & 75.2 \\
\textsc{deepseek-v4-flash} & 74.4 & 86.5 & 71.4 & 78.2 \\
\textsc{qwen3.6-plus} & 81.2 & 87.6 & 88.3 & 93.6 \\
\textsc{kimi-k2.6} & 77.1 & 81.2 & 82.3 & 86.8 \\
\midrule
\textsc{gpt-oss-120b} & 69.2 & 75.9 & 71.4 & 84.6 \\
\textsc{llama-3.3-70b-instruct} & 22.9 & 52.6 & 27.8 & 37.6 \\
\bottomrule
\end{tabular}
\renewcommand{\arraystretch}{1}
\end{table}

\FloatBarrier
\subsection{Tool-use}
\label{exp:tools}

\cref{tab:mathconstraint-tools} evaluates the same \textsc{MathConstraint} instances with a sandboxed Python tool budget. Tool access changes both absolute accuracy and the ranking. \textsc{gpt-5.5} rises from $66.9\%$ to $80.9\%$, \textsc{claude-4.6-sonnet} rises from $18.5\%$ to $70.5\%$, and \textsc{gemini-3.1-pro} rises from $51.7\%$ to $72.9\%$. The gains are not uniform: models that already emit compact, direct answers sometimes improve only modestly, while models that can synthesize solver encodings from the prompt gain sharply. Thus tool access measures the additional skill of translating informal constraints into executable search under a fixed interaction budget. Representative traces in \cref{app:traces} illustrate these regimes.

\begin{table}[htbp]
\caption{Tool-enabled performance on the full 329-instance \textsc{MathConstraint} release. Accuracy and SAT Acc. are defined in \cref{app:metrics}.}
\label{tab:mathconstraint-tools}
\centering
\small
\renewcommand{\arraystretch}{0.96}
\setlength{\tabcolsep}{3pt}
\begin{tabular}{@{}lccc@{}}
\toprule
Model & Correct & Accuracy (\%) & SAT Acc. (\%) \\
\midrule
\textsc{gpt-5.5} & \textbf{266} & \textbf{80.9} & \textbf{94.2} \\
\textsc{claude-opus-4.7} & \underline{255} & \underline{77.5} & 90.9 \\
\textsc{claude-4.6-sonnet} & 232 & 70.5 & \underline{91.8} \\
\textsc{gemini-3.1-pro} & 240 & 72.9 & 86.3 \\
\textsc{gemini-3.1-flash-lite} & 205 & 62.3 & 87.5 \\
\textsc{grok-4.20} & 219 & 66.6 & 83.3 \\
\midrule
\textsc{deepseek-v4-pro} & 136 & 41.3 & 51.1 \\
\textsc{deepseek-v4-flash} & 120 & 36.5 & 47.7 \\
\textsc{qwen3.6-plus} & 218 & 66.3 & 83.3 \\
\textsc{kimi-k2.6} & 197 & 59.9 & 71.4 \\
\midrule
\textsc{gpt-oss-120b} & 120 & 36.5 & 57.8 \\
\textsc{llama-3.3-70b-instruct} & 105 & 31.9 & 48.0 \\
\bottomrule
\end{tabular}
\renewcommand{\arraystretch}{1}
\end{table}

\paragraph{Tools substitute for inference-time reasoning.}
The cohort's largest tool-induced gain is \textsc{claude-4.6-sonnet}: $18.5\%$ no-tools $\to$ $70.5\%$ ($+52$\,pp). Sonnet runs in our setup with no extended-thinking budget (\cref{app:models}). With less inference-time reasoning, it commits to invalid witnesses, producing the cohort's largest witness gap ($+53.5$\,pp) (\cref{tab:witness-gap}). Tool access restores the missing computation. Solver calls externalize witness construction, and the witness gap closes. Tool budget and inference-time reasoning are partially fungible. \textsc{sim@}$k$ \cref{exp:budget} measures how much orchestration depth the substitution requires.

\begin{figure}[t]
    \centering
    \begin{minipage}[t]{0.48\linewidth}
        \centering
        \includegraphics[width=\linewidth]{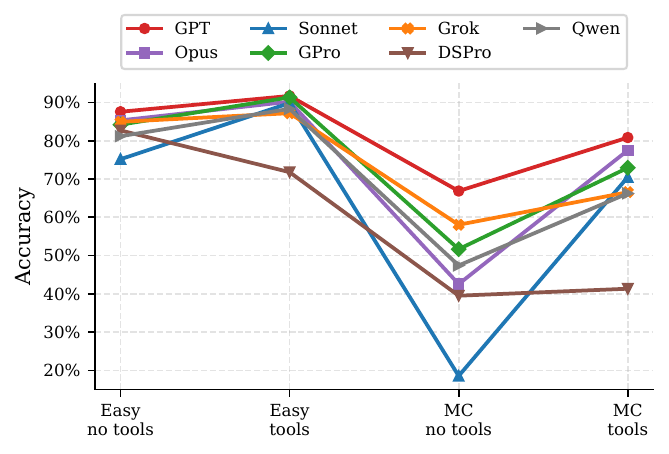}
        \caption{Accuracy across datasets and tool conditions for selected models. The plotted aggregates are reported in \cref{tab:mathconstraint-no-tools,tab:mathconstraint-tools,tab:easy-combined}.}
        \label{fig:four-condition-accuracy}
    \end{minipage}\hfill
    \begin{minipage}[t]{0.48\linewidth}
        \centering
        \includegraphics[width=\linewidth]{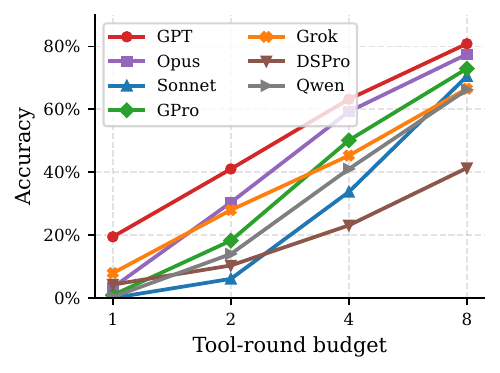}
        \caption{\textsc{sim@}$k$ curves on \textsc{MathConstraint}. Halving the budget from $8$ to $4$ rounds erases up to $37$\,pp. \Cref{tab:mathconstraint-tool-budget} reports sensitivities for all models.}
        \label{fig:simk-tool-budget}
    \end{minipage}
\end{figure}

On \textsc{MathConstraint-Easy}, tools further compress the frontier-model spread. Top models cluster around $90\%$ verifier accuracy (\cref{tab:easy-combined}). This reinforces \textsc{MathConstraint-Easy}'s position as a useful but less discriminative slice rather than the headline benchmark, motivating the adaptive construction of \textsc{MathConstraint} in \cref{sec:methodology:adaptivity}. A four-condition comparison across datasets and tool settings is provided in \cref{fig:four-condition-accuracy}.

\paragraph{Tool fingerprints.}
Models also differ in what they call. On \textsc{MathConstraint} \texttt{tools}, \textsc{grok-4.20} uses \texttt{z3} for $44\%$ of its \texttt{execute\_python} calls; \textsc{gemini-3.1-pro} balances \texttt{pysat} ($32\%$) and \texttt{z3} ($26\%$); the \textsc{claude} models default to brute-force loops ($60\%$ for \textsc{sonnet}, $54\%$ for \textsc{opus}) with \texttt{pysat} as the secondary mode; \textsc{llama-3.3-70b-instruct} relies on \texttt{networkx} ($50\%$). A full breakdown can be found in \cref{tab:tool-call-composition}.

\subsection{\textsc{sim@}k: Tool-Budget Replay}
\label{exp:budget}
A single $8$-round number hides large variance (\cref{tab:mathconstraint-tool-budget}). We replay logged tool trajectories under smaller caps: \textsc{sim@}$k$ truncates each trace after at most $k$ tool rounds and re-applies the verifier and force-submit logic (\cref{app:metrics}). \textsc{sim@}$k$ is a conservative lower bound on what a budget-aware policy could achieve; a model told it has $k$ rounds in advance might prune exploratory calls and do better.


The curve shapes expose policy differences that a scalar leaderboard hides. \textsc{llama-3.3-70b-instruct} is early-saturating: it reaches its full replayed accuracy by \textsc{sim@}2. \textsc{gpt-5.5} and \textsc{grok-4.20} are steadier, with useful work spread across the budget. \textsc{claude-4.6-sonnet} and \textsc{gemini-3.1-flash-lite} are late-gain profiles, with much of their accepted mass appearing only after four rounds. These patterns indicate that most successful tool runs are not one-shot solver calls, but multi-round attempts to formulate, debug, execute, and submit a certificate.

The same behavior appears from two other angles in the appendix. Models with steep \textsc{sim@}k curves also tend to use many rounds on \textsc{MathConstraint} (\cref{tab:tool-use-behavior-mathconstraint}), and runs without explicit \texttt{submit\_answer} can still be rescued by final-response parsing (\cref{tab:force-submit-fallback}). This rescue effect is substantial for some models: \textsc{gpt-5.5} passes $67/88$ force-submitted traces ($76.1\%$), while \textsc{deepseek-v4-flash} passes only $19/180$ ($10.6\%$). Thus tool-budget sensitivity, round usage, and force-submit rescue are three views of the same agentic behavior.

\FloatBarrier

\subsection{Difficulty and Efficiency}
\label{exp:diff}

\paragraph{Parameter ranges as an LLM difficulty dial.}
\textsc{MathConstraint} is harder at generation time than \textsc{MathConstraint-Easy}: median solver time rises from $0.20$ to $2.52$ seconds, and p99 rises from $11.8$ to $653$ seconds (\cref{tab:dataset-difficulty}). The same shift transfers to models. Frontier no-tools accuracy drops from the $72.6$--$87.6\%$ band on \textsc{MathConstraint-Easy} to $18.5$--$66.9\%$ on \textsc{MathConstraint}, reopening separation that the easy slice largely compresses (\cref{tab:easy-combined,tab:mathconstraint-no-tools}).

\paragraph{Solver time is not a within-regime proxy.}
Within the hard release, reference solver time weakly predicts model accuracy: log solver time correlates only $-0.15$ with no-tools cross-model accuracy and $-0.14$ with tool-enabled accuracy, and \cref{fig:solver-time-accuracy} is roughly flat across solver-time quartiles. We treat this as a diagnostic result rather than a failure of the difficulty dial. Cranked ranges reliably harden the distribution, but ranking individual natural-language instances by LLM difficulty requires features that solver wall time alone does not expose.

Tool use also changes efficiency. Some models become both more accurate and cheaper per accepted answer because tool calls shorten long free-form reasoning traces; others spend many rounds encoding or debugging. We report per-instance tokens, latency, and provider-billed cost in \cref{tab:eval-efficiency-mathconstraint-no-tools,tab:eval-efficiency-mathconstraint-tools,tab:eval-efficiency-easy-no-tools,tab:eval-efficiency-easy-tools}, token-type breakdowns in \cref{tab:token-breakdown-mathconstraint-no-tools,tab:token-breakdown-mathconstraint-tools}, and the cost--accuracy scatter in \cref{fig:cost-accuracy}.

\FloatBarrier

\subsection{Diagnostics}
\label{exp:diagnostics}

Additional diagnostics appear in the appendix: model-correctness correlations, polarity/backend stratification, failure buckets, and per-type breakdowns (\cref{fig:model-correlations,tab:stratification-mathconstraint-no-tools,tab:stratification-mathconstraint-tools,tab:failure-buckets-no-tools,tab:failure-buckets-tools,fig:failure-buckets,tab:per-type-mathconstraint-no-tools,tab:per-type-mathconstraint-tools,tab:per-type-easy-no-tools,tab:per-type-easy-tools}).

\section{Conclusion}

\textsc{MathConstraint} is designed as evaluation infrastructure for combinatorial reasoning rather than another fixed benchmark. By grounding each instance in solver-certified satisfiability and verifier-checked witnesses, the framework can be rerun at harder settings as models improve. The \textsc{MathConstraint-Easy}$\to$\textsc{MathConstraint} transition is the first example of that refresh cycle: the same generator-verifier machinery, drawing harder parameters and problems from our two solver-backed registries (\texttt{pycsp3-models} and \texttt{pysms}), restores separation between frontier models that the easier benchmark had begun to compress.

Our experiments also show that tool-enabled evaluation measures a different capability, i.e., whether models can turn natural language constraints into working solver programs under limited budgets. On the harder release, access to a sandboxed Python environment with generic SAT/SMT solvers raises frontier mean accuracy by $28$ points, with gains as large as $52$ points for \textsc{claude-4.6-sonnet}. The \textsc{sim@}$k$ analysis makes the same point from another angle: halving the tool budget from $8$ to $4$ rounds erases up to $37$ points of accuracy. Strong reasoning increasingly depends not only on finding the right answer, but also on using external computation effectively.

Our study also has limitations. The released benchmarks use a fixed prompt format and a fixed maximum tool budget, and \textsc{sim@}$k$ only approximates how models would behave under smaller native budgets. Future work could explore adaptive prompting, budget-aware policies, and broader classes of reasoning tools.


More broadly, \textsc{MathConstraint} moves toward a different kind of reasoning benchmark that is adaptive, solver-verified, and tool-aware. As models increasingly rely on external computation, future evaluations should measure not only reasoning ability, but also how effectively models coordinate search, verification, and tool use together.


\FloatBarrier

\section*{Acknowledgments}
We gratefully acknowledge KNOWIDEA Technologies for providing model inference
credits used in our evaluations. We also thank Vidhi Patra for helpful
discussions and feedback.

\bibliographystyle{plainnat}
\bibliography{refs}


\appendix

\section{Implementation Details}
\label{app:impl}

This appendix records the implementation choices needed to reproduce
the generation and evaluation contract.

\paragraph{Generation.} For \texttt{pysms} graph problems we solve with PySAT/Cadical and bypass \texttt{smsg}'s canonical-form filter, which produced false negatives on non-canonical valid graphs. For \texttt{pycsp} problems we use the upstream-recommended backend (Cadical, z3, pycosat, or AbsCon) per type. Each candidate instance is bounded by a $3600$\,s reference-solver cap; timeouts are dropped rather than released. Generation parallelizes across CPU workers with per-worker checkpointing, so a profile can resume without duplicating accepted instances.

\paragraph{Sandbox.} \texttt{execute\_python} dispatches each tool call to a fresh subprocess in a pinned virtualenv with the Python standard library plus \texttt{python-sat}, \texttt{pycosat}, and \texttt{z3-solver==4.13.0.0}. Network access and outbound \texttt{subprocess} spawning are blocked, and the \texttt{math\_constraint}, \texttt{pycsp3}, and \texttt{pysms} packages are excluded so models cannot reach the benchmark verifier. Per-call wall time defaults to $15$\,s with a $60$\,s hard cap; \texttt{stdout}/\texttt{stderr} are truncated at $32$\,KB. Traces were audited for sandbox-escape attempts; none were found.

\paragraph{Force-submit fallback.} If a tool-enabled run reaches the round cap without calling \texttt{submit\_answer}, the evaluator parses the final natural-language response as a fallback: strict JSON, then fenced JSON blocks (last valid block when several appear), then brace-balanced extraction anchored at the first \texttt{"satisfiable"} key. The model is not told the round budget, so logged \texttt{tool\_call\_chain} sequences can be replayed at smaller caps for the \textsc{sim@}$k$ analysis.

\paragraph{Verification.} For SAT claims, the submitted solution is added as hard unit constraints on top of the original encoding and re-solved by the reference backend (PySAT/Cadical only for \texttt{pysms}); witnesses are accepted up to backend-level isomorphism. For UNSAT claims, only polarity is checked. Hinted \textsc{MathConstraint-Easy} instances additionally require compliance with the fixed partial assignment. Each verification attempt has a $60$\,s hard timeout, recorded in the trace rather than treated as accepted.

\paragraph{System prompt.} Identical across all models and conditions.
The italicized block is concatenated only in the \texttt{tools}
condition.
\begin{verbatim}
You are solving constraint satisfaction problems. Analyze the problem
carefully and determine:
1. Whether the problem is satisfiable (SAT) or unsatisfiable (UNSAT)
2. If satisfiable, provide a valid solution

Respond with a JSON object in the following format:
{
    "satisfiable": true/false,
    "solution": [list of values] or null if unsatisfiable,
    "reasoning": "Brief explanation of your reasoning"
}

Important:
- For satisfiability, true means a solution exists, false means it's
  mathematically impossible
- Solutions should be provided as a list of integers
- Be precise with your answer - incorrect solutions will be rejected
[tools-only:]
You may optionally call tools before submitting your final answer:
- execute_python(code, timeout_seconds): run Python code in an isolated
  subprocess. Standard library plus pysat/z3/pycosat available; cannot
  import math_constraint/pycsp3/pysms, access network, or spawn
  subprocesses. Default timeout 15s, max 60s.
- submit_answer(satisfiable, solution, reasoning): submit your final
  answer.
If you use tools, always finish by calling submit_answer.
\end{verbatim}

\section{Methodological Notes}
\label{app:method-notes}

This appendix collects three points of justification for methodological design choices beyond what the main text provides.

\subsection{Frozen dataset, single generator seed}
\label{app:method-notes:seeds}

\textsc{MathConstraint} is released as two frozen slices
(\textsc{Easy} and \textsc{MathConstraint}) generated under fixed
profile seeds. We do not resample per evaluation, and we do not
release multiple seeded draws of the same profile. Frozen instances
are necessary for reproducible comparison across future models, and
multi-seed generation would leak difficulty variance into every
reported number since two draws from the same profile produce slices
of different empirical hardness. Adaptivity enters through explicit
refresh cycles described in \cref{sec:methodology:adaptivity} rather than
per-evaluation resampling, and the \textsc{Easy}$\to$\textsc{MathConstraint}
progression is the first such refresh. Model-side, all results use
temperature $0$ with one run per (model, instance, condition) cell,
following standard practice for frontier-API benchmarks. With $n=329$
instances, the binomial standard error is at most $2.8$\,pp at $50\%$
accuracy and below $2.3$\,pp at top-model accuracy levels, well under
the gaps driving the headline claims.

\subsection{The frontier-failure admission filter}
\label{app:method-notes:filter}

The adaptivity loop in \cref{fig:adaptivity} admits a candidate
only if at least one frontier model fails it. This is a common sense choice because
a refresh whose new instances are already solved by the cohort would not restore discrimination,
which is the entire purpose of running the loop. The filter does
couple the released slice to the cohort used at construction time, so
a model evaluated post-release stands in a slightly different
relationship to the benchmark than a cohort member; we treat this as
a property of filter-based adaptive evaluation rather than an artifact
specific to \textsc{MathConstraint}.

\subsection{\textsc{sim@}$k$ as a lower bound}
\label{app:method-notes:simk}

Models are never told their tool-round budget; the cap of $8$ is
enforced by the harness, not communicated in the prompt
(\cref{app:impl}). \textsc{sim@}$k$ takes the traces
produced under that condition and truncates them to at most $k$
rounds. It is therefore a lower bound in two ways. First, the original
$8$-round runs themselves were budget-unaware, so the trajectories
include exploration that a budget-aware policy might prune. Second,
truncating at $k<8$ scores a strictly shorter prefix of an already
budget-unaware trace; a model told in advance that it had $k$ rounds
could allocate differently and likely do better. We retain this
framing intentionally. Native-budget reruns at every $k$ for every
model would multiply evaluation cost without changing the qualitative
finding, which is that \emph{measured} tool-use capability can
collapse by tens of points as the cap shrinks (\textsc{Sonnet}:
$70.5 \to 33.7$ at $k=4$). A floor that low is informative regardless
of whether budget-aware policies could lift it. We therefore treat
\textsc{sim@}$k$ as a cheap, conservative diagnostic that exposes
policy structure invisible to a single-budget number, and leave
native-budget studies to future work that targets specific models
rather than the full cohort.

\subsection{Thin-tail problem types}
\label{sec:thin-tail}

A handful of problem types in \textsc{MathConstraint} are represented by very few instances. Per-type accuracies for $n \leq 2$ types should be read as existence proofs rather than rates; the benchmark is not designed to support per-type significance testing on these tail entries. We retain them to preserve registry coverage of qualitatively distinct combinatorial phenomena that would otherwise drop out of the headline aggregates.

\FloatBarrier
\section{Evaluation Metrics}
\label{app:metrics}

Let \(D\) be an evaluated dataset and let \(y_i \in \{\mathrm{SAT},\mathrm{UNSAT}\}\)
be the solver-certified polarity of instance \(i\). The evaluator parses a
model output into a submitted polarity \(\hat{y}_i\) and, for SAT claims, an
optional witness \(\hat{x}_i\). Let
\[
  p_i = \mathbf{1}\{\hat{y}_i = y_i\}
\]
denote polarity correctness, and let
\[
  v_i = \mathbf{1}\{(\hat{y}_i = \mathrm{UNSAT} \land y_i=\mathrm{UNSAT})
  \lor (\hat{y}_i = \mathrm{SAT} \land y_i=\mathrm{SAT}
  \land \hat{x}_i \text{ satisfies the verifier})\}
\]
denote end-to-end verifier acceptance.

\paragraph{Accuracy and SAT Acc.}
The main accuracy metric is verifier acceptance:
\[
  \mathrm{Accuracy}(D) = \frac{1}{|D|}\sum_{i \in D} v_i.
\]
\textsc{SAT Acc.} is polarity accuracy:
\[
  \mathrm{SATAcc}(D) = \frac{1}{|D|}\sum_{i \in D} p_i.
\]
Thus \textsc{SAT Acc.} can be high even when accuracy is low: a model may
correctly decide that a SAT instance is satisfiable but still submit an invalid
witness. On UNSAT instances, correct polarity is sufficient for verifier
acceptance because no witness is required.

\paragraph{Tool-budget replay.}
For a logged tool-enabled trace \(i\), let \(r_i\) be the number of tool rounds
used before the accepted final submission in the original run. We report
\[
  \mathrm{sim@}k(D) = \frac{1}{|D|}\sum_{i \in D} v_i\,\mathbf{1}\{r_i \le k\}.
\]
This is a conservative replay metric: it truncates observed traces and applies
the same parsing and verification logic, but it does not allow a model to adapt
its strategy after being told a smaller budget. Consequently \textsc{sim@}k is
best interpreted as a lower bound on what the same model might achieve if run
natively with a \(k\)-round budget and an adapted policy.

\paragraph{Force-submit and rescue.}
An explicit submission is an instance in which the model calls
\texttt{submit\_answer}. A force-submitted instance is one with no explicit
submission, so the evaluator parses the final natural-language response. The
rescue rate is
\[
  \mathrm{RescueRate} =
  \frac{\sum_i v_i\,\mathbf{1}\{\text{force-submitted } i\}}
       {\sum_i \mathbf{1}\{\text{force-submitted } i\}},
\]
with value \(0\) when the denominator is zero.

\paragraph{Efficiency and behavior.}
Token, latency, and cost columns are per-instance means from provider usage
logs. Tool calls and rounds are per-instance means; execute and submit columns
are total tool-call counts. Failure-bucket tables partition non-accepted runs
after parsing and verification, while stratification tables recompute
\(\mathrm{Accuracy}\) on SAT, UNSAT, \texttt{pycsp}, and \texttt{pysms}
subsets.

\FloatBarrier
\section{Extended Dataset Description}
\label{app:problems}

This section expands the dataset-level statistics used in the main text. We report \textsc{MathConstraint} first, then \textsc{MathConstraint-Easy}, matching the order of the experiments.

\begin{table}[htbp]
\caption{Extended dataset difficulty statistics. Solver-time columns are reference generation-time wall time in seconds. Structural columns report medians and 90th percentiles from each instance's recorded difficulty metadata; search-space values are $\log_{10}$ of the recorded search-space proxy.}
\label{tab:dataset-difficulty-appendix}
\centering
\small
\setlength{\tabcolsep}{4pt}
\begin{tabular*}{\textwidth}{@{\extracolsep{\fill}}lrrrrrrrrrrr@{}}
\toprule
& & \multicolumn{1}{c}{Backend} & \multicolumn{3}{c}{Solver time (s)} & \multicolumn{2}{c}{Variables} & \multicolumn{2}{c}{Constraints} & \multicolumn{2}{c}{$\log_{10}$ space} \\
\cmidrule(lr){4-6}\cmidrule(lr){7-8}\cmidrule(lr){9-10}\cmidrule(lr){11-12}
Dataset & $n$ & CP/SMS & p50 & p90 & p99 & p50 & p90 & p50 & p90 & p50 & p90 \\
\midrule
\textsc{MathConstraint-Easy} & 266 & 110/156 & 0.20 & 1.79 & 11.8 & 17.5 & 42.1 & 16.5 & 263 & 8.68 & 27.9 \\
\textsc{MathConstraint} & 329 & 227/102 & 2.52 & 75.6 & 653 & 46.0 & 210 & 154 & 949 & 18.7 & 106 \\
\bottomrule
\end{tabular*}
\end{table}

\begin{table}[htbp]
\caption{Problem-type composition of \textsc{MathConstraint}.}
\label{tab:problem-type-splits-mathconstraint}
\centering
\scriptsize
\setlength{\tabcolsep}{3pt}
\begin{tabular*}{\textwidth}{@{\extracolsep{\fill}}lrrrlrrr@{}}
\toprule
Problem type & $n$ & SAT & UNSAT & Problem type & $n$ & SAT & UNSAT \\
\midrule
\texttt{all\_interval} & 5 & 5 & 0 & \texttt{number\_partitioning} & 10 & 10 & 0 \\
\texttt{antimagic\_square} & 4 & 4 & 0 & \texttt{ortholatin} & 2 & 2 & 0 \\
\texttt{bibd} & 9 & 9 & 0 & \texttt{pysms\_chromatic\_girth} & 47 & 20 & 27 \\
\texttt{costas\_array} & 8 & 8 & 0 & \texttt{pysms\_clique\_coloring} & 4 & 4 & 0 \\
\texttt{debruijn} & 1 & 1 & 0 & \texttt{pysms\_combined\_graph} & 1 & 0 & 1 \\
\texttt{golomb} & 5 & 5 & 0 & \texttt{pysms\_contains\_cliques} & 1 & 0 & 1 \\
\texttt{graceful\_graph} & 21 & 21 & 0 & \texttt{pysms\_degree\_bounds} & 2 & 2 & 0 \\
\texttt{graph\_k\_coloring} & 22 & 18 & 4 & \texttt{pysms\_girth\_degree} & 32 & 15 & 17 \\
\texttt{hadamard} & 4 & 4 & 0 & \texttt{pysms\_graph\_builder} & 6 & 2 & 4 \\
\texttt{hamilton\_cycle} & 1 & 0 & 1 & \texttt{pysms\_independent\_connectivity} & 6 & 6 & 0 \\
\texttt{knight\_tour} & 4 & 4 & 0 & \texttt{pysms\_min\_girth} & 1 & 1 & 0 \\
\texttt{langford} & 9 & 9 & 0 & \texttt{pysms\_mtf} & 1 & 1 & 0 \\
\texttt{latin\_square\_completion} & 4 & 4 & 0 & \texttt{pysms\_ramsey} & 1 & 0 & 1 \\
\texttt{low\_autocorrelation} & 14 & 14 & 0 & \texttt{quasigroup\_idempotent} & 5 & 2 & 3 \\
\texttt{magic\_sequence} & 1 & 0 & 1 & \texttt{queens} & 6 & 6 & 0 \\
\texttt{magic\_square} & 2 & 2 & 0 & \texttt{ramsey} & 14 & 13 & 1 \\
\texttt{max\_clique} & 26 & 15 & 11 & \texttt{social\_golfers} & 16 & 16 & 0 \\
\texttt{max\_independent\_set} & 22 & 1 & 21 & \texttt{sudoku} & 2 & 2 & 0 \\
\texttt{non\_transitive\_dice} & 2 & 2 & 0 & \texttt{van\_der\_waerden} & 3 & 0 & 3 \\
 & & & & \texttt{vertex\_cover} & 5 & 0 & 5 \\
\bottomrule
\end{tabular*}
\end{table}

\begin{table}[htbp]
\caption{Problem-type composition of \textsc{MathConstraint-Easy}.}
\label{tab:problem-type-splits-easy}
\centering
\scriptsize
\setlength{\tabcolsep}{3pt}
\begin{tabular*}{\textwidth}{@{\extracolsep{\fill}}lrrrlrrr@{}}
\toprule
Problem type & $n$ & SAT & UNSAT & Problem type & $n$ & SAT & UNSAT \\
\midrule
\texttt{all\_interval} & 12 & 12 & 0 & \texttt{pysms\_degree\_bounds} & 12 & 11 & 1 \\
\texttt{costas\_array} & 9 & 9 & 0 & \texttt{pysms\_girth\_degree} & 12 & 6 & 6 \\
\texttt{golomb} & 6 & 6 & 0 & \texttt{pysms\_graph\_builder} & 12 & 6 & 6 \\
\texttt{graceful\_graph} & 8 & 8 & 0 & \texttt{pysms\_independent\_connectivity} & 12 & 12 & 0 \\
\texttt{knight\_tour} & 4 & 4 & 0 & \texttt{pysms\_min\_connectivity} & 12 & 12 & 0 \\
\texttt{langford} & 8 & 6 & 2 & \texttt{pysms\_min\_degree} & 12 & 12 & 0 \\
\texttt{low\_autocorrelation} & 12 & 12 & 0 & \texttt{pysms\_min\_girth} & 12 & 12 & 0 \\
\texttt{magic\_sequence} & 12 & 10 & 2 & \texttt{pysms\_mtf} & 12 & 12 & 0 \\
\texttt{pigeons} & 12 & 0 & 12 & \texttt{pysms\_num\_edges\_bounds} & 12 & 12 & 0 \\
\texttt{pysms\_chromatic\_girth} & 12 & 9 & 3 & \texttt{queens} & 12 & 12 & 0 \\
\texttt{pysms\_clique\_coloring} & 12 & 12 & 0 & \texttt{ramsey} & 12 & 8 & 4 \\
\texttt{pysms\_combined\_graph} & 12 & 0 & 12 & \texttt{sudoku} & 3 & 3 & 0 \\
\texttt{pysms\_contains\_cliques} & 12 & 0 & 12 & & & & \\
\bottomrule
\end{tabular*}
\end{table}

\begin{table}[htbp]
\caption{No-tools hint stratification on \textsc{MathConstraint-Easy}. Entries are verifier accuracy percentages for hinted SAT, unhinted SAT, and UNSAT instances.}
\label{tab:hint-stratification-easy-no-tools}
\centering
\scriptsize
\setlength{\tabcolsep}{4pt}
\begin{tabular*}{0.82\textwidth}{@{\extracolsep{\fill}}lrrr@{}}
\toprule
Model & Hinted SAT & Unhinted SAT & UNSAT \\
\midrule
\textsc{gpt-5.5} & 76 & 95 & 95 \\
\textsc{claude-opus-4.7} & 75 & 93 & 90 \\
\textsc{claude-4.6-sonnet} & 66 & 80 & 83 \\
\textsc{gemini-3.1-pro} & 75 & 91 & 88 \\
\textsc{gemini-3.1-flash-lite} & 57 & 81 & 85 \\
\textsc{grok-4.20} & 75 & 91 & 92 \\
\midrule
\textsc{deepseek-v4-pro} & 72 & 90 & 88 \\
\textsc{deepseek-v4-flash} & 61 & 79 & 90 \\
\textsc{qwen3.6-plus} & 70 & 86 & 92 \\
\textsc{kimi-k2.6} & 62 & 87 & 85 \\
\midrule
\textsc{gpt-oss-120b} & 53 & 74 & 88 \\
\textsc{llama-3.3-70b-instruct} & 3 & 26 & 52 \\
\bottomrule
\end{tabular*}
\end{table}

\begin{table}[htbp]
\caption{Tool-enabled hint stratification on \textsc{MathConstraint-Easy}. Entries are verifier accuracy percentages for hinted SAT, unhinted SAT, and UNSAT instances.}
\label{tab:hint-stratification-easy-tools}
\centering
\scriptsize
\setlength{\tabcolsep}{4pt}
\begin{tabular*}{0.82\textwidth}{@{\extracolsep{\fill}}lrrr@{}}
\toprule
Model & Hinted SAT & Unhinted SAT & UNSAT \\
\midrule
\textsc{gpt-5.5} & 83 & 98 & 95 \\
\textsc{claude-opus-4.7} & 81 & 99 & 92 \\
\textsc{claude-4.6-sonnet} & 84 & 98 & 85 \\
\textsc{gemini-3.1-pro} & 83 & 100 & 90 \\
\textsc{gemini-3.1-flash-lite} & 82 & 99 & 87 \\
\textsc{grok-4.20} & 74 & 96 & 95 \\
\midrule
\textsc{deepseek-v4-pro} & 58 & 79 & 83 \\
\textsc{deepseek-v4-flash} & 60 & 81 & 75 \\
\textsc{qwen3.6-plus} & 79 & 97 & 90 \\
\textsc{kimi-k2.6} & 72 & 91 & 85 \\
\midrule
\textsc{gpt-oss-120b} & 52 & 80 & 90 \\
\textsc{llama-3.3-70b-instruct} & 5 & 17 & 85 \\
\bottomrule
\end{tabular*}
\end{table}

\FloatBarrier
\section{Models, Configurations, and Pricing}
\label{app:models}

\paragraph{Model configuration.}

All models use temperature $0$, \texttt{max\_tokens}$=16{,}384$, and OpenRouter routing with fallbacks. The tool condition allows up to eight tool-calling rounds. Reasoning controls follow provider-native options through OpenRouter (\texttt{configs/models.yaml}): models exposing a discrete reasoning-effort level run at the maximum (\textsc{gemini-3.1-flash-lite} at medium); Anthropic's API uses a \texttt{budget\_tokens} cap on thinking tokens rather than a discrete effort level, set to $8$k for \textsc{claude-opus-4.7} and unset for \textsc{claude-4.6-sonnet}; remaining models run with no reasoning configuration.

\paragraph{Frontier cohort.}

Throughout this paper, ``frontier'' refers to a fixed cohort of six models used both for the admission filter (Section~\cref{app:method-notes:filter}) and for cohort-level summary statistics: \textsc{GPT-5.5}, \textsc{Claude-Opus-4.7}, \textsc{Claude-4.6-Sonnet}, \textsc{Grok-4.20}, \textsc{Gemini-3.1-Pro}, and \textsc{Gemini-3.1-Flash-Lite}. The remaining six evaluated models (\textsc{DeepSeek-V4-Pro}, \textsc{DeepSeek-V4-Flash}, \textsc{Qwen3.6-Plus}, \textsc{Kimi-K2.6}, \textsc{GPT-OSS-120B}, \textsc{Llama-3.3-70B-Instruct}) are reported alongside the frontier cohort but did not participate in admission. Future refreshes will update this cohort and document the change in the corresponding release notes.

\paragraph{Model slugs.}
The exact model slugs are \texttt{openai/gpt-5.5-20260423}, \texttt{anthropic/claude-opus-4.7}, \texttt{anthropic/claude-4.6-sonnet-20260217}, \texttt{google/gemini-3.1-pro-preview-20260219}, \\\texttt{google/gemini-3.1-flash-lite-preview-20260303}, \texttt{x-ai/grok-4.20-20260309}, \texttt{deepseek/deepseek-v4-pro-20260423}, \texttt{deepseek/deepseek-v4-flash-20260423}, \texttt{qwen/qwen3.6-plus-04-02}, \texttt{moonshotai/kimi-k2.6-20260420}, \texttt{openai/gpt-oss-120b:free}, and \texttt{meta-llama/llama-3.3-70b-instruct}.

\begin{table}[htbp]
\caption{Model configuration and token pricing. Prices are USD per million tokens from \texttt{configs/model\_pricing.yaml}, correct as of 2026-04-30. Dashes indicate models without an explicit paid price entry in the pricing file.}
\label{tab:model-config-pricing}
\centering
\scriptsize
\setlength{\tabcolsep}{4pt}
\begin{tabular*}{\textwidth}{@{\extracolsep{\fill}}l>{\raggedright\arraybackslash}p{0.46\textwidth}rr@{}}
\toprule
Model & Reasoning config & Input & Output \\
\midrule
\textsc{gpt-5.5} & \texttt{effort=high, exclude=False} & 5 & 30 \\
\textsc{claude-opus-4.7} & \texttt{max\_tokens=8000, exclude=False} & 5 & 25 \\
\textsc{claude-4.6-sonnet} & \texttt{none} & 3 & 15 \\
\textsc{gemini-3.1-pro} & \texttt{effort=high, exclude=False} & 2 & 12 \\
\textsc{gemini-3.1-flash-lite} & \texttt{effort=medium, exclude=False} & 0.25 & 1.5 \\
\textsc{grok-4.20} & \texttt{effort=high, exclude=False} & 2 & 6 \\
\midrule
\textsc{deepseek-v4-pro} & \texttt{effort=high, exclude=False} & 0.435 & 0.87 \\
\textsc{deepseek-v4-flash} & \texttt{none} & 0.14 & 0.28 \\
\textsc{qwen3.6-plus} & \texttt{effort=high, exclude=False} & 0.325 & 1.95 \\
\textsc{kimi-k2.6} & \texttt{effort=high, exclude=False} & 0.7448 & 4.655 \\
\midrule
\textsc{gpt-oss-120b} & \texttt{none} & -- & -- \\
\textsc{llama-3.3-70b-instruct} & \texttt{none} & -- & -- \\
\bottomrule
\end{tabular*}
\end{table}

\FloatBarrier
\section{Extended Experimental Results}
\label{app:extended-results}

The following tables are audit material for the main results: efficiency, token composition, tool behavior, force-submit behavior, failure categories, stratification, and per-type accuracy.

\begin{table}[htbp]
\caption{Tool-budget sensitivity on the full 329-instance \textsc{MathConstraint} release. \textsc{sim@}k is a conservative replay metric defined in \cref{app:metrics}; \textsc{sim@}8 is the full tool-enabled setting.}
\label{tab:mathconstraint-tool-budget}
\centering
\small
\renewcommand{\arraystretch}{0.96}
\setlength{\tabcolsep}{3pt}
\begin{tabular}{@{}lcccc@{}}
\toprule
Model & \textsc{sim@}1 & \textsc{sim@}2 & \textsc{sim@}4 & \textsc{sim@}8 \\
\midrule
\textsc{gpt-5.5} & \textbf{19.5} & \textbf{41.0} & \textbf{63.2} & \textbf{80.9} \\
\textsc{claude-opus-4.7} & 3.3 & 30.4 & \underline{59.3} & \underline{77.5} \\
\textsc{claude-4.6-sonnet} & 0.0 & 6.1 & 33.7 & 70.5 \\
\textsc{gemini-3.1-pro} & 0.9 & 18.2 & 50.2 & 72.9 \\
\textsc{gemini-3.1-flash-lite} & 0.6 & 4.9 & 28.0 & 62.3 \\
\textsc{grok-4.20} & 7.9 & 28.0 & 45.3 & 66.6 \\
\midrule
\textsc{deepseek-v4-pro} & 4.3 & 10.3 & 23.1 & 41.3 \\
\textsc{deepseek-v4-flash} & 3.0 & 7.0 & 23.1 & 36.5 \\
\textsc{qwen3.6-plus} & 0.3 & 14.0 & 41.0 & 66.3 \\
\textsc{kimi-k2.6} & 2.7 & 10.9 & 28.9 & 59.9 \\
\midrule
\textsc{gpt-oss-120b} & \underline{13.1} & 23.4 & 31.0 & 36.5 \\
\textsc{llama-3.3-70b-instruct} & 1.5 & \underline{31.9} & 31.9 & 31.9 \\
\bottomrule
\end{tabular}
\renewcommand{\arraystretch}{1}
\end{table}

\begin{table}[htbp]
\caption{No-tools evaluation efficiency on the full 329-instance \textsc{MathConstraint} release. Token and latency columns are per-instance averages from API usage logs; cost is dollars per instance from provider-reported billing.}
\label{tab:eval-efficiency-mathconstraint-no-tools}
\centering
\scriptsize
\setlength{\tabcolsep}{4pt}
\begin{tabular*}{0.82\textwidth}{@{\extracolsep{\fill}}lrrr@{}}
\toprule
Model & Tokens & Lat. (s) & Cost \\
\midrule
\textsc{gpt-5.5} & 10880 & 174.8 & 0.282 \\
\textsc{claude-opus-4.7} & 18170 & 125.6 & 0.425 \\
\textsc{claude-4.6-sonnet} & 5420 & 51.1 & 0.067 \\
\textsc{gemini-3.1-pro} & 13221 & 129.7 & 0.141 \\
\textsc{gemini-3.1-flash-lite} & 3994 & 9.2 & 0.005 \\
\textsc{grok-4.20} & 21899 & 235.9 & 0.068 \\
\midrule
\textsc{deepseek-v4-pro} & 23822 & 778.9 & 0.073 \\
\textsc{deepseek-v4-flash} & 22295 & 275.0 & 0.006 \\
\textsc{qwen3.6-plus} & 52707 & 976.0 & 0.101 \\
\textsc{kimi-k2.6} & 26030 & 936.9 & 0.100 \\
\midrule
\textsc{gpt-oss-120b} & 1353 & 21.5 & 0.000 \\
\textsc{llama-3.3-70b-instruct} & 2365 & 38.0 & 0.001 \\
\bottomrule
\end{tabular*}
\end{table}

\begin{table}[htbp]
\caption{Tool-enabled evaluation efficiency on the full 329-instance \textsc{MathConstraint} release. Token and latency columns are per-instance averages from API usage logs; cost is dollars per instance from provider-reported billing.}
\label{tab:eval-efficiency-mathconstraint-tools}
\centering
\scriptsize
\setlength{\tabcolsep}{4pt}
\begin{tabular*}{0.82\textwidth}{@{\extracolsep{\fill}}lrrr@{}}
\toprule
Model & Tokens & Lat. (s) & Cost \\
\midrule
\textsc{gpt-5.5} & 12081 & 91.0 & 0.151 \\
\textsc{claude-opus-4.7} & 23441 & 68.3 & 0.208 \\
\textsc{claude-4.6-sonnet} & 35046 & 94.0 & 0.174 \\
\textsc{gemini-3.1-pro} & 23181 & 115.0 & 0.132 \\
\textsc{gemini-3.1-flash-lite} & 33977 & 48.8 & 0.015 \\
\textsc{grok-4.20} & 22526 & 131.9 & 0.039 \\
\midrule
\textsc{deepseek-v4-pro} & 32361 & 362.5 & 0.058 \\
\textsc{deepseek-v4-flash} & 21711 & 283.3 & 0.004 \\
\textsc{qwen3.6-plus} & 35036 & 211.3 & 0.027 \\
\textsc{kimi-k2.6} & 37166 & 704.5 & 0.064 \\
\midrule
\textsc{gpt-oss-120b} & 12717 & 78.3 & 0.000 \\
\textsc{llama-3.3-70b-instruct} & 4012 & 27.3 & 0.001 \\
\bottomrule
\end{tabular*}
\end{table}

\begin{table}[htbp]
\caption{No-tools evaluation efficiency on \textsc{MathConstraint-Easy} ($n=266$). Token and latency columns are per-instance averages from API usage logs; cost is dollars per instance from provider-reported billing.}
\label{tab:eval-efficiency-easy-no-tools}
\centering
\scriptsize
\setlength{\tabcolsep}{4pt}
\begin{tabular*}{0.82\textwidth}{@{\extracolsep{\fill}}lrrr@{}}
\toprule
Model & Tokens & Lat. (s) & Cost \\
\midrule
\textsc{gpt-5.5} & 3249 & 48.9 & 0.091 \\
\textsc{claude-opus-4.7} & 4548 & 30.6 & 0.105 \\
\textsc{claude-4.6-sonnet} & 2222 & 23.2 & 0.029 \\
\textsc{gemini-3.1-pro} & 5097 & 51.5 & 0.052 \\
\textsc{gemini-3.1-flash-lite} & 2330 & 7.0 & 0.003 \\
\textsc{grok-4.20} & 8102 & 89.1 & 0.047 \\
\midrule
\textsc{deepseek-v4-pro} & 7620 & 242.5 & 0.025 \\
\textsc{deepseek-v4-flash} & 6693 & 80.3 & 0.002 \\
\textsc{qwen3.6-plus} & 21593 & 421.2 & 0.042 \\
\textsc{kimi-k2.6} & 8593 & 334.0 & 0.034 \\
\midrule
\textsc{gpt-oss-120b} & 955 & 22.5 & 0.000 \\
\textsc{llama-3.3-70b-instruct} & 1580 & 33.0 & 0.001 \\
\bottomrule
\end{tabular*}
\end{table}

\begin{table}[htbp]
\caption{Tool-enabled evaluation efficiency on \textsc{MathConstraint-Easy} ($n=266$). Token and latency columns are per-instance averages from API usage logs; cost is dollars per instance from provider-reported billing.}
\label{tab:eval-efficiency-easy-tools}
\centering
\scriptsize
\setlength{\tabcolsep}{4pt}
\begin{tabular*}{0.82\textwidth}{@{\extracolsep{\fill}}lrrr@{}}
\toprule
Model & Tokens & Lat. (s) & Cost \\
\midrule
\textsc{gpt-5.5} & 2126 & 21.9 & 0.035 \\
\textsc{claude-opus-4.7} & 6249 & 19.7 & 0.058 \\
\textsc{claude-4.6-sonnet} & 13493 & 36.4 & 0.068 \\
\textsc{gemini-3.1-pro} & 6549 & 48.2 & 0.047 \\
\textsc{gemini-3.1-flash-lite} & 8069 & 14.5 & 0.005 \\
\textsc{grok-4.20} & 7815 & 48.3 & 0.027 \\
\midrule
\textsc{deepseek-v4-pro} & 10499 & 314.2 & 0.027 \\
\textsc{deepseek-v4-flash} & 12620 & 174.6 & 0.002 \\
\textsc{qwen3.6-plus} & 9072 & 68.8 & 0.008 \\
\textsc{kimi-k2.6} & 7864 & 225.8 & 0.017 \\
\midrule
\textsc{gpt-oss-120b} & 5022 & 40.5 & 0.000 \\
\textsc{llama-3.3-70b-instruct} & 1853 & 11.3 & 0.000 \\
\bottomrule
\end{tabular*}
\end{table}

\begin{table}[htbp]
\caption{No-tools token breakdown on the full 329-instance \textsc{MathConstraint} release. Entries are per-instance averages; completion tokens are the response-length proxy.}
\label{tab:token-breakdown-mathconstraint-no-tools}
\centering
\scriptsize
\setlength{\tabcolsep}{4pt}
\begin{tabular*}{0.88\textwidth}{@{\extracolsep{\fill}}lrrrr@{}}
\toprule
Model & Prompt & Compl. & Reason. & Cached \\
\midrule
\textsc{gpt-5.5} & 1761 & 9119 & 8849 & 113 \\
\textsc{claude-opus-4.7} & 1438 & 16732 & 2185 & 0 \\
\textsc{claude-4.6-sonnet} & 1161 & 4259 & 0 & 0 \\
\textsc{gemini-3.1-pro} & 1717 & 11504 & 11082 & 0 \\
\textsc{gemini-3.1-flash-lite} & 1029 & 2965 & 1849 & 0 \\
\textsc{grok-4.20} & 909 & 20991 & 18967 & 162 \\
\midrule
\textsc{deepseek-v4-pro} & 1236 & 22586 & 21161 & 226 \\
\textsc{deepseek-v4-flash} & 1100 & 21195 & 16896 & 288 \\
\textsc{qwen3.6-plus} & 924 & 51783 & 48207 & 0 \\
\textsc{kimi-k2.6} & 1288 & 24743 & 17886 & 144 \\
\midrule
\textsc{gpt-oss-120b} & 849 & 504 & 274 & 172 \\
\textsc{llama-3.3-70b-instruct} & 1090 & 1276 & 0 & 18 \\
\bottomrule
\end{tabular*}
\end{table}

\begin{table}[htbp]
\caption{Tool-enabled token breakdown on the full 329-instance \textsc{MathConstraint} release. Entries are per-instance averages; completion tokens are the response-length proxy.}
\label{tab:token-breakdown-mathconstraint-tools}
\centering
\scriptsize
\setlength{\tabcolsep}{4pt}
\begin{tabular*}{0.88\textwidth}{@{\extracolsep{\fill}}lrrrr@{}}
\toprule
Model & Prompt & Compl. & Reason. & Cached \\
\midrule
\textsc{gpt-5.5} & 8051 & 4030 & 2018 & 2281 \\
\textsc{claude-opus-4.7} & 18894 & 4547 & 191 & 0 \\
\textsc{claude-4.6-sonnet} & 29344 & 5702 & 0 & 0 \\
\textsc{gemini-3.1-pro} & 14349 & 8831 & 5272 & 1554 \\
\textsc{gemini-3.1-flash-lite} & 27238 & 6740 & 979 & 7220 \\
\textsc{grok-4.20} & 12629 & 9897 & 6361 & 5959 \\
\midrule
\textsc{deepseek-v4-pro} & 23648 & 8713 & 3490 & 9283 \\
\textsc{deepseek-v4-flash} & 15716 & 5995 & 1850 & 3589 \\
\textsc{qwen3.6-plus} & 25594 & 9442 & 3643 & 0 \\
\textsc{kimi-k2.6} & 24213 & 12953 & 6078 & 8560 \\
\midrule
\textsc{gpt-oss-120b} & 10815 & 1902 & 310 & 7231 \\
\textsc{llama-3.3-70b-instruct} & 3119 & 893 & 0 & 0 \\
\bottomrule
\end{tabular*}
\end{table}

\begin{table}[htbp]
\caption{Tool-use behavior on \textsc{MathConstraint-Easy}. Calls and rounds are per-instance averages; execute and submit are total tool-call counts. Submit and forced are instance rates in percent. Submit is the explicit \texttt{submit\_answer} instance rate; forced is the evaluator \texttt{forced\_submit} flag rate." Also change both column headers from "Forced \%" to "Flagged \%}
\label{tab:tool-use-behavior-easy}
\centering
\scriptsize
\setlength{\tabcolsep}{3pt}
\begin{tabular*}{\textwidth}{@{\extracolsep{\fill}}lrrrrrr@{}}
\toprule
Model & Calls & Rounds & Execute & Submit & Submit \% & Forced \% \\
\midrule
\textsc{gpt-5.5} & 0.73 & 1.41 & 111 & 83 & 31.2 & 2.6 \\
\textsc{claude-opus-4.7} & 2.08 & 2.09 & 295 & 259 & 97.4 & 1.9 \\
\textsc{claude-4.6-sonnet} & 3.57 & 3.77 & 753 & 197 & 74.1 & 25.6 \\
\textsc{gemini-3.1-pro} & 2.68 & 2.68 & 455 & 258 & 97.0 & 2.3 \\
\textsc{gemini-3.1-flash-lite} & 3.13 & 3.17 & 592 & 241 & 90.6 & 7.9 \\
\textsc{grok-4.20} & 2.83 & 2.87 & 534 & 218 & 82.0 & 13.9 \\
\midrule
\textsc{deepseek-v4-pro} & 2.63 & 2.93 & 539 & 161 & 60.5 & 16.9 \\
\textsc{deepseek-v4-flash} & 3.35 & 3.53 & 693 & 199 & 74.8 & 6.0 \\
\textsc{qwen3.6-plus} & 2.79 & 2.85 & 500 & 243 & 91.4 & 5.6 \\
\textsc{kimi-k2.6} & 2.73 & 2.74 & 476 & 248 & 93.2 & 0.8 \\
\midrule
\textsc{gpt-oss-120b} & 2.51 & 3.35 & 667 & 1 & 0.4 & 18.4 \\
\textsc{llama-3.3-70b-instruct} & 1.89 & 1.89 & 236 & 266 & 100.0 & 0.0 \\
\bottomrule
\end{tabular*}
\end{table}

\begin{table}[htbp]
\caption{Tool-use behavior on the full 329-instance \textsc{MathConstraint} release. Calls and rounds are per-instance averages; execute and submit are total tool-call counts. Submit and forced are instance rates in percent. Submit is the explicit \texttt{submit\_answer} instance rate; forced is the evaluator \texttt{forced\_submit} flag rate." Also change both column headers from "Forced \%" to "Flagged \%}
\label{tab:tool-use-behavior-mathconstraint}
\centering
\scriptsize
\setlength{\tabcolsep}{3pt}
\begin{tabular*}{\textwidth}{@{\extracolsep{\fill}}lrrrrrr@{}}
\toprule
Model & Calls & Rounds & Execute & Submit & Submit \% & Forced \% \\
\midrule
\textsc{gpt-5.5} & 2.87 & 3.07 & 703 & 241 & 73.3 & 4.0 \\
\textsc{claude-opus-4.7} & 3.65 & 3.65 & 915 & 287 & 87.2 & 11.2 \\
\textsc{claude-4.6-sonnet} & 5.34 & 5.39 & 1543 & 215 & 65.3 & 34.0 \\
\textsc{gemini-3.1-pro} & 4.25 & 4.26 & 1130 & 267 & 81.2 & 11.2 \\
\textsc{gemini-3.1-flash-lite} & 5.31 & 5.34 & 1525 & 222 & 67.5 & 31.9 \\
\textsc{grok-4.20} & 4.50 & 4.53 & 1270 & 209 & 63.5 & 29.8 \\
\midrule
\textsc{deepseek-v4-pro} & 4.49 & 4.72 & 1303 & 173 & 52.6 & 11.6 \\
\textsc{deepseek-v4-flash} & 3.95 & 4.33 & 1150 & 149 & 45.3 & 9.7 \\
\textsc{qwen3.6-plus} & 4.56 & 4.59 & 1255 & 244 & 74.2 & 13.1 \\
\textsc{kimi-k2.6} & 5.20 & 5.22 & 1496 & 213 & 64.7 & 14.3 \\
\midrule
\textsc{gpt-oss-120b} & 3.96 & 4.62 & 1300 & 4 & 1.2 & 35.9 \\
\textsc{llama-3.3-70b-instruct} & 1.84 & 1.84 & 278 & 329 & 100.0 & 0.0 \\
\bottomrule
\end{tabular*}
\end{table}

\begin{table}[htbp]
\caption{Tool-call composition on \textsc{MathConstraint} \textsc{tools}. Fractions of \texttt{execute\_python} calls categorized by primary library import or use; classification follows priority order \texttt{pysat} $\to$ \texttt{z3} $\to$ \texttt{pycosat} $\to$ \texttt{networkx} $\to$ brute-force (loops without solver imports) $\to$ other. \textsc{Calls} is the total number of \texttt{execute\_python} invocations in MC traces.}
\label{tab:tool-call-composition}
\centering
\small
\setlength{\tabcolsep}{4pt}
\begin{tabular}{@{}lr@{\hskip 8pt}cccccc@{}}
\toprule
Model & Calls & \texttt{pysat} & \texttt{z3} & \texttt{pycosat} & \texttt{networkx} & brute-force & other \\
\midrule
\textsc{gpt-5.5} & 703 & 21 & 20 & 0 & 1 & 47 & 10 \\
\textsc{claude-opus-4.7} & 915 & 26 & 8 & 0 & 3 & 54 & 8 \\
\textsc{claude-4.6-sonnet} & 1542 & 22 & 8 & 0 & 3 & 60 & 7 \\
\textsc{gemini-3.1-pro} & 1130 & 32 & 26 & 0 & 8 & 26 & 8 \\
\textsc{gemini-3.1-flash-lite} & 1525 & 7 & 13 & 0 & 11 & 61 & 9 \\
\textsc{grok-4.20} & 1270 & 4 & 44 & 0 & 8 & 39 & 5 \\
\textsc{deepseek-v4-pro} & 1297 & 35 & 24 & 1 & 2 & 36 & 3 \\
\textsc{deepseek-v4-flash} & 1143 & 19 & 24 & 1 & 1 & 49 & 6 \\
\textsc{qwen3.6-plus} & 1252 & 29 & 18 & 0 & 1 & 50 & 2 \\
\textsc{kimi-k2.6} & 1489 & 21 & 21 & 2 & 1 & 49 & 7 \\
\textsc{gpt-oss-120b} & 106 & 1 & 18 & 0 & 4 & 55 & 23 \\
\textsc{llama-3.3-70b-instruct} & 275 & 1 & 0 & 0 & 50 & 48 & 1 \\
\bottomrule
\end{tabular}
\end{table}

\begin{table}[htbp]
\caption{Force-submit fallback analysis for tool-enabled runs on the full 329-instance \textsc{MathConstraint} release. Explicit submissions, force submissions, and rescue rate are defined in \cref{app:metrics}. If every force-submitted instance is scored incorrect, the top-three model set is unchanged, though its order becomes \textsc{claude-opus-4.7}, \textsc{gemini-3.1-pro}, \textsc{gpt-5.5}.}
\label{tab:force-submit-fallback}
\centering
\small
\setlength{\tabcolsep}{4pt}
\begin{tabular*}{\textwidth}{@{\extracolsep{\fill}}lrrrrr@{}}
\toprule
Model & Total & Explicit & Forced & Forced correct & Rescue (\%) \\
\midrule
\textsc{gpt-5.5} & 329 & 241 & 88 & 67 & 76.1 \\
\textsc{claude-opus-4.7} & 329 & 287 & 42 & 24 & 57.1 \\
\textsc{claude-4.6-sonnet} & 329 & 215 & 114 & 56 & 49.1 \\
\textsc{gemini-3.1-pro} & 329 & 267 & 62 & 23 & 37.1 \\
\textsc{gemini-3.1-flash-lite} & 329 & 222 & 107 & 44 & 41.1 \\
\textsc{grok-4.20} & 329 & 209 & 120 & 58 & 48.3 \\
\midrule
\textsc{deepseek-v4-pro} & 329 & 173 & 156 & 32 & 20.5 \\
\textsc{deepseek-v4-flash} & 329 & 149 & 180 & 19 & 10.6 \\
\textsc{qwen3.6-plus} & 329 & 244 & 85 & 32 & 37.6 \\
\textsc{kimi-k2.6} & 329 & 213 & 116 & 41 & 35.3 \\
\midrule
\textsc{gpt-oss-120b} & 329 & 4 & 325 & 118 & 36.3 \\
\textsc{llama-3.3-70b-instruct} & 329 & 329 & 0 & 0 & 0.0 \\
\bottomrule
\end{tabular*}
\end{table}

\begin{table}[htbp]
\caption{No-tools failure buckets on the full 329-instance \textsc{MathConstraint} release. Accepted counts are verifier-accepted answers; other columns partition non-accepted runs by the evaluator's failure bucket. Other includes API/content-filter buckets not shown separately.}
\label{tab:failure-buckets-no-tools}
\centering
\scriptsize
\setlength{\tabcolsep}{3pt}
\begin{tabular*}{\textwidth}{@{\extracolsep{\fill}}lrrrrrrr@{}}
\toprule
Model & Accept & \makecell{Wrong\\pol.} & \makecell{Wrong\\sol.} & Length & Parse & \makecell{Max\\rounds} & Other \\
\midrule
\textsc{gpt-5.5} & 220 & 10 & 40 & 55 & 0 & 0 & 4 \\
\textsc{claude-opus-4.7} & 140 & 12 & 47 & 128 & 0 & 0 & 2 \\
\textsc{claude-4.6-sonnet} & 61 & 75 & 175 & 16 & 2 & 0 & 0 \\
\textsc{gemini-3.1-pro} & 170 & 18 & 51 & 31 & 56 & 0 & 3 \\
\textsc{gemini-3.1-flash-lite} & 82 & 73 & 169 & 5 & 0 & 0 & 0 \\
\textsc{grok-4.20} & 191 & 101 & 32 & 1 & 0 & 0 & 4 \\
\midrule
\textsc{deepseek-v4-pro} & 130 & 20 & 32 & 135 & 2 & 0 & 10 \\
\textsc{deepseek-v4-flash} & 112 & 49 & 90 & 68 & 6 & 0 & 4 \\
\textsc{qwen3.6-plus} & 156 & 36 & 82 & 1 & 21 & 0 & 33 \\
\textsc{kimi-k2.6} & 118 & 14 & 23 & 163 & 0 & 0 & 11 \\
\midrule
\textsc{gpt-oss-120b} & 130 & 172 & 27 & 0 & 0 & 0 & 0 \\
\textsc{llama-3.3-70b-instruct} & 28 & 78 & 179 & 4 & 39 & 0 & 1 \\
\bottomrule
\end{tabular*}
\end{table}

\begin{table}[htbp]
\caption{Tool-enabled failure buckets on the full 329-instance \textsc{MathConstraint} release. Accepted counts are verifier-accepted answers; other columns partition non-accepted runs by the evaluator's failure bucket. Other includes API/content-filter buckets not shown separately.}
\label{tab:failure-buckets-tools}
\centering
\scriptsize
\setlength{\tabcolsep}{3pt}
\begin{tabular*}{\textwidth}{@{\extracolsep{\fill}}lrrrrrrr@{}}
\toprule
Model & Accept & \makecell{Wrong\\pol.} & \makecell{Wrong\\sol.} & Length & Parse & \makecell{Max\\rounds} & Other \\
\midrule
\textsc{gpt-5.5} & 266 & 7 & 44 & 10 & 0 & 0 & 2 \\
\textsc{claude-opus-4.7} & 255 & 25 & 44 & 4 & 0 & 0 & 1 \\
\textsc{claude-4.6-sonnet} & 232 & 25 & 70 & 1 & 1 & 0 & 0 \\
\textsc{gemini-3.1-pro} & 240 & 20 & 44 & 3 & 0 & 22 & 0 \\
\textsc{gemini-3.1-flash-lite} & 205 & 39 & 83 & 1 & 0 & 1 & 0 \\
\textsc{grok-4.20} & 219 & 40 & 55 & 0 & 0 & 14 & 1 \\
\midrule
\textsc{deepseek-v4-pro} & 136 & 43 & 32 & 11 & 27 & 57 & 23 \\
\textsc{deepseek-v4-flash} & 120 & 24 & 37 & 9 & 36 & 44 & 59 \\
\textsc{qwen3.6-plus} & 218 & 13 & 56 & 0 & 3 & 36 & 3 \\
\textsc{kimi-k2.6} & 197 & 25 & 38 & 41 & 6 & 17 & 5 \\
\midrule
\textsc{gpt-oss-120b} & 120 & 130 & 70 & 0 & 0 & 9 & 0 \\
\textsc{llama-3.3-70b-instruct} & 105 & 171 & 53 & 0 & 0 & 0 & 0 \\
\bottomrule
\end{tabular*}
\end{table}

\begin{table}[htbp]
\caption{Witness gap on \textsc{MathConstraint} no-tools. \textsc{Gap} is SAT polarity accuracy minus end-to-end verifier accuracy. It measures how often a model decides existence correctly but fails to construct a witness that survives exact verification (\cref{app:metrics}). Sorted by gap descending.}
\label{tab:witness-gap}
\centering
\small
\setlength{\tabcolsep}{4pt}
\begin{tabular}{@{}lccc@{}}
\toprule
Model & SAT Acc.\ (\%) & Accuracy (\%) & Gap (pp) \\
\midrule
\textsc{llama-3.3-70b-instruct} & 62.9 & 8.5 & 54.4 \\
\textsc{claude-4.6-sonnet} & 72.0 & 18.5 & 53.5 \\
\textsc{gemini-3.1-flash-lite} & 76.3 & 24.9 & 51.4 \\
\textsc{deepseek-v4-flash} & 61.4 & 34.0 & 27.4 \\
\textsc{qwen3.6-plus} & 72.3 & 47.4 & 24.9 \\
\textsc{gemini-3.1-pro} & 67.2 & 51.7 & 15.5 \\
\textsc{claude-opus-4.7} & 56.8 & 42.6 & 14.3 \\
\textsc{gpt-5.5} & 79.0 & 66.9 & 12.2 \\
\textsc{grok-4.20} & 67.8 & 58.1 & 9.7 \\
\textsc{deepseek-v4-pro} & 49.2 & 39.5 & 9.7 \\
\textsc{gpt-oss-120b} & 47.7 & 39.5 & 8.2 \\
\textsc{kimi-k2.6} & 42.9 & 35.9 & 7.0 \\
\bottomrule
\end{tabular}
\end{table}

\begin{table}[htbp]
\caption{No-tools polarity and backend stratification on \textsc{MathConstraint-Easy}. Entries are verifier accuracy percentages. CP denotes \texttt{pycsp}; SMS denotes \texttt{pysms}.}
\label{tab:stratification-easy-no-tools}
\centering
\scriptsize
\setlength{\tabcolsep}{4pt}
\begin{tabular*}{0.82\textwidth}{@{\extracolsep{\fill}}lrrrr@{}}
\toprule
Model & SAT & UNSAT & CP & SMS \\
\midrule
\textsc{gpt-5.5} & 85 & 95 & 75 & 97 \\
\textsc{claude-opus-4.7} & 84 & 90 & 71 & 96 \\
\textsc{claude-4.6-sonnet} & 73 & 83 & 51 & 92 \\
\textsc{gemini-3.1-pro} & 83 & 88 & 68 & 96 \\
\textsc{gemini-3.1-flash-lite} & 69 & 85 & 45 & 92 \\
\textsc{grok-4.20} & 83 & 92 & 69 & 96 \\
\midrule
\textsc{deepseek-v4-pro} & 81 & 88 & 65 & 95 \\
\textsc{deepseek-v4-flash} & 70 & 90 & 58 & 86 \\
\textsc{qwen3.6-plus} & 78 & 92 & 65 & 92 \\
\textsc{kimi-k2.6} & 75 & 85 & 63 & 87 \\
\midrule
\textsc{gpt-oss-120b} & 64 & 88 & 43 & 88 \\
\textsc{llama-3.3-70b-instruct} & 15 & 52 & 22 & 24 \\
\bottomrule
\end{tabular*}
\end{table}

\begin{table}[htbp]
\caption{Tool-enabled polarity and backend stratification on \textsc{MathConstraint-Easy}. Entries are verifier accuracy percentages. CP denotes \texttt{pycsp}; SMS denotes \texttt{pysms}.}
\label{tab:stratification-easy-tools}
\centering
\scriptsize
\setlength{\tabcolsep}{4pt}
\begin{tabular*}{0.82\textwidth}{@{\extracolsep{\fill}}lrrrr@{}}
\toprule
Model & SAT & UNSAT & CP & SMS \\
\midrule
\textsc{gpt-5.5} & 91 & 95 & 84 & 97 \\
\textsc{claude-opus-4.7} & 90 & 92 & 84 & 95 \\
\textsc{claude-4.6-sonnet} & 91 & 85 & 84 & 94 \\
\textsc{gemini-3.1-pro} & 92 & 90 & 86 & 95 \\
\textsc{gemini-3.1-flash-lite} & 90 & 87 & 83 & 94 \\
\textsc{grok-4.20} & 85 & 95 & 80 & 92 \\
\midrule
\textsc{deepseek-v4-pro} & 68 & 83 & 65 & 77 \\
\textsc{deepseek-v4-flash} & 70 & 75 & 76 & 68 \\
\textsc{qwen3.6-plus} & 88 & 90 & 82 & 93 \\
\textsc{kimi-k2.6} & 82 & 85 & 77 & 86 \\
\midrule
\textsc{gpt-oss-120b} & 66 & 90 & 47 & 88 \\
\textsc{llama-3.3-70b-instruct} & 11 & 85 & 34 & 24 \\
\bottomrule
\end{tabular*}
\end{table}

\begin{table}[htbp]
\caption{No-tools polarity and backend stratification on the full 329-instance \textsc{MathConstraint} release. Entries are verifier accuracy percentages. CP denotes \texttt{pycsp}; SMS denotes \texttt{pysms}.}
\label{tab:stratification-mathconstraint-no-tools}
\centering
\scriptsize
\setlength{\tabcolsep}{4pt}
\begin{tabular*}{0.82\textwidth}{@{\extracolsep{\fill}}lrrrr@{}}
\toprule
Model & SAT & UNSAT & CP & SMS \\
\midrule
\textsc{gpt-5.5} & 62 & 77 & 56 & 91 \\
\textsc{claude-opus-4.7} & 45 & 38 & 33 & 64 \\
\textsc{claude-4.6-sonnet} & 19 & 17 & 12 & 32 \\
\textsc{gemini-3.1-pro} & 46 & 64 & 40 & 77 \\
\textsc{gemini-3.1-flash-lite} & 18 & 41 & 11 & 56 \\
\textsc{grok-4.20} & 43 & 93 & 49 & 78 \\
\midrule
\textsc{deepseek-v4-pro} & 40 & 38 & 30 & 60 \\
\textsc{deepseek-v4-flash} & 29 & 46 & 21 & 64 \\
\textsc{qwen3.6-plus} & 43 & 56 & 35 & 75 \\
\textsc{kimi-k2.6} & 36 & 36 & 30 & 50 \\
\midrule
\textsc{gpt-oss-120b} & 15 & 95 & 28 & 66 \\
\textsc{llama-3.3-70b-instruct} & 1 & 25 & 3 & 21 \\
\bottomrule
\end{tabular*}
\end{table}

\begin{table}[htbp]
\caption{Tool-enabled polarity and backend stratification on the full 329-instance \textsc{MathConstraint} release. Entries are verifier accuracy percentages. CP denotes \texttt{pycsp}; SMS denotes \texttt{pysms}.}
\label{tab:stratification-mathconstraint-tools}
\centering
\scriptsize
\setlength{\tabcolsep}{4pt}
\begin{tabular*}{0.82\textwidth}{@{\extracolsep{\fill}}lrrrr@{}}
\toprule
Model & SAT & UNSAT & CP & SMS \\
\midrule
\textsc{gpt-5.5} & 75 & 95 & 75 & 94 \\
\textsc{claude-opus-4.7} & 73 & 87 & 74 & 84 \\
\textsc{claude-4.6-sonnet} & 66 & 80 & 72 & 68 \\
\textsc{gemini-3.1-pro} & 70 & 79 & 73 & 74 \\
\textsc{gemini-3.1-flash-lite} & 54 & 82 & 60 & 68 \\
\textsc{grok-4.20} & 56 & 91 & 62 & 77 \\
\midrule
\textsc{deepseek-v4-pro} & 41 & 42 & 41 & 42 \\
\textsc{deepseek-v4-flash} & 42 & 25 & 38 & 32 \\
\textsc{qwen3.6-plus} & 61 & 77 & 68 & 63 \\
\textsc{kimi-k2.6} & 57 & 67 & 63 & 54 \\
\midrule
\textsc{gpt-oss-120b} & 21 & 71 & 22 & 70 \\
\textsc{llama-3.3-70b-instruct} & 4 & 96 & 25 & 48 \\
\bottomrule
\end{tabular*}
\end{table}

\begin{sidewaystable}[p]
\caption{Per-type no-tools performance on \textsc{MathConstraint}. Each entry is Accuracy/SAT Acc. in percent.}
\label{tab:per-type-mathconstraint-no-tools}
\centering
\scriptsize
\setlength{\tabcolsep}{2.5pt}
\resizebox{\textheight}{!}{%
\begin{tabular}{@{}lrrrrrrrrrrrr@{}}
\toprule
Problem type & \textsc{gpt-5.5} & \textsc{claude-opus-4.7} & \textsc{claude-4.6-sonnet} & \textsc{gemini-3.1-pro} & \textsc{gemini-3.1-flash-lite} & \textsc{grok-4.20} & \textsc{deepseek-v4-pro} & \textsc{deepseek-v4-flash} & \textsc{qwen3.6-plus} & \textsc{kimi-k2.6} & \textsc{gpt-oss-120b} & \textsc{llama-3.3-70b-instruct} \\
\midrule
\texttt{all\_interval} & 20/100 & 20/100 & 20/80 & 20/100 & 20/80 & 20/60 & 20/80 & 20/80 & 20/100 & 20/80 & 80/80 & 0/100 \\
\texttt{antimagic\_square} & 50/50 & 0/0 & 0/100 & 0/0 & 0/75 & 0/0 & 0/25 & 0/0 & 0/75 & 0/0 & 0/0 & 0/100 \\
\texttt{bibd} & 56/100 & 0/78 & 0/89 & 11/11 & 0/78 & 0/44 & 33/56 & 0/11 & 22/33 & 11/11 & 0/0 & 0/67 \\
\texttt{costas\_array} & 88/88 & 88/88 & 25/100 & 88/100 & 12/100 & 62/75 & 88/88 & 38/75 & 88/100 & 88/88 & 25/75 & 0/100 \\
\texttt{debruijn} & 100/100 & 0/0 & 0/100 & 0/0 & 0/0 & 0/0 & 0/0 & 0/0 & 0/0 & 0/0 & 0/0 & 0/100 \\
\texttt{golomb} & 60/100 & 80/100 & 20/100 & 100/100 & 60/80 & 60/100 & 100/100 & 100/100 & 80/100 & 100/100 & 20/60 & 0/80 \\
\texttt{graceful\_graph} & 24/38 & 14/19 & 0/90 & 10/38 & 5/71 & 14/19 & 5/19 & 10/52 & 10/67 & 10/14 & 0/0 & 0/100 \\
\texttt{graph\_k\_coloring} & 41/91 & 23/64 & 5/82 & 36/95 & 0/82 & 36/59 & 36/64 & 27/82 & 32/82 & 41/59 & 18/18 & 5/77 \\
\texttt{hadamard} & 75/75 & 75/100 & 25/100 & 75/75 & 0/100 & 75/75 & 75/75 & 50/100 & 75/75 & 75/75 & 0/25 & 0/100 \\
\texttt{hamilton\_cycle} & 100/100 & 100/100 & 100/100 & 100/100 & 100/100 & 100/100 & 100/100 & 100/100 & 100/100 & 100/100 & 100/100 & 0/0 \\
\texttt{knight\_tour} & 50/50 & 50/50 & 0/100 & 50/50 & 0/100 & 50/50 & 0/0 & 0/0 & 50/75 & 25/25 & 0/0 & 0/75 \\
\texttt{langford} & 100/100 & 44/44 & 0/100 & 22/56 & 0/100 & 44/78 & 44/56 & 0/67 & 44/78 & 22/33 & 0/0 & 0/100 \\
\texttt{latin\_square\_completion} & 75/75 & 0/0 & 0/100 & 25/25 & 0/100 & 25/50 & 0/0 & 0/75 & 25/75 & 0/0 & 0/0 & 0/75 \\
\texttt{low\_autocorrelation} & 7/14 & 7/14 & 0/93 & 0/50 & 0/79 & 7/7 & 7/21 & 0/36 & 0/36 & 0/7 & 0/14 & 0/93 \\
\texttt{magic\_sequence} & 100/100 & 100/100 & 100/100 & 0/0 & 0/0 & 100/100 & 100/100 & 100/100 & 100/100 & 0/0 & 100/100 & 0/0 \\
\texttt{magic\_square} & 100/100 & 100/100 & 100/100 & 100/100 & 100/100 & 100/100 & 100/100 & 100/100 & 50/50 & 100/100 & 50/100 & 50/100 \\
\texttt{max\_clique} & 73/73 & 54/58 & 38/58 & 65/65 & 19/58 & 81/81 & 27/27 & 27/54 & 50/58 & 35/42 & 46/46 & 4/62 \\
\texttt{max\_independent\_set} & 55/55 & 5/5 & 0/0 & 41/45 & 0/5 & 95/95 & 0/0 & 0/5 & 14/18 & 9/9 & 95/95 & 5/9 \\
\texttt{non\_transitive\_dice} & 100/100 & 100/100 & 0/100 & 100/100 & 50/100 & 100/100 & 100/100 & 0/50 & 100/100 & 100/100 & 0/0 & 0/100 \\
\texttt{number\_partitioning} & 40/100 & 30/100 & 20/100 & 50/100 & 20/100 & 50/100 & 60/100 & 50/100 & 40/100 & 40/90 & 30/100 & 0/100 \\
\texttt{ortholatin} & 100/100 & 100/100 & 50/100 & 100/100 & 50/100 & 100/100 & 100/100 & 50/100 & 100/100 & 100/100 & 0/100 & 0/100 \\
\texttt{pysms\_chromatic\_girth} & 94/94 & 55/60 & 23/49 & 74/74 & 60/83 & 79/83 & 43/45 & 60/70 & 74/89 & 28/28 & 72/77 & 19/36 \\
\texttt{pysms\_clique\_coloring} & 100/100 & 100/100 & 100/100 & 100/100 & 50/100 & 100/100 & 100/100 & 75/75 & 100/100 & 100/100 & 50/50 & 0/25 \\
\texttt{pysms\_combined\_graph} & 100/100 & 100/100 & 0/0 & 100/100 & 100/100 & 100/100 & 100/100 & 100/100 & 100/100 & 100/100 & 100/100 & 100/100 \\
\texttt{pysms\_contains\_cliques} & 100/100 & 100/100 & 0/0 & 100/100 & 100/100 & 100/100 & 100/100 & 0/0 & 100/100 & 100/100 & 0/0 & 0/0 \\
\texttt{pysms\_degree\_bounds} & 100/100 & 50/100 & 50/50 & 100/100 & 100/100 & 100/100 & 100/100 & 50/100 & 100/100 & 100/100 & 50/50 & 0/0 \\
\texttt{pysms\_girth\_degree} & 94/94 & 66/66 & 25/62 & 81/88 & 47/84 & 75/78 & 69/72 & 69/72 & 69/88 & 62/66 & 59/62 & 28/50 \\
\texttt{pysms\_graph\_builder} & 50/50 & 33/33 & 33/33 & 33/33 & 17/33 & 33/33 & 33/33 & 50/50 & 33/33 & 33/33 & 33/33 & 17/33 \\
\texttt{pysms\_independent\_connectivity} & 100/100 & 100/100 & 83/100 & 100/100 & 67/100 & 100/100 & 100/100 & 83/83 & 100/100 & 83/83 & 100/100 & 0/0 \\
\texttt{pysms\_min\_girth} & 100/100 & 100/100 & 0/100 & 0/0 & 100/100 & 100/100 & 100/100 & 0/100 & 100/100 & 100/100 & 0/0 & 0/0 \\
\texttt{pysms\_mtf} & 0/0 & 100/100 & 100/100 & 100/100 & 100/100 & 100/100 & 100/100 & 100/100 & 100/100 & 100/100 & 100/100 & 0/0 \\
\texttt{pysms\_ramsey} & 100/100 & 100/100 & 100/100 & 100/100 & 100/100 & 100/100 & 100/100 & 100/100 & 100/100 & 100/100 & 100/100 & 100/100 \\
\texttt{quasigroup\_idempotent} & 60/60 & 40/40 & 0/40 & 60/60 & 20/60 & 60/60 & 60/60 & 20/40 & 20/60 & 20/20 & 60/60 & 0/40 \\
\texttt{queens} & 67/67 & 50/50 & 0/100 & 33/50 & 0/100 & 17/17 & 33/50 & 33/83 & 17/67 & 33/33 & 0/17 & 0/100 \\
\texttt{ramsey} & 71/79 & 29/57 & 7/100 & 43/50 & 14/93 & 43/50 & 29/36 & 21/57 & 29/50 & 21/43 & 14/29 & 0/64 \\
\texttt{social\_golfers} & 50/100 & 25/75 & 0/100 & 25/69 & 6/94 & 44/69 & 6/44 & 0/81 & 50/94 & 12/25 & 0/12 & 0/100 \\
\texttt{sudoku} & 50/50 & 0/0 & 0/100 & 0/0 & 0/100 & 0/0 & 0/0 & 0/50 & 0/50 & 0/0 & 0/0 & 0/50 \\
\texttt{van\_der\_waerden} & 100/100 & 100/100 & 100/100 & 100/100 & 100/100 & 100/100 & 100/100 & 100/100 & 100/100 & 100/100 & 100/100 & 100/100 \\
\texttt{vertex\_cover} & 80/80 & 60/60 & 20/20 & 60/60 & 0/0 & 100/100 & 40/40 & 40/40 & 60/60 & 60/60 & 100/100 & 0/0 \\
\bottomrule
\end{tabular}}
\end{sidewaystable}

\begin{sidewaystable}[p]
\caption{Per-type tool-enabled performance on \textsc{MathConstraint}. Each entry is Accuracy/SAT Acc. in percent.}
\label{tab:per-type-mathconstraint-tools}
\centering
\scriptsize
\setlength{\tabcolsep}{2.5pt}
\resizebox{\textheight}{!}{%
\begin{tabular}{@{}lrrrrrrrrrrrr@{}}
\toprule
Problem type & \textsc{gpt-5.5} & \textsc{claude-opus-4.7} & \textsc{claude-4.6-sonnet} & \textsc{gemini-3.1-pro} & \textsc{gemini-3.1-flash-lite} & \textsc{grok-4.20} & \textsc{deepseek-v4-pro} & \textsc{deepseek-v4-flash} & \textsc{qwen3.6-plus} & \textsc{kimi-k2.6} & \textsc{gpt-oss-120b} & \textsc{llama-3.3-70b-instruct} \\
\midrule
\texttt{all\_interval} & 20/100 & 20/100 & 20/100 & 20/100 & 20/80 & 20/80 & 0/80 & 0/40 & 20/80 & 20/80 & 20/60 & 0/40 \\
\texttt{antimagic\_square} & 75/75 & 75/100 & 75/100 & 75/75 & 75/75 & 0/0 & 25/25 & 0/0 & 75/75 & 0/0 & 0/0 & 0/0 \\
\texttt{bibd} & 67/100 & 67/100 & 44/89 & 56/100 & 22/89 & 56/89 & 33/67 & 33/44 & 33/67 & 22/33 & 0/22 & 0/22 \\
\texttt{costas\_array} & 100/100 & 100/100 & 100/100 & 100/100 & 88/100 & 75/88 & 62/62 & 88/88 & 88/88 & 100/100 & 50/100 & 12/75 \\
\texttt{debruijn} & 100/100 & 100/100 & 100/100 & 100/100 & 0/100 & 100/100 & 0/0 & 0/100 & 100/100 & 100/100 & 0/0 & 0/0 \\
\texttt{golomb} & 60/100 & 100/100 & 100/100 & 100/100 & 100/100 & 100/100 & 80/80 & 80/80 & 100/100 & 80/80 & 20/100 & 20/60 \\
\texttt{graceful\_graph} & 57/67 & 57/76 & 43/100 & 43/57 & 19/67 & 24/38 & 19/29 & 14/33 & 43/67 & 38/52 & 5/19 & 0/57 \\
\texttt{graph\_k\_coloring} & 32/100 & 27/100 & 23/100 & 27/100 & 23/100 & 27/100 & 14/50 & 32/82 & 27/100 & 23/100 & 32/82 & 18/18 \\
\texttt{hadamard} & 100/100 & 100/100 & 100/100 & 75/100 & 50/75 & 75/75 & 25/25 & 50/50 & 75/100 & 75/75 & 0/25 & 25/25 \\
\texttt{hamilton\_cycle} & 100/100 & 100/100 & 100/100 & 100/100 & 100/100 & 100/100 & 100/100 & 100/100 & 100/100 & 100/100 & 100/100 & 100/100 \\
\texttt{knight\_tour} & 100/100 & 75/75 & 75/100 & 100/100 & 75/75 & 75/75 & 75/75 & 100/100 & 100/100 & 100/100 & 0/50 & 0/25 \\
\texttt{langford} & 89/89 & 100/100 & 100/100 & 100/100 & 89/100 & 89/100 & 11/33 & 78/78 & 78/78 & 89/89 & 44/67 & 0/33 \\
\texttt{latin\_square\_completion} & 100/100 & 100/100 & 100/100 & 100/100 & 50/75 & 100/100 & 75/75 & 100/100 & 75/75 & 100/100 & 25/50 & 25/25 \\
\texttt{low\_autocorrelation} & 43/71 & 43/57 & 43/64 & 36/50 & 43/64 & 21/43 & 36/36 & 21/36 & 43/71 & 43/57 & 0/50 & 21/43 \\
\texttt{magic\_sequence} & 100/100 & 100/100 & 100/100 & 100/100 & 100/100 & 100/100 & 100/100 & 100/100 & 100/100 & 100/100 & 100/100 & 100/100 \\
\texttt{magic\_square} & 100/100 & 50/50 & 100/100 & 100/100 & 100/100 & 100/100 & 100/100 & 100/100 & 100/100 & 100/100 & 50/100 & 0/50 \\
\texttt{max\_clique} & 100/100 & 100/100 & 100/100 & 92/96 & 100/100 & 100/100 & 65/65 & 23/31 & 100/100 & 88/88 & 35/54 & 42/42 \\
\texttt{max\_independent\_set} & 100/100 & 100/100 & 100/100 & 100/100 & 100/100 & 100/100 & 36/36 & 14/14 & 100/100 & 95/95 & 5/9 & 95/95 \\
\texttt{non\_transitive\_dice} & 100/100 & 50/50 & 100/100 & 100/100 & 100/100 & 100/100 & 50/50 & 100/100 & 100/100 & 50/50 & 0/0 & 0/0 \\
\texttt{number\_partitioning} & 40/100 & 30/100 & 50/100 & 0/100 & 0/100 & 0/100 & 0/80 & 0/40 & 60/100 & 50/100 & 40/80 & 0/90 \\
\texttt{ortholatin} & 100/100 & 100/100 & 100/100 & 100/100 & 100/100 & 100/100 & 100/100 & 100/100 & 100/100 & 100/100 & 0/100 & 50/100 \\
\texttt{pysms\_chromatic\_girth} & 94/96 & 89/89 & 81/87 & 60/62 & 64/79 & 79/85 & 28/28 & 21/23 & 64/64 & 38/43 & 74/79 & 53/53 \\
\texttt{pysms\_clique\_coloring} & 100/100 & 100/100 & 100/100 & 100/100 & 100/100 & 100/100 & 100/100 & 75/75 & 100/100 & 100/100 & 50/75 & 0/0 \\
\texttt{pysms\_combined\_graph} & 100/100 & 100/100 & 100/100 & 100/100 & 100/100 & 100/100 & 0/0 & 100/100 & 100/100 & 100/100 & 100/100 & 100/100 \\
\texttt{pysms\_contains\_cliques} & 100/100 & 100/100 & 100/100 & 100/100 & 0/0 & 100/100 & 100/100 & 100/100 & 0/0 & 100/100 & 0/0 & 100/100 \\
\texttt{pysms\_degree\_bounds} & 100/100 & 100/100 & 100/100 & 100/100 & 100/100 & 100/100 & 100/100 & 50/50 & 100/100 & 100/100 & 100/100 & 0/0 \\
\texttt{pysms\_girth\_degree} & 100/100 & 84/84 & 38/69 & 91/91 & 66/81 & 69/72 & 47/47 & 34/38 & 59/72 & 62/66 & 62/69 & 53/53 \\
\texttt{pysms\_graph\_builder} & 50/50 & 33/33 & 33/33 & 33/33 & 33/33 & 50/50 & 17/17 & 17/17 & 33/33 & 33/33 & 33/33 & 67/67 \\
\texttt{pysms\_independent\_connectivity} & 100/100 & 67/83 & 100/100 & 83/100 & 100/100 & 100/100 & 100/100 & 50/50 & 67/100 & 83/83 & 100/100 & 0/0 \\
\texttt{pysms\_min\_girth} & 100/100 & 100/100 & 100/100 & 100/100 & 100/100 & 100/100 & 0/0 & 0/0 & 0/100 & 100/100 & 100/100 & 0/0 \\
\texttt{pysms\_mtf} & 100/100 & 100/100 & 100/100 & 100/100 & 100/100 & 100/100 & 100/100 & 100/100 & 100/100 & 0/100 & 100/100 & 0/0 \\
\texttt{pysms\_ramsey} & 100/100 & 100/100 & 100/100 & 100/100 & 100/100 & 100/100 & 0/0 & 100/100 & 100/100 & 100/100 & 100/100 & 100/100 \\
\texttt{quasigroup\_idempotent} & 80/80 & 80/100 & 80/100 & 60/60 & 60/60 & 40/60 & 20/20 & 0/0 & 80/80 & 60/60 & 20/20 & 20/20 \\
\texttt{queens} & 100/100 & 100/100 & 100/100 & 100/100 & 100/100 & 83/83 & 83/83 & 100/100 & 100/100 & 100/100 & 33/83 & 0/17 \\
\texttt{ramsey} & 100/100 & 79/93 & 93/93 & 100/100 & 93/100 & 93/100 & 57/57 & 71/79 & 93/93 & 71/71 & 7/64 & 7/21 \\
\texttt{social\_golfers} & 56/100 & 81/100 & 44/100 & 94/100 & 0/100 & 25/81 & 31/62 & 12/56 & 6/88 & 25/44 & 6/12 & 0/56 \\
\texttt{sudoku} & 100/100 & 100/100 & 100/100 & 100/100 & 100/100 & 100/100 & 100/100 & 100/100 & 100/100 & 50/50 & 50/100 & 0/0 \\
\texttt{van\_der\_waerden} & 100/100 & 100/100 & 100/100 & 100/100 & 100/100 & 100/100 & 100/100 & 100/100 & 100/100 & 100/100 & 100/100 & 100/100 \\
\texttt{vertex\_cover} & 100/100 & 100/100 & 100/100 & 100/100 & 100/100 & 100/100 & 80/80 & 60/60 & 100/100 & 100/100 & 80/80 & 100/100 \\
\bottomrule
\end{tabular}}
\end{sidewaystable}

\begin{sidewaystable}[p]
\caption{Per-type no-tools performance on \textsc{MathConstraint-Easy}. Each entry is Accuracy/SAT Acc. in percent.}
\label{tab:per-type-easy-no-tools}
\centering
\scriptsize
\setlength{\tabcolsep}{2.5pt}
\resizebox{\textheight}{!}{%
\begin{tabular}{@{}lrrrrrrrrrrrr@{}}
\toprule
Problem type & \textsc{gpt-5.5} & \textsc{claude-opus-4.7} & \textsc{claude-4.6-sonnet} & \textsc{gemini-3.1-pro} & \textsc{gemini-3.1-flash-lite} & \textsc{grok-4.20} & \textsc{deepseek-v4-pro} & \textsc{deepseek-v4-flash} & \textsc{qwen3.6-plus} & \textsc{kimi-k2.6} & \textsc{gpt-oss-120b} & \textsc{llama-3.3-70b-instruct} \\
\midrule
\texttt{all\_interval} & 67/100 & 67/100 & 67/92 & 67/100 & 67/92 & 67/83 & 67/92 & 67/92 & 67/100 & 67/92 & 67/92 & 33/92 \\
\texttt{costas\_array} & 89/89 & 89/89 & 44/100 & 89/100 & 22/100 & 78/78 & 78/78 & 67/89 & 67/78 & 89/89 & 33/78 & 11/100 \\
\texttt{golomb} & 67/100 & 83/100 & 50/100 & 100/100 & 83/83 & 67/100 & 100/100 & 100/100 & 83/83 & 100/100 & 50/67 & 17/67 \\
\texttt{graceful\_graph} & 25/62 & 38/50 & 0/75 & 25/75 & 12/88 & 38/50 & 12/25 & 25/50 & 25/62 & 25/38 & 0/0 & 0/100 \\
\texttt{knight\_tour} & 50/50 & 50/50 & 0/100 & 50/50 & 0/100 & 50/50 & 0/0 & 0/0 & 50/75 & 25/25 & 0/0 & 0/75 \\
\texttt{langford} & 100/100 & 88/88 & 38/100 & 62/62 & 12/88 & 88/100 & 75/88 & 25/75 & 88/88 & 50/62 & 38/50 & 0/75 \\
\texttt{low\_autocorrelation} & 17/25 & 17/17 & 8/92 & 8/42 & 0/83 & 17/17 & 17/25 & 8/33 & 8/25 & 8/17 & 0/0 & 0/92 \\
\texttt{magic\_sequence} & 100/100 & 100/100 & 100/100 & 92/92 & 83/92 & 100/100 & 100/100 & 100/100 & 100/100 & 92/92 & 58/67 & 8/83 \\
\texttt{pigeons} & 100/100 & 100/100 & 100/100 & 100/100 & 100/100 & 100/100 & 100/100 & 100/100 & 100/100 & 100/100 & 100/100 & 100/100 \\
\texttt{pysms\_chromatic\_girth} & 100/100 & 92/92 & 75/75 & 83/83 & 83/83 & 92/92 & 83/83 & 92/92 & 92/92 & 75/75 & 100/100 & 25/25 \\
\texttt{pysms\_clique\_coloring} & 100/100 & 100/100 & 100/100 & 100/100 & 100/100 & 100/100 & 100/100 & 83/92 & 92/92 & 100/100 & 67/75 & 8/17 \\
\texttt{pysms\_combined\_graph} & 100/100 & 100/100 & 92/92 & 100/100 & 100/100 & 100/100 & 100/100 & 100/100 & 100/100 & 100/100 & 92/92 & 83/83 \\
\texttt{pysms\_contains\_cliques} & 100/100 & 100/100 & 92/92 & 100/100 & 100/100 & 100/100 & 100/100 & 92/92 & 100/100 & 100/100 & 92/92 & 25/25 \\
\texttt{pysms\_degree\_bounds} & 100/100 & 92/100 & 92/92 & 100/100 & 100/100 & 100/100 & 100/100 & 83/92 & 100/100 & 83/83 & 92/92 & 8/8 \\
\texttt{pysms\_girth\_degree} & 92/92 & 92/92 & 83/83 & 92/92 & 83/83 & 100/100 & 83/83 & 100/100 & 92/92 & 92/92 & 83/92 & 33/33 \\
\texttt{pysms\_graph\_builder} & 75/75 & 67/67 & 67/67 & 67/67 & 67/67 & 67/67 & 67/67 & 58/67 & 58/58 & 67/67 & 50/58 & 42/42 \\
\texttt{pysms\_independent\_connectivity} & 100/100 & 100/100 & 100/100 & 100/100 & 92/100 & 100/100 & 100/100 & 67/75 & 100/100 & 92/92 & 92/92 & 33/33 \\
\texttt{pysms\_min\_connectivity} & 100/100 & 100/100 & 100/100 & 100/100 & 100/100 & 100/100 & 100/100 & 67/83 & 92/92 & 75/75 & 100/100 & 8/25 \\
\texttt{pysms\_min\_degree} & 100/100 & 100/100 & 100/100 & 100/100 & 92/100 & 100/100 & 100/100 & 83/100 & 100/100 & 92/100 & 83/92 & 8/17 \\
\texttt{pysms\_min\_girth} & 100/100 & 100/100 & 100/100 & 100/100 & 92/100 & 100/100 & 100/100 & 100/100 & 83/83 & 92/92 & 92/92 & 17/25 \\
\texttt{pysms\_mtf} & 92/92 & 100/100 & 100/100 & 100/100 & 92/100 & 92/92 & 100/100 & 92/100 & 100/100 & 75/92 & 100/100 & 17/25 \\
\texttt{pysms\_num\_edges\_bounds} & 100/100 & 100/100 & 100/100 & 100/100 & 100/100 & 100/100 & 100/100 & 100/100 & 92/92 & 92/92 & 100/100 & 0/17 \\
\texttt{queens} & 83/83 & 75/75 & 50/100 & 67/75 & 42/100 & 58/58 & 67/75 & 58/92 & 58/83 & 58/58 & 42/58 & 42/100 \\
\texttt{ramsey} & 100/100 & 75/92 & 50/100 & 92/100 & 42/100 & 92/92 & 75/83 & 58/92 & 75/92 & 67/83 & 42/58 & 0/58 \\
\texttt{sudoku} & 67/67 & 33/33 & 33/100 & 33/33 & 0/100 & 33/33 & 33/33 & 33/67 & 33/67 & 33/33 & 33/33 & 0/67 \\
\bottomrule
\end{tabular}}
\end{sidewaystable}

\begin{sidewaystable}[p]
\caption{Per-type tool-enabled performance on \textsc{MathConstraint-Easy}. Each entry is Accuracy/SAT Acc. in percent.}
\label{tab:per-type-easy-tools}
\centering
\scriptsize
\setlength{\tabcolsep}{2.5pt}
\resizebox{\textheight}{!}{%
\begin{tabular}{@{}lrrrrrrrrrrrr@{}}
\toprule
Problem type & \textsc{gpt-5.5} & \textsc{claude-opus-4.7} & \textsc{claude-4.6-sonnet} & \textsc{gemini-3.1-pro} & \textsc{gemini-3.1-flash-lite} & \textsc{grok-4.20} & \textsc{deepseek-v4-pro} & \textsc{deepseek-v4-flash} & \textsc{qwen3.6-plus} & \textsc{kimi-k2.6} & \textsc{gpt-oss-120b} & \textsc{llama-3.3-70b-instruct} \\
\midrule
\texttt{all\_interval} & 67/100 & 67/100 & 67/100 & 67/100 & 67/92 & 67/92 & 33/67 & 67/83 & 67/92 & 58/83 & 58/100 & 42/58 \\
\texttt{costas\_array} & 100/100 & 100/100 & 100/100 & 100/100 & 89/100 & 100/100 & 67/67 & 89/89 & 100/100 & 100/100 & 44/100 & 44/78 \\
\texttt{golomb} & 67/100 & 100/100 & 100/100 & 100/100 & 100/100 & 100/100 & 83/83 & 67/67 & 100/100 & 83/83 & 50/83 & 17/67 \\
\texttt{graceful\_graph} & 62/88 & 62/100 & 50/100 & 62/100 & 50/88 & 38/75 & 25/50 & 12/50 & 38/75 & 62/100 & 0/25 & 0/25 \\
\texttt{knight\_tour} & 100/100 & 75/75 & 75/100 & 100/100 & 75/75 & 75/75 & 75/75 & 100/100 & 100/100 & 100/100 & 25/75 & 0/25 \\
\texttt{langford} & 88/88 & 100/100 & 100/100 & 100/100 & 88/100 & 100/100 & 38/62 & 88/88 & 75/75 & 100/100 & 38/50 & 0/38 \\
\texttt{low\_autocorrelation} & 33/67 & 33/50 & 33/58 & 33/50 & 33/58 & 33/42 & 33/33 & 33/50 & 33/67 & 33/50 & 0/25 & 33/58 \\
\texttt{magic\_sequence} & 100/100 & 100/100 & 100/100 & 100/100 & 100/100 & 83/92 & 92/92 & 100/100 & 100/100 & 83/83 & 83/83 & 25/33 \\
\texttt{pigeons} & 100/100 & 100/100 & 100/100 & 100/100 & 100/100 & 100/100 & 100/100 & 100/100 & 100/100 & 100/100 & 92/92 & 100/100 \\
\texttt{pysms\_chromatic\_girth} & 92/100 & 92/92 & 83/83 & 75/83 & 92/92 & 100/100 & 75/75 & 50/58 & 92/92 & 67/92 & 100/100 & 25/25 \\
\texttt{pysms\_clique\_coloring} & 100/100 & 100/100 & 100/100 & 100/100 & 100/100 & 92/92 & 25/25 & 33/33 & 83/92 & 92/92 & 75/92 & 0/0 \\
\texttt{pysms\_combined\_graph} & 100/100 & 100/100 & 100/100 & 100/100 & 100/100 & 100/100 & 83/83 & 92/92 & 100/100 & 92/92 & 100/100 & 100/100 \\
\texttt{pysms\_contains\_cliques} & 100/100 & 100/100 & 100/100 & 100/100 & 83/83 & 100/100 & 100/100 & 75/75 & 92/92 & 100/100 & 92/92 & 58/58 \\
\texttt{pysms\_degree\_bounds} & 100/100 & 100/100 & 100/100 & 100/100 & 100/100 & 75/83 & 75/75 & 58/75 & 100/100 & 83/83 & 100/100 & 17/25 \\
\texttt{pysms\_girth\_degree} & 100/100 & 92/92 & 75/83 & 100/100 & 92/92 & 100/100 & 67/67 & 67/67 & 100/100 & 75/75 & 83/83 & 50/50 \\
\texttt{pysms\_graph\_builder} & 75/75 & 67/67 & 67/67 & 67/67 & 58/67 & 75/75 & 58/58 & 50/58 & 67/67 & 58/58 & 42/42 & 50/50 \\
\texttt{pysms\_independent\_connectivity} & 100/100 & 83/92 & 100/100 & 92/100 & 100/100 & 100/100 & 67/67 & 25/25 & 92/100 & 67/67 & 83/83 & 0/0 \\
\texttt{pysms\_min\_connectivity} & 100/100 & 100/100 & 100/100 & 100/100 & 100/100 & 92/100 & 83/83 & 75/83 & 83/100 & 92/92 & 92/92 & 0/17 \\
\texttt{pysms\_min\_degree} & 100/100 & 100/100 & 100/100 & 100/100 & 100/100 & 92/92 & 92/92 & 92/100 & 100/100 & 100/100 & 92/100 & 8/8 \\
\texttt{pysms\_min\_girth} & 100/100 & 100/100 & 100/100 & 100/100 & 100/100 & 92/100 & 100/100 & 92/100 & 100/100 & 100/100 & 92/100 & 0/0 \\
\texttt{pysms\_mtf} & 100/100 & 100/100 & 100/100 & 100/100 & 100/100 & 83/100 & 92/100 & 92/100 & 100/100 & 92/100 & 100/100 & 0/0 \\
\texttt{pysms\_num\_edges\_bounds} & 100/100 & 100/100 & 100/100 & 100/100 & 100/100 & 100/100 & 83/83 & 83/100 & 100/100 & 100/100 & 100/100 & 0/17 \\
\texttt{queens} & 100/100 & 100/100 & 100/100 & 100/100 & 100/100 & 92/92 & 83/83 & 100/100 & 100/100 & 92/92 & 67/100 & 42/58 \\
\texttt{ramsey} & 100/100 & 83/100 & 92/92 & 100/100 & 100/100 & 92/100 & 67/67 & 75/83 & 92/92 & 67/67 & 33/83 & 25/33 \\
\texttt{sudoku} & 100/100 & 100/100 & 100/100 & 100/100 & 100/100 & 100/100 & 100/100 & 100/100 & 100/100 & 67/67 & 33/67 & 0/0 \\
\bottomrule
\end{tabular}}
\end{sidewaystable}

\FloatBarrier
\section{Appendix Figures}
\label{app:figures}

This appendix collects the remaining diagnostic figures referenced in the main text.

\begin{figure}[!htbp]
    \centering
    \includegraphics[width=0.82\linewidth]{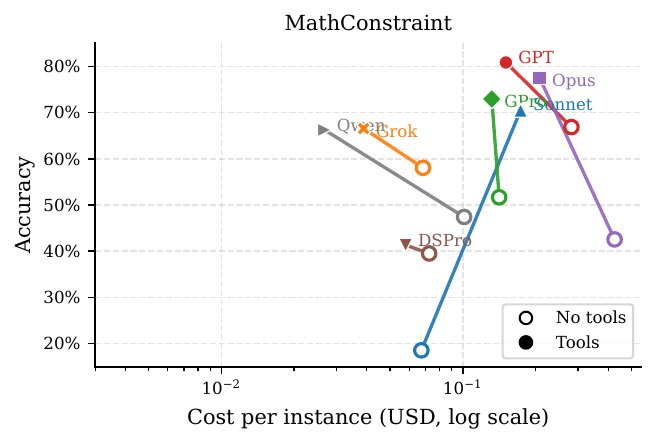}
    \caption{Cost--accuracy tradeoff on \textsc{MathConstraint}. Open markers are no-tools runs; filled markers are tool-enabled runs. Costs and accuracies are tabulated in \cref{tab:eval-efficiency-mathconstraint-no-tools,tab:eval-efficiency-mathconstraint-tools,tab:mathconstraint-no-tools,tab:mathconstraint-tools}.}
    \label{fig:cost-accuracy}
\end{figure}

\begin{figure}[!htbp]
    \centering
    \includegraphics[width=\linewidth]{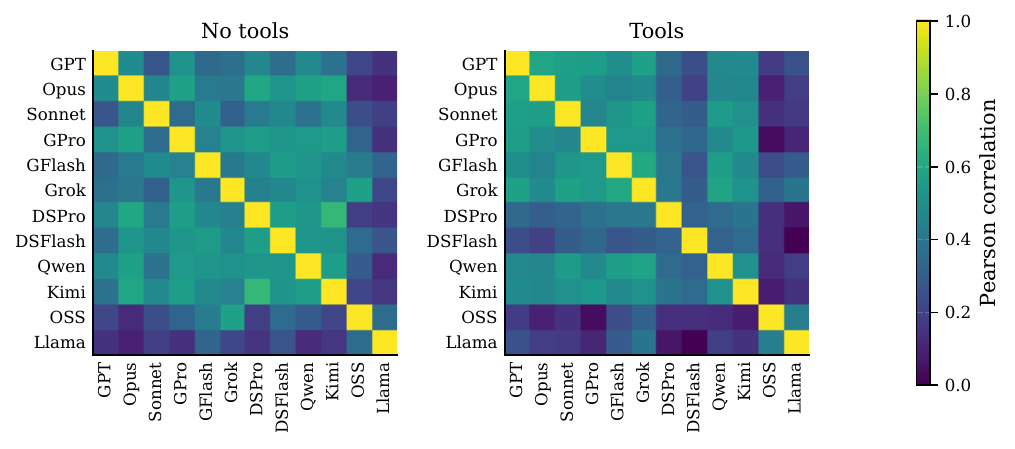}
    \caption{Pairwise correctness correlations across models on \textsc{MathConstraint}, shown separately for no-tools and tool-enabled runs. Aggregate accuracies for the same runs are in \cref{tab:mathconstraint-no-tools,tab:mathconstraint-tools}.}
    \label{fig:model-correlations}
\end{figure}

\begin{figure}[t]
    \centering
    \includegraphics[width=0.86\linewidth]{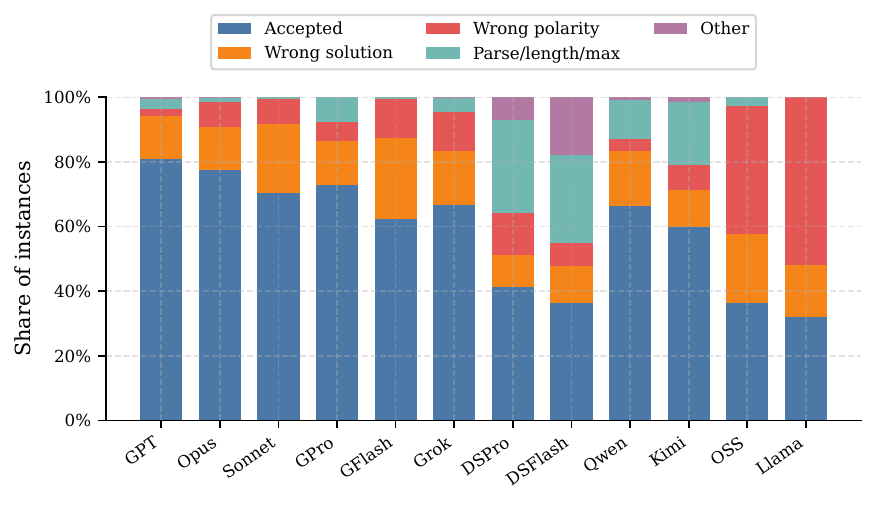}
    \caption{Failure-bucket decomposition on tool-enabled \textsc{MathConstraint} runs. Buckets are assigned by the evaluator after parsing and verification; counts are tabulated in \cref{tab:failure-buckets-tools}.}
    \label{fig:failure-buckets}
\end{figure}

\begin{figure}[t]
    \centering
    \includegraphics[width=\linewidth]{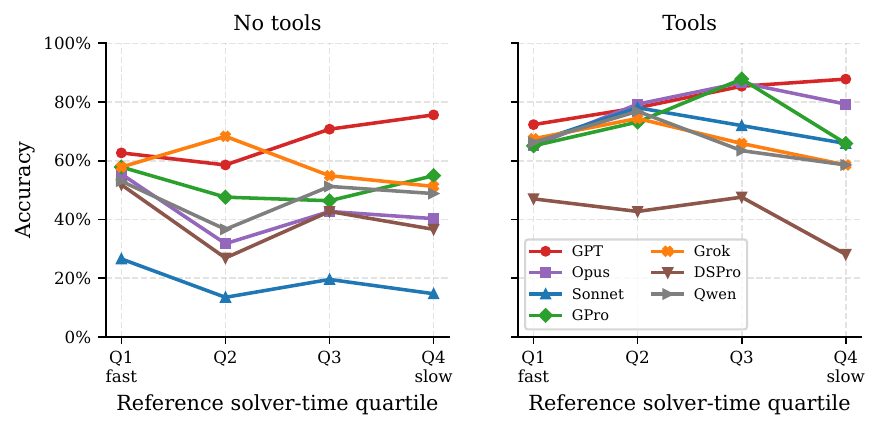}
    \caption{Accuracy stratified by generation-time solver-time quartile on \textsc{MathConstraint}. Solver-time distribution summaries are in \cref{tab:dataset-difficulty,tab:dataset-difficulty-appendix}; overall accuracies are in \cref{tab:mathconstraint-no-tools,tab:mathconstraint-tools}.}
    \label{fig:solver-time-accuracy}
\end{figure}

\begin{figure}[t]
    \centering
    \includegraphics[width=0.7\linewidth]{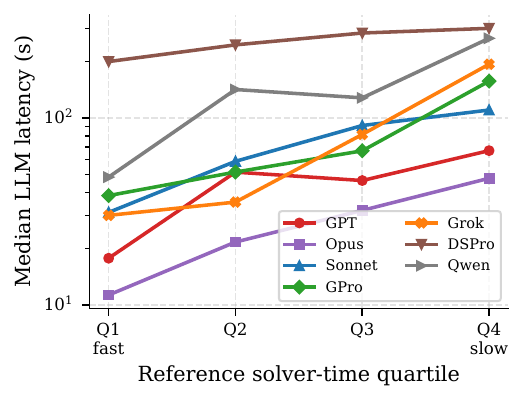}
    \caption{Median tool-enabled LLM latency by generation-time solver-time quartile on \textsc{MathConstraint}. Solver-time summaries are in \cref{tab:dataset-difficulty,tab:dataset-difficulty-appendix}; per-instance average latency is in \cref{tab:eval-efficiency-mathconstraint-tools}.}
    \label{fig:solver-time-latency}
\end{figure}

\begin{figure}[t]
    \centering
    \includegraphics[width=0.9\linewidth]{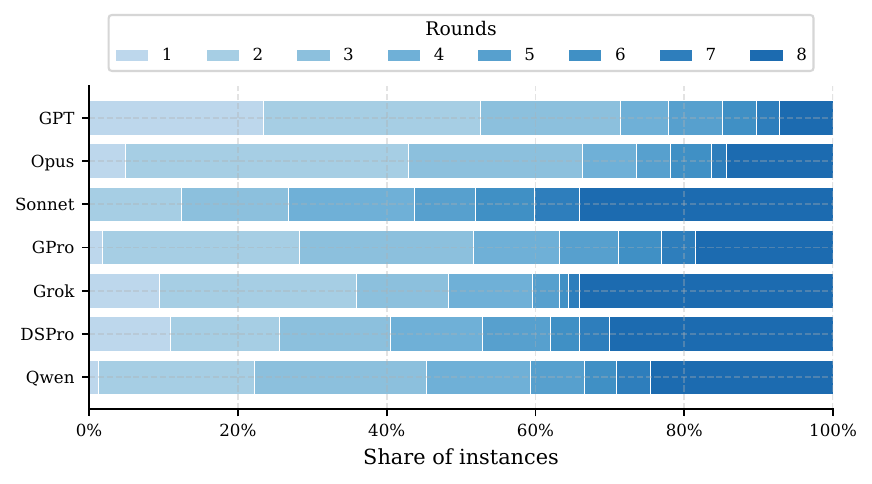}
    \caption{Distribution of tool rounds used by each model on \textsc{MathConstraint}. Average calls, rounds, submissions, and force-submit rates are in \cref{tab:tool-use-behavior-mathconstraint}.}
    \label{fig:tool-round-histogram}
\end{figure}

\begin{figure}[t]
    \centering
    \includegraphics[width=\linewidth]{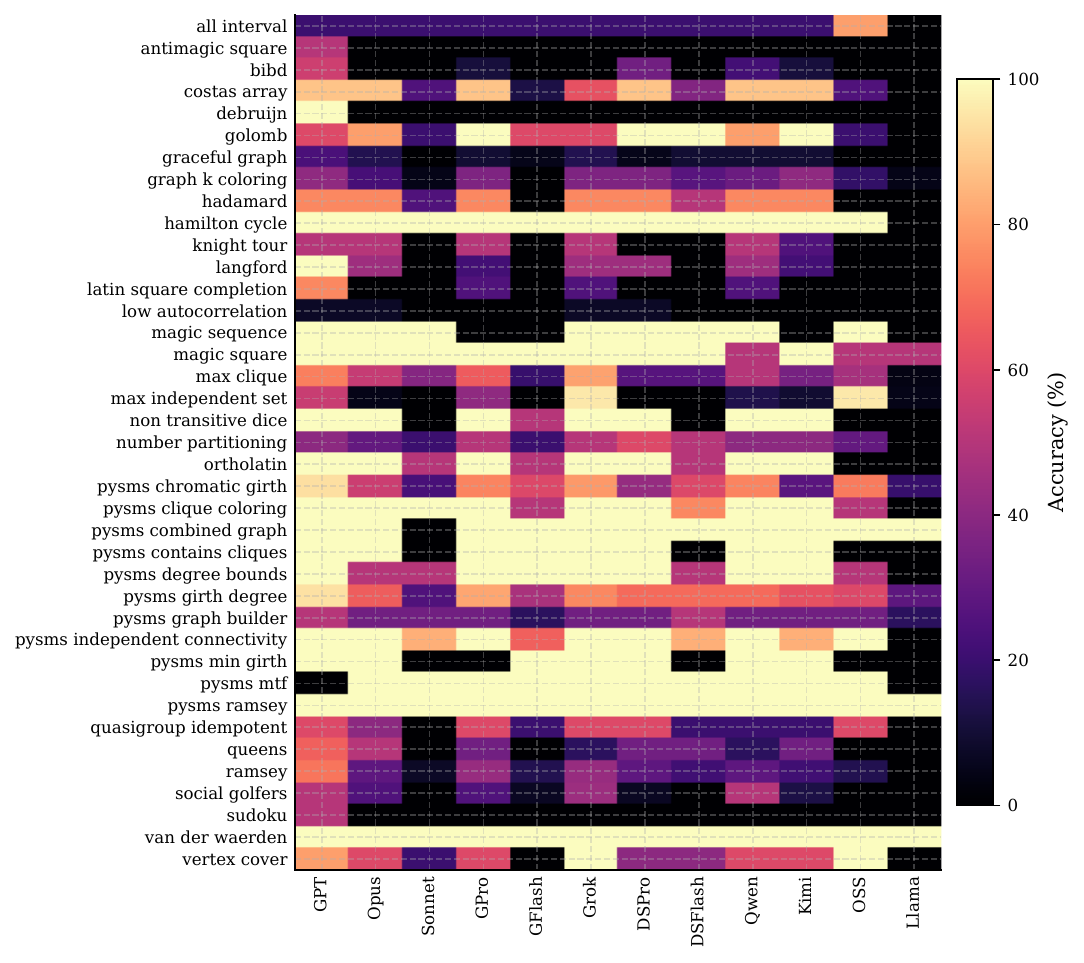}
    \caption{Per-type no-tools accuracy heatmap on \textsc{MathConstraint}. The full per-type no-tools table is \cref{tab:per-type-mathconstraint-no-tools}.}
    \label{fig:per-type-accuracy-heatmap}
\end{figure}

\FloatBarrier

\section{Example Traces}
\label{app:traces}
This appendix shows representative traces from the released evaluations. Each trace is split into the same observable fields used by the evaluation harness: prompt, reasoning trace summary, tool calls, final output, and verifier verdict. We do not reproduce hidden chain-of-thought verbatim; the reasoning boxes below are concise summaries of the recorded behavior, while tool code, final submissions, and verifier outcomes are copied from the stored trace records. The stored trace JSON records tool call status and latency, but not full stdout/stderr payloads.

\tcbset{
  tracepromptstyle/.style={
    breakable,
    colback=gray!5,
    colframe=gray!50!black,
    boxsep=3pt, top=5pt, bottom=5pt, left=6pt, right=6pt,
    fontupper=\footnotesize\ttfamily,
    fonttitle=\bfseries\footnotesize,
    arc=2pt,
  },
  tracereasonstyle/.style={
    breakable,
    colback=blue!5,
    colframe=blue!55!black,
    boxsep=3pt, top=5pt, bottom=5pt, left=6pt, right=6pt,
    fontupper=\footnotesize,
    fonttitle=\bfseries\footnotesize,
    arc=2pt,
  },
  tracecodestyle/.style={
    breakable,
    colback=orange!8,
    colframe=orange!65!black,
    boxsep=3pt, top=5pt, bottom=5pt, left=6pt, right=6pt,
    fontupper=\scriptsize\ttfamily,
    fonttitle=\bfseries\footnotesize,
    arc=2pt,
  },
  traceoutputstyle/.style={
    breakable,
    colback=yellow!7,
    colframe=yellow!45!black,
    boxsep=3pt, top=5pt, bottom=5pt, left=6pt, right=6pt,
    fontupper=\scriptsize\ttfamily,
    fonttitle=\bfseries\footnotesize,
    arc=2pt,
  },
  traceanswerstyle/.style={
    breakable,
    colback=green!7,
    colframe=green!55!black,
    boxsep=3pt, top=5pt, bottom=5pt, left=6pt, right=6pt,
    fontupper=\scriptsize\ttfamily,
    fonttitle=\bfseries\footnotesize,
    arc=2pt,
  },
  traceverdictokstyle/.style={
    breakable,
    colback=green!7,
    colframe=green!55!black,
    boxsep=3pt, top=5pt, bottom=5pt, left=6pt, right=6pt,
    fontupper=\footnotesize\ttfamily,
    fonttitle=\bfseries\footnotesize,
    arc=2pt,
  },
  traceverdictbadstyle/.style={
    breakable,
    colback=red!5,
    colframe=red!45!black,
    boxsep=3pt, top=5pt, bottom=5pt, left=6pt, right=6pt,
    fontupper=\footnotesize\ttfamily,
    fonttitle=\bfseries\footnotesize,
    arc=2pt,
  },
}

\subsection{Trace 1: Successful tool use}

\noindent\textbf{Model:} \texttt{deepseek/deepseek-v4-flash-20260423}. \textbf{Condition:} \texttt{tools}. \textbf{Problem:} \texttt{costas\_array\_n10\_\_v0\_nh}.

\begin{tcolorbox}[tracepromptstyle, title={Prompt}]
Find a Costas array of order 10: place 10 marks on an 10x10 grid, one per row and one per column, such that all 10*(9)/2 displacement vectors (dr, dc) between pairs of marks (with dc > 0) are pairwise distinct.

Return x as a list of 10 integers in 0..9, where x[c] is the row of the mark in column c (permutation), or state "UNSATISFIABLE" if no such array exists.
\end{tcolorbox}

\begin{tcolorbox}[tracereasonstyle, title={Reasoning trace summary}]
A tool-enabled model writes and verifies a backtracking search, then submits a valid witness.

Recorded final reasoning: Found a Costas array of order 10 via backtracking search with constraint propagation. The permutation [0, 1, 3, 7, 4, 9, 8, 6, 2, 5] places one mark per row and column, and all 45 displacement vectors (dr, dc) between pairs of marks are pairwise distinct, satisfying the Costas array property.
\end{tcolorbox}

\begin{tcolorbox}[tracecodestyle, title={Tool code}]
Tool call 1: execute\_python
import itertools
import sys

\# Let's try a backtracking approach first to see if we can find a Costas array of order 10

def is\_costas(perm):
    n = len(perm)
    vectors = set()
    for i in range(n):
        for j in range(i+1, n):
            dr = perm[j] - perm[i]
            dc = j - i
            if (dr, dc) in vectors:
                return False
            vectors.add((dr, dc))
    return True

\# Known Costas arrays of order 10 exist. Let me try to find one using known constructions.
\# The Welch construction works for primes. For n=10, we can use the Lempel construction or search.

\# Let's try a known Costas array of order 10
\# Known example from literature:
known = [0, 2, 7, 5, 3, 8, 6, 4, 9, 1]  \# This is a known one? Let me check

print("Checking known array:", known)
print("Is Costas?", is\_costas(known))

\# Let me try another known one
known2 = [0, 1, 3, 6, 2, 7, 5, 8, 4, 9]
print("Checking known2:", known2)
print("Is Costas?", is\_costas(known2))

\# Let me search using backtracking with pruning
def find\_costas(n):
    \# Use backtracking with constraint propagation
    used\_rows = [False] * n
    \# Store all displacement vectors seen so far
    vectors = set()
    
    def backtrack(col, perm):
        if col == n:
            return perm[:]
        
        for row in range(n):
            if used\_rows[row]:
                continue
            
            \# Check if adding this position creates any duplicate vectors
            valid = True
            new\_vectors = []
            for prev\_col in range(col):
                prev\_row = perm[prev\_col]
                dr = row - prev\_row
                dc = col - prev\_col
                if (dr, dc) in vectors:
                    valid = False
                    break
                new\_vectors.append(
... [truncated]

Tool call 2: execute\_python
def verify\_costas(perm):
    n = len(perm)
    print(f"Permutation: \{perm\}")
    print(f"Length: \{n\}")
    
    \# Check one per row
    if sorted(perm) != list(range(n)):
        print("FAIL: Not a permutation")
        return False
    
    vectors = \{\}
    for i in range(n):
        for j in range(i+1, n):
            dr = perm[j] - perm[i]
            dc = j - i
            key = (dr, dc)
            if key in vectors:
                print(f"FAIL: Duplicate vector \{key\} from columns (\{i\},\{j\}) and \{vectors[key]\}")
                return False
            vectors[key] = (i, j)
    
    print(f"All \{n*(n-1)//2\} vectors are distinct!")
    print("Vectors:", sorted(vectors.keys()))
    return True

perm = [0, 1, 3, 7, 4, 9, 8, 6, 2, 5]
verify\_costas(perm)

\# Also visualize
print("\textbackslash{}nGrid visualization:")
for r in range(9, -1, -1):
    line = ""
    for c in range(10):
        if perm[c] == r:
            line += "X "
        else:
            line += ". "
    print(f"\{r:2d\} \{line\}")
print("   ", end="")
for c in range(10):
    print(f"\{c\} ", end="")
print()

Tool call 3: submit\_answer
\{
  "satisfiable": true,
  "solution": [
    0,
    1,
    3,
    7,
    4,
    9,
    8,
    6,
    2,
    5
  ],
  "reasoning": "Found a Costas array of order 10 via backtracking search with constraint propagation. The permutation [0, 1, 3, 7, 4, 9, 8, 6, 2, 5] places one mark per row and column, and all 45 displacement vectors (dr, dc) between pairs of marks are pairwise distinct, satisfying the Costas array property."
\}
\end{tcolorbox}
\begin{tcolorbox}[traceoutputstyle, title={Tool responses}]
Tool call 1: execute\_python
ok=True latency\_ms=34.71
Note: stdout/stderr text is not present in the stored trace JSON; only call status and latency are recorded.

Tool call 2: execute\_python
ok=True latency\_ms=32.83
Note: stdout/stderr text is not present in the stored trace JSON; only call status and latency are recorded.

Tool call 3: submit\_answer
ok=True latency\_ms=0.0
\end{tcolorbox}

\begin{tcolorbox}[traceanswerstyle, title={Final output}]
\{
  "satisfiable": true,
  "solution": [
    0,
    1,
    3,
    7,
    4,
    9,
    8,
    6,
    2,
    5
  ],
  "reasoning": "Found a Costas array of order 10 via backtracking search with constraint propagation. The permutation [0, 1, 3, 7, 4, 9, 8, 6, 2, 5] places one mark per row and column, and all 45 displacement vectors (dr, dc) between pairs of marks are pairwise distinct, satisfying the Costas array property."
\}
\end{tcolorbox}

\begin{tcolorbox}[traceverdictokstyle, title={Verification verdict}]
ground\_truth\_satisfiable=True
submitted\_satisfiable=True
correct=True
satisfiability\_correct=True
solution\_correct=True
validation\_details=Solution verified by solver
failure\_bucket=none
\end{tcolorbox}

\subsection{Trace 2: Successful no-tools construction}

\noindent\textbf{Model:} \texttt{deepseek/deepseek-v4-flash-20260423}. \textbf{Condition:} \texttt{no\_tools}. \textbf{Problem:} \texttt{all\_interval\_n11\_\_v8\_nh}.

\begin{tcolorbox}[tracepromptstyle, title={Prompt}]
Find an all-interval series of size 11. This is a permutation of 0 to 10 such that the absolute differences between consecutive elements are also a permutation of 1 to 10.

Return a list of 11 integers forming a valid permutation, or state "UNSATISFIABLE" if no solution exists.
\end{tcolorbox}

\begin{tcolorbox}[tracereasonstyle, title={Reasoning trace summary}]
A no-tools model uses a compact constructive pattern and submits a valid witness directly.

Recorded final reasoning: A known construction for all-interval series of size n (where n is odd) is to interleave the smallest and largest remaining numbers: starting with 0, then n-1, then 1, then n-2, etc. For n=11, this yields the sequence 0,10,1,9,2,8,3,7,4,6,5, and the absolute differences are 10,9,8,7,6,5,4,3,2,1, which is a permutation of 1 to 10.
\end{tcolorbox}

\begin{tcolorbox}[tracecodestyle, title={Tool code}]
No tool calls.
\end{tcolorbox}
\begin{tcolorbox}[traceoutputstyle, title={Tool responses}]
No tool responses.
\end{tcolorbox}

\begin{tcolorbox}[traceanswerstyle, title={Final output}]
\{
  "satisfiable": true,
  "solution": [
    0,
    10,
    1,
    9,
    2,
    8,
    3,
    7,
    4,
    6,
    5
  ],
  "reasoning": "A known construction for all-interval series of size n (where n is odd) is to interleave the smallest and largest remaining numbers: starting with 0, then n-1, then 1, then n-2, etc. For n=11, this yields the sequence 0,10,1,9,2,8,3,7,4,6,5, and the absolute differences are 10,9,8,7,6,5,4,3,2,1, which is a permutation of 1 to 10."
\}
\end{tcolorbox}

\begin{tcolorbox}[traceverdictokstyle, title={Verification verdict}]
ground\_truth\_satisfiable=True
submitted\_satisfiable=True
correct=True
satisfiability\_correct=True
solution\_correct=True
validation\_details=Solution verified by solver
failure\_bucket=none
\end{tcolorbox}

\subsection{Trace 3: Tool-use failure despite local check}

\noindent\textbf{Model:} \texttt{deepseek/deepseek-v4-flash-20260423}. \textbf{Condition:} \texttt{tools}. \textbf{Problem:} \texttt{all\_interval\_n2\_\_v0\_h}.

\begin{tcolorbox}[tracepromptstyle, title={Prompt}]
Find an all-interval series of size 2. This is a permutation of 0 to 1 such that the absolute differences between consecutive elements are also a permutation of 1 to 1.

Return a list of 2 integers forming a valid permutation, or state "UNSATISFIABLE" if no solution exists.

Partial assignment (fixed values that must be respected):
- x[0]=0
- d[0]=1
Return a complete solution consistent with these fixed assignments.
\end{tcolorbox}

\begin{tcolorbox}[tracereasonstyle, title={Reasoning trace summary}]
The model locally checks the mathematical object but submits the wrong schema for a hinted instance.

Recorded final reasoning: We need a permutation of [0,1] such that absolute differences between consecutive elements form a permutation of [1]. Given x[0]=0 and d[0]=1, we need |x[1]-0|=1, so x[1]=1 (since it must be from \{0,1\} and 0 is already used). This gives differences [1] which is a permutation of [1]. The solution [0,1] satisfies all constraints.
\end{tcolorbox}

\begin{tcolorbox}[tracecodestyle, title={Tool code}]
Tool call 1: execute\_python
\# Quick verification
x = [0, 1]
d = [abs(x[1] - x[0])]
print("x:", x)
print("d:", d)
print("Is x a permutation of [0,1]?", sorted(x) == [0, 1])
print("Is d a permutation of [1]?", sorted(d) == [1])

Tool call 2: submit\_answer
\{
  "satisfiable": true,
  "solution": [
    0,
    1
  ],
  "reasoning": "We need a permutation of [0,1] such that absolute differences between consecutive elements form a permutation of [1]. Given x[0]=0 and d[0]=1, we need |x[1]-0|=1, so x[1]=1 (since it must be from \{0,1\} and 0 is already used). This gives differences [1] which is a permutation of [1]. The solution [0,1] satisfies all constraints."
\}
\end{tcolorbox}
\begin{tcolorbox}[traceoutputstyle, title={Tool responses}]
Tool call 1: execute\_python
ok=True latency\_ms=33.06
Note: stdout/stderr text is not present in the stored trace JSON; only call status and latency are recorded.

Tool call 2: submit\_answer
ok=True latency\_ms=0.0
\end{tcolorbox}

\begin{tcolorbox}[traceanswerstyle, title={Final output}]
\{
  "satisfiable": true,
  "solution": [
    0,
    1
  ],
  "reasoning": "We need a permutation of [0,1] such that absolute differences between consecutive elements form a permutation of [1]. Given x[0]=0 and d[0]=1, we need |x[1]-0|=1, so x[1]=1 (since it must be from \{0,1\} and 0 is already used). This gives differences [1] which is a permutation of [1]. The solution [0,1] satisfies all constraints."
\}
\end{tcolorbox}

\begin{tcolorbox}[traceverdictbadstyle, title={Verification verdict}]
ground\_truth\_satisfiable=True
submitted\_satisfiable=True
correct=False
satisfiability\_correct=True
solution\_correct=False
validation\_details=Partial assignment: submitted solution not keyed by variable
failure\_bucket=wrong\_solution
\end{tcolorbox}

\subsection{Trace 4: UNSAT after failed search}

\noindent\textbf{Model:} \texttt{deepseek/deepseek-v4-flash-20260423}. \textbf{Condition:} \texttt{tools}. \textbf{Problem:} \texttt{langford\_n6\_\_v8}.

\begin{tcolorbox}[tracepromptstyle, title={Prompt}]
Construct a Langford sequence L(2,6): a sequence of length 12 containing exactly 2 copies of each integer from 1 to 6, such that the two occurrences of each value v are exactly v+1 positions apart (so copies of 1 are 1 position apart, copies of 2 are 2 positions apart, etc.). Example: L(2,3) has solution [2, 3, 1, 2, 1, 3].

Return seq as a list of 12 integers in 1..6, or state "UNSATISFIABLE" if no such sequence exists.
\end{tcolorbox}

\begin{tcolorbox}[tracereasonstyle, title={Reasoning trace summary}]
The model tries search, sees no construction, invokes the Langford parity theorem, and correctly submits UNSAT.

Recorded final reasoning: Langford sequences L(2,n) exist if and only if n ≡ 0 or 3 (mod 4). Since 6 ≡ 2 (mod 4), no valid Langford sequence of length 12 exists for n=6. This is a well-known result in combinatorial design theory, and exhaustive backtracking search confirms unsatisfiability.
\end{tcolorbox}

\begin{tcolorbox}[tracecodestyle, title={Tool code}]
Tool call 1: execute\_python
\# Let's solve L(2,6) - Langford pairing problem
\# We need to place two copies of each number 1..6 in a sequence of length 12
\# such that the distance between the two copies of v is exactly v+1

\# Let's use backtracking

def solve\_langford(n):
    """Solve L(2,n) Langford sequence"""
    length = 2 * n
    seq = [0] * length
    
    \# positions[v] will store the position of the first occurrence of v
    \# We'll place numbers from n down to 1 (heuristic for faster solving)
    
    def backtrack(v):
        if v == 0:
            return True
        
        \# Try to place the two copies of v
        \# They must be v+1 apart
        for i in range(length - v - 1):
            if seq[i] == 0 and seq[i + v + 1] == 0:
                seq[i] = v
                seq[i + v + 1] = v
                if backtrack(v - 1):
                    return True
                seq[i] = 0
                seq[i + v + 1] = 0
        return False
    
    if backtrack(n):
        return seq
    return None

result = solve\_langford(6)
print(result)

Tool call 2: execute\_python
\# Langford sequences L(2,n) exist iff n ≡ 0 or 3 (mod 4)
\# Let's check: 6 mod 4 = 2, so L(2,6) should be UNSATISFIABLE

\# Let's verify with a more thorough search anyway
import itertools

def check\_langford(seq, n):
    """Check if seq is a valid Langford sequence L(2,n)"""
    length = 2 * n
    if len(seq) != length:
        return False
    \# Check each value 1..n appears exactly twice
    for v in range(1, n+1):
        positions = [i for i, x in enumerate(seq) if x == v]
        if len(positions) != 2:
            return False
        if positions[1] - positions[0] != v + 1:
            return False
    return True

\# Known theorem: Langford sequences L(2,n) exist iff n ≡ 0 or 3 (mod 4)
\# 6 ≡ 2 (mod 4), so no solution exists
print(f"6 mod 4 = \{6 \% 4\}")
print("Known theorem: L(2,n) exists iff n ≡ 0 or 3 (mod 4)")
print("Since 6 ≡ 2 (mod 4), L(2,6) is UNSATISFIABLE")

\# Let's also do a quick exhaustive search for n=6 to confirm
\# But 12! is too large, so let's use constraint programming

\# Actually let's use pysat to confirm
from pysat import SATSolver
from pysat.formula import CNF

\# Or just use a smarter backtracking

def solve\_langford\_bruteforce(n):
    """More systematic backtracking"""
    length = 2 * n
    seq = [0] * length
    
    \# Track which numbers have been placed
    placed = [False] * (n + 1)
    
    def backtrack(pos):
        if pos == length:
            return True
        
        if seq[pos] != 0:
            return backtrack(pos + 1)
        
        \# Try placing each number v that hasn't been placed yet
        for v in range(1, n + 1):
            if not placed[v]:
                second\_pos = pos + v + 1
                if second\_pos < length and seq[second\_pos] == 0:
                    seq[pos] = v
                    seq[second\_pos] =
... [truncated]

Tool call 3: execute\_python
def solve\_langford(n):
    """More systematic backtracking"""
    length = 2 * n
    seq = [0] * length
    
    placed = [False] * (n + 1)
    
    def backtrack(pos):
        if pos == length:
            return True
        
        if seq[pos] != 0:
            return backtrack(pos + 1)
        
        for v in range(1, n + 1):
            if not placed[v]:
                second\_pos = pos + v + 1
                if second\_pos < length and seq[second\_pos] == 0:
                    seq[pos] = v
                    seq[second\_pos] = v
                    placed[v] = True
                    if backtrack(pos + 1):
                        return True
                    seq[pos] = 0
                    seq[second\_pos] = 0
                    placed[v] = False
        return False
    
    if backtrack(0):
        return seq
    return None

\# Test for various n
for n in [3, 4, 5, 6, 7]:
    r = solve\_langford(n)
    print(f"n=\{n\}: \{'SAT' if r else 'UNSAT'\} (n mod 4 = \{n \% 4\}), solution: \{r\}")

Tool call 4: submit\_answer
\{
  "satisfiable": false,
  "solution": null,
  "reasoning": "Langford sequences L(2,n) exist if and only if n ≡ 0 or 3 (mod 4). Since 6 ≡ 2 (mod 4), no valid Langford sequence of length 12 exists for n=6. This is a well-known result in combinatorial design theory, and exhaustive backtracking search confirms unsatisfiability."
\}
\end{tcolorbox}
\begin{tcolorbox}[traceoutputstyle, title={Tool responses}]
Tool call 1: execute\_python
ok=True latency\_ms=34.8
Note: stdout/stderr text is not present in the stored trace JSON; only call status and latency are recorded.

Tool call 2: execute\_python
ok=False latency\_ms=35.3
Note: stdout/stderr text is not present in the stored trace JSON; only call status and latency are recorded.

Tool call 3: execute\_python
ok=True latency\_ms=33.5
Note: stdout/stderr text is not present in the stored trace JSON; only call status and latency are recorded.

Tool call 4: submit\_answer
ok=True latency\_ms=0.0
\end{tcolorbox}

\begin{tcolorbox}[traceanswerstyle, title={Final output}]
\{
  "satisfiable": false,
  "solution": null,
  "reasoning": "Langford sequences L(2,n) exist if and only if n ≡ 0 or 3 (mod 4). Since 6 ≡ 2 (mod 4), no valid Langford sequence of length 12 exists for n=6. This is a well-known result in combinatorial design theory, and exhaustive backtracking search confirms unsatisfiability."
\}
\end{tcolorbox}

\begin{tcolorbox}[traceverdictokstyle, title={Verification verdict}]
ground\_truth\_satisfiable=False
submitted\_satisfiable=False
correct=True
satisfiability\_correct=True
solution\_correct=True
validation\_details=Correct UNSAT
failure\_bucket=none
\end{tcolorbox}


\section{Problem Catalog}
\label{app:problems-cat}

This appendix catalogs the 39 problems in \textsc{MathConstraint}. For each problem we provide a definition, the canonical historical or modeling reference, and a prompt example as presented to evaluated models. For the random-graph and partial-clue problems the variable data block (\texttt{<edges>}, \texttt{<clues>}) is elided for space and is generated per instance from the seed.

\subsection{All-Interval Series}
\label{app:all_interval}
Find a permutation of \(\{0,\dots,n-1\}\) such that the absolute differences between consecutive elements also form a permutation of \(\{1,\dots,n-1\}\)~\cite{hoos1999sat}.

\begin{tcolorbox}[promptstyle, title={$n = 10$}]
Find an all-interval series of size 10. This is a permutation of 0 to 9 such that the absolute differences between consecutive elements are also a permutation of 1 to 9.

Return a list of 10 integers forming a valid permutation, or state "UNSATISFIABLE" if no solution exists.
\end{tcolorbox}

\subsection{Antimagic Square}
\label{app:antimagic_square}
Arrange the integers \(1,\dots,n^2\) in an \(n\times n\) grid so that the \(2n+2\) line sums (rows, columns, both main diagonals) form \(2n+2\) consecutive integers~\cite{lindon1962antimagic}.

\begin{tcolorbox}[promptstyle, title={$n = 5$}]
Construct an antimagic square of order 5: an n×n grid filled with the integers 1..25 (each used exactly once) such that the 12 line sums (n rows, n cols, both main diagonals) form 12 CONSECUTIVE integers (all distinct, with max - min = 11).

Return the grid + sums as a flat list of 37 integers: first 25 are the grid (row-major; cell (i,j) at index i*5+j), next 12 are the sums (rows 0..4, then cols 0..4, then main diagonal, then anti-diagonal). State "UNSATISFIABLE" if no such square exists.
\end{tcolorbox}

\subsection{Balanced Incomplete Block Design}
\label{app:bibd}
Construct a balanced incomplete block design \(\mathrm{BIBD}(v,k,\lambda)\): a \(v\times b\) binary incidence matrix where every block has \(k\) points and every pair of points appears in exactly \(\lambda\) common blocks~\cite{https://doi.org/10.1111/j.1469-1809.1936.tb02134.x}.

\begin{tcolorbox}[promptstyle, title={$v = 7,\ k = 3,\ \lambda = 1$ (Fano plane)}]
Construct a Balanced Incomplete Block Design BIBD(7, 3, 1): an v×b binary incidence matrix x (where v=7 points and b=7 blocks) such that:
  - Every row sums to r=3 (each point in r blocks)
  - Every column sums to k=3 (each block has k points)
  - For every pair of distinct rows $i_1, i_2$: the inner product $\sum_j x[i_1][j] * x[i_2][j] = \lambda=1$ (every pair of points appears in exactly lambda common blocks)
Existence requires Fisher's inequality (b >= v) and integrality of b, r.

Return the 7x7 matrix as a flat list of 49 integers in row-major order (cell (i,j) at index i*7+j), or state "UNSATISFIABLE".
\end{tcolorbox}

\subsection{Costas Array}
\label{app:costas_array}
Place \(n\) marks on an \(n\times n\) grid (one per row and one per column) so that all \(\binom{n}{2}\) displacement vectors between mark pairs are pairwise distinct~\cite{1457235}.

\begin{tcolorbox}[promptstyle, title={$n = 8$}]
Find a Costas array of order 8: place 8 marks on an 8x8 grid, one per row and one per column, such that all 8*(7)/2 displacement vectors (dr, dc) between pairs of marks (with dc > 0) are pairwise distinct.

Return x as a list of 8 integers in 0..7, where x[c] is the row of the mark in column c (permutation), or state "UNSATISFIABLE" if no such array exists.
\end{tcolorbox}

\subsection{de Bruijn Sequence}
\label{app:debruijn}
Construct a de~Bruijn sequence \(B(b,n)\): a cyclic sequence of length \(b^n\) over a \(b\)-letter alphabet in which every length-\(n\) subsequence appears exactly once~\cite{de1946combinatorial}.

\begin{tcolorbox}[promptstyle, title={$b = 2,\ n = 4$}]
Construct a De Bruijn sequence B(2, 4): a cyclic sequence of length $16 = 2^4$ over the alphabet {0, ..., 1} such that every n-length subsequence (read cyclically) appears EXACTLY ONCE. Existence is guaranteed (de Bruijn 1946) but finding one at scale is non-trivial.

Return the sequence as a flat list of 16 integers in {0, ..., 1}, or state "UNSATISFIABLE".
\end{tcolorbox}

\subsection{Golomb Ruler}
\label{app:golomb}
Find a Golomb ruler with \(n\) marks: a set of \(n\) integers (with the first at 0) such that all pairwise distances between marks are distinct~\cite{1454786}.

\begin{tcolorbox}[promptstyle, title={$n = 7$}]
Find a Golomb ruler with 7 marks. A Golomb ruler is a set of 7 integers (marks) such that all pairwise distances between marks are unique. The first mark should be at position 0.

Return a list of 7 integers in increasing order representing the mark positions, or state "UNSATISFIABLE" if no solution exists.
\end{tcolorbox}

\subsection{Graceful Graph Labeling}
\label{app:graceful_graph}
Find a graceful labeling of \(p\) disjoint \(K_k\) cliques chained by inter-clique edges: assign each vertex a unique label so that all edge labels (absolute label differences) are distinct~\cite{rosa1966certain}.

\begin{tcolorbox}[promptstyle, title={$k = 3,\ p = 4$}]
Find a graceful labeling for the graph $G_{3,4}$: 4 disjoint $K_3$ cliques (numbered 0 through 3), where each pair of consecutive cliques (g, g+1) is connected by 3 edges that link vertex i of clique g to vertex i of clique g+1 for each i in 0..2. The graph has 12 vertices and 21 edges in total.

A graceful labeling assigns each vertex a unique label from 0 to 21 such that all edge labels |label(u) - label(v)| are distinct.

Return a list of 12 integers giving the vertex labels, ordered by clique then vertex within clique (so the i-th block of 3 consecutive entries gives the labels for clique i), or state "UNSATISFIABLE" if no such labeling exists.
\end{tcolorbox}

\subsection{Graph $k$-Coloring}
\label{app:graph_k_coloring}
Decide whether a random graph admits a proper \(k\)-coloring; classical NP-complete decision~\cite{Karp1972}.

\begin{tcolorbox}[promptstyle, title={$n = 20,\ k = 3$}]
Decide whether the following random graph G with 20 vertices (numbered 0 to 19) admits a proper 3-coloring (each vertex assigned a color from {0, 1, ..., 2} such that no two adjacent vertices share a color). NP-complete decision.

Edges of G (<edge\_count> edges):
<edges>

Return the coloring as a flat list of 20 integers in {0, ..., 2}, or state "UNSATISFIABLE" if no proper 3-coloring exists.
\end{tcolorbox}

\subsection{Hadamard/Legendre Pair}
\label{app:hadamard}
Find a Legendre pair: two \(\pm 1\) sequences of odd length \(n\) whose cyclic cross-correlations all equal \(-2\), corresponding to Hadamard matrix construction~\cite{hadamard1893resolution}.

\begin{tcolorbox}[promptstyle, title={$n = 11$}]
Find two binary sequences x and y of length 11 (with n odd), each over the alphabet {-1, +1}, such that:
  (a) Sum of x equals 1, Sum of y equals 1
  (b) For every k in 1..5, the cyclic cross-correlation
            $\sum_{i=0}^{10} x[i] * x[(i+k) \bmod 11]  +  \sum_{i=0}^{10} y[i] * y[(i+k) \bmod 11]$   equals -2.

Return the concatenation x ++ y as a flat list of 22 integers (first 11 are x, next 11 are y), or state "UNSATISFIABLE".
\end{tcolorbox}

\subsection{Hamiltonian Cycle}
\label{app:hamilton_cycle}
Decide whether a random graph contains a Hamiltonian cycle visiting every vertex exactly once; classical NP-complete decision~\cite{Karp1972}.

\begin{tcolorbox}[promptstyle, title={$n = 15$}]
Decide whether the following random graph G with 15 vertices (numbered 0 to 14) contains a Hamiltonian cycle - a cyclic ordering visiting every vertex exactly once where every consecutive pair (and the pair (last, first)) is an edge of G. NP-complete.

Edges of G (<edge\_count> edges):
<edges>

Return the Hamiltonian cycle as a flat list of 15 distinct integers in [0, 15) giving the cycle order (with implicit wrap-around between last and first), starting at vertex 0. State "UNSATISFIABLE" if no Hamiltonian cycle exists.
\end{tcolorbox}

\subsection{Knight\'s Tour}
\label{app:knight_tour}
Find an open knight's tour on an \(n\times n\) board starting from cell 0~\cite{euler1766solution}.

\begin{tcolorbox}[promptstyle, title={$n = 6$}]
Find an open knight's tour on an 6x6 chessboard that starts at cell 0 (row 0, column 0). A knight's tour visits every cell exactly once via knight moves; cells are numbered in row-major order so cell k is at row k//6, column k

Return a list of 36 integers giving the cell visited at each step (a permutation of 0..35 starting with 0), or state "UNSATISFIABLE" if no such tour exists.
\end{tcolorbox}

\subsection{Langford Pairing}
\label{app:langford}
Construct a Langford pairing \(L(2,n)\): a sequence of length \(2n\) containing two copies of each integer in \(\{1,\dots,n\}\), with the two copies of value \(v\) exactly \(v+1\) positions apart~\cite{Langford_1958}.

\begin{tcolorbox}[promptstyle, title={$n = 8$}]
Construct a Langford sequence L(2,8): a sequence of length 16 containing exactly 2 copies of each integer from 1 to 8, such that the two occurrences of each value v are exactly v+1 positions apart (so copies of 1 are 1 position apart, copies of 2 are 2 positions apart, etc.). Example: L(2,3) has solution [2, 3, 1, 2, 1, 3].

Return seq as a list of 16 integers in 1..8, or state "UNSATISFIABLE" if no such sequence exists.
\end{tcolorbox}

\subsection{Latin Square Completion}
\label{app:latin_square_completion}
Complete a partial Latin square: fill the empty cells of an \(n\times n\) grid so each value in \(\{0,\dots,n-1\}\) appears exactly once per row and per column~\cite{colbourn1984complexity}.

\begin{tcolorbox}[promptstyle, title={$n = 7$, density 40\%}]
Complete the partial Latin square of order 7. The grid is 7x7; some cells are pre-filled with values from {0, ..., 6}, others are empty (denoted -1). Fill every empty cell so that each of the 7 values appears exactly once in every row and exactly once in every column.

The pre-filled clues (flat row-major; cell (i,j) at index i*7+j; -1 = empty):
<clues>

Return the completed grid as a flat list of 49 integers in row-major order (cell (i,j) at index i*7+j), or state "UNSATISFIABLE".
\end{tcolorbox}

\subsection{Low Autocorrelation Binary Sequence}
\label{app:low_autocorrelation}
Find a \(\pm 1\) binary sequence of length \(n\) whose squared aperiodic autocorrelations sum below a threshold~\cite{bernasconi1987low}.

\begin{tcolorbox}[promptstyle, title={$n = 12$}]
Find a binary sequence of length 12 over the alphabet {-1, +1} with low aperiodic autocorrelation: specifically, the sum over k=1..11 of $C_k^2$, where $C_k = \sum_{i=0}^{n-k-1} seq[i]*seq[i+k]$, must be at most 14.

Return seq as a list of 12 integers, each -1 or +1, or state "UNSATISFIABLE" if no such sequence exists.
\end{tcolorbox}

\subsection{Magic Sequence}
\label{app:magic_sequence}
Find a sequence \(x[0],\dots,x[n-1]\) such that each \(x[i]\) equals the count of \(i\)'s appearance in the sequence~\cite{van1989constraint}.

\begin{tcolorbox}[promptstyle, title={$n = 8$}]
Find a magic sequence of length 8. A magic sequence is a sequence x[0], x[1], ..., x[7] where each x[i] equals the count of how many times i appears in the sequence.

Return a list of 8 integers, or state "UNSATISFIABLE" if no solution exists.
\end{tcolorbox}

\subsection{Magic Square}
\label{app:magic_square}
Arrange \(1,\dots,n^2\) in an \(n\times n\) grid so every row, column, and main diagonal sums to the magic constant~\cite{andrews1917magic}.

\begin{tcolorbox}[promptstyle, title={$n = 4$}]
Construct an n x n magic square of order 4: arrange the integers 1..16 so each value appears exactly once and the sum of every row, every column, and both main diagonals equals 34.

Pre-filled clues (flat row-major; cell (i,j) at index i*4+j; 0 = empty):
[0, 0, 0, 0, 0, 0, 0, 0, 0, 0, 0, 0, 0, 0, 0, 0]

Return the magic square as a flat list of 16 integers in row-major order (cell (i,j) at index i*4+j), or state "UNSATISFIABLE".
\end{tcolorbox}

\subsection{Maximum Clique}
\label{app:max_clique}
Decide whether a random graph contains a clique of size at least \(k\); NP-hard~\cite{Karp1972}.

\begin{tcolorbox}[promptstyle, title={$n = 30,\ k = 6$}]
Decide whether the following random graph G with 30 vertices (numbered 0 to 29) contains a clique of size at least 6: a subset of 6 vertices, all pairwise adjacent. NP-hard.

Edges of G (<edge\_count> edges):
<edges>

Return the clique as a flat 0/1 vector of length 30 (entry i is 1 iff vertex i is in the clique). The number of 1s must be at least 6. State "UNSATISFIABLE" if no such clique exists.
\end{tcolorbox}

\subsection{Maximum Independent Set}
\label{app:max_independent_set}
Decide whether a random graph contains an independent set of size at least \(k\); NP-hard~\cite{Karp1972}.

\begin{tcolorbox}[promptstyle, title={$n = 25,\ k = 8$}]
Decide whether the following random graph G with 25 vertices (numbered 0 to 24) has an independent set of size at least 8 (a subset of vertices, no two of which are adjacent). NP-hard.

Edges of G (<edge\_count> edges):
<edges>

Return an independent set as a flat 0/1 vector of length 25 (entry i is 1 iff vertex i is in the set). The number of 1s must be at least 8. State "UNSATISFIABLE" if no such set exists.
\end{tcolorbox}

\subsection{Non-Transitive Dice}
\label{app:non_transitive_dice}
Design \(n\) dice with \(m\) faces such that die \(i\) beats die \((i{+}1) \bmod n\) in head-to-head face comparison, forming a cyclic ``rock--paper--scissors'' relation (Efron's dice)~\cite{Gardner:1970:MGl}.

\begin{tcolorbox}[promptstyle, title={$n = 3,\ m = 6$}]
Design 3 cyclically intransitive dice, each with 6 faces showing values in {0, ..., 11}. The dice must satisfy: die i 'beats' die (i+1) mod 3 in head-to-head face comparison - i.e., for every i in [0, 3): \#\{(a,b) : face(i,a) > face((i+1) mod 3,b)\} > 18, out of 36 possible (a,b) face pairs. Famous problem (Efron's dice).

Return a flat list of 18 integers: entry i*6+j is the j-th face value (0-indexed) of die i (faces sorted ascending). State "UNSATISFIABLE" if no such dice exist.
\end{tcolorbox}

\subsection{Number Partitioning}
\label{app:number_partitioning}
Partition \(\{1,\dots,n\}\) into \(k\) disjoint subsets of equal sum~\cite{Karp1972}.

\begin{tcolorbox}[promptstyle, title={$n = 12,\ k = 3$}]
Partition the integers {1, 2, ..., 12} into 3 disjoint subsets, each summing to the same value 26. (The total sum is 12*(12+1)/2 = 78, so the target is 78/3 = 26.) NP-hard partition problem.

Return a flat list of 12 integers in {0, ..., 2} where the i-th entry (0-indexed) gives the subset assignment of integer i+1. State "UNSATISFIABLE" if no such partition exists.
\end{tcolorbox}

\subsection{Orthogonal Latin Squares}
\label{app:ortholatin}
Construct a pair of orthogonal Latin squares of order \(n\) (a Greco-Latin square): impossible for \(n \in \{2, 6\}\), satisfiable for all other \(n \geq 3\)~\cite{euler1782recherches}. Note we give the model a small hint about existence on this instance.

\begin{tcolorbox}[promptstyle, title={$n = 5$}]
Construct a pair of orthogonal Latin squares of order 5 (also called a Greco-Latin square): two n×n grids X and Y, each filled with values from {0, 1, ..., 4}, such that:
  - X is a Latin square (every row and every column is a permutation of 0..4)
  - Y is a Latin square
  - X and Y are orthogonal: the 25 pairs (X[i][j], Y[i][j]) for i,j in [0, 5) are all distinct
Famously, no such pair exists for n=2 or n=6 (Tarry's theorem); they exist for every other n >= 3.

Return the concatenation X.flatten() ++ Y.flatten() as a flat list of 50 integers (first 25 are X row-major, next 25 are Y row-major), or state "UNSATISFIABLE".
\end{tcolorbox}

\subsection{Chromatic Girth Graph}
\label{app:pysms_chromatic_girth}
Generate a graph on \(v\) vertices with bounded chromatic number, lower-bounded girth, and a minimum edge count~\cite{erdos1959graph}.

\begin{tcolorbox}[promptstyle, title={\(v=14\), \(\chi \le 3\), girth \(\ge 7\), edges \(\ge 18\)}]
Generate a graph with 14 vertices that is colorable with at most 3 colors, has girth (shortest cycle) at least 7, and has at least 18 edges.

Return the graph as a list of edges (u, v) with 0 <= u < v < 14, or state "UNSATISFIABLE" if no graph exists.
\end{tcolorbox}

\subsection{Clique Coloring Graph}
\label{app:pysms_clique_coloring}
Generate a graph with bounded clique number, bounded chromatic number, and a minimum-degree constraint --- the Mycielski-style separation problem~\cite{mycielski1955coloriage}.

\begin{tcolorbox}[promptstyle, title={\(v=15\), \(\omega \le 4\), \(\chi \le 4\), \(\delta \ge 5\)}]
Generate a graph with 15 vertices where the maximum clique size is at most 4, the chromatic number is at most 4, and every vertex has degree at least 5.

Return the graph as a list of edges (u, v) with 0 <= u < v < 15, or state "UNSATISFIABLE" if no graph exists.
\end{tcolorbox}

\subsection{Combined Graph Constraints}
\label{app:pysms_combined_graph}
A compound graph-generation problem stacking up to ten simultaneous structural constraints (degrees, edges, clique number, independence number, chromatic number, connectivity, girth, $K_k$-freeness, etc.); always UNSAT in our calibration~\cite{kirchweger2021sat}.

\begin{tcolorbox}[promptstyle, title={mixed constraints, \(v = 12\)}]
Generate a graph with 12 vertices that has the following constraints:
  - minimum degree at least 2
  - maximum degree at most 4
  - minimum number of edges: 14
  - maximum number of edges: 18
  - maximum clique size at most 3
  - maximum independent set size at most 4
  - chromatic number at most 3
  - vertex-connectivity at least 2
  - girth at least 4
  - free of $K_4$ subgraphs

Return the graph as a list of edges (u, v) with 0 <= u < v < 12, or state "UNSATISFIABLE" if no graph exists.
\end{tcolorbox}

\subsection{Disjoint Clique Graph}
\label{app:pysms_contains_cliques}
Generate a graph that is exactly the disjoint union of \(c\) cliques of size \(s\), with no other edges --- SAT iff \(c\cdot s = v\)~\cite{kirchweger2021sat}.

\begin{tcolorbox}[promptstyle, title={\(v=10\), 2 cliques of size 5}]
Generate a graph with 10 vertices that consists of exactly 2 vertex-disjoint clique(s), each of size 5. Every vertex must belong to one of these cliques, and the only edges in the graph are those within these cliques.

Return the graph as a list of edges (u, v) with 0 <= u < v < 10, or state "UNSATISFIABLE" if no graph exists.
\end{tcolorbox}

\subsection{Degree-Bounded Graph}
\label{app:pysms_degree_bounds}
Generate a graph in which every vertex has degree within prescribed bounds~\cite{turan1941external,mantel1907problem}.

\begin{tcolorbox}[promptstyle, title={\(v=10\), degrees in \([2,4]\)}]
Generate a graph with 10 vertices where every vertex has degree between 2 and 4.

Return the graph as a list of edges (u, v) with 0 <= u < v < 10, or state "UNSATISFIABLE" if no graph exists.
\end{tcolorbox}

\subsection{Girth-and-Degree Graph}
\label{app:pysms_girth_degree}
Generate a graph (a near-cage) with both girth and degree constraints~\cite{erdos1963regulare}.

\begin{tcolorbox}[promptstyle, title={\(v=10\), girth \(\ge 5\), degrees in \([2,4]\)}]
Generate a graph with 10 vertices where the girth (shortest cycle) is at least 5 and every vertex has degree between 2 and 4.

Return the graph as a list of edges (u, v) with 0 <= u < v < 10, or state "UNSATISFIABLE" if no graph exists.
\end{tcolorbox}

\subsection{Graph Builder}
\label{app:pysms_graph_builder}
A multi-constraint graph generation problem that mixes degree bounds, edge counts, and chromatic constraints into a single specification~\cite{kirchweger2021sat}.

\begin{tcolorbox}[promptstyle, title={\(v=10\), mixed constraints}]
Generate a graph with 10 vertices satisfying:
  - minimum degree at least 2
  - maximum degree at most 4
  - number of edges between 10 and 14
  - chromatic number between 2 and 3

Return the graph as a list of edges (u, v) with 0 <= u < v < 10, or state "UNSATISFIABLE" if no graph exists.
\end{tcolorbox}

\subsection{Independence and Connectivity Graph}
\label{app:pysms_independent_connectivity}
Generate a graph with both bounded independence number and lower-bounded vertex connectivity --- the Chv\'atal--Erd\H{o}s setting~\cite{chvatal1972note}.

\begin{tcolorbox}[promptstyle, title={\(v=10\), \(\alpha \le 4\), \(\kappa \ge 2\)}]
Generate a graph with 10 vertices where the maximum independent set size is at most 4 and the vertex-connectivity is at least 2.

Return the graph as a list of edges (u, v) with 0 <= u < v < 10, or state "UNSATISFIABLE" if no graph exists.
\end{tcolorbox}

\subsection{Minimum-Girth Graph}
\label{app:pysms_min_girth}
Generate a graph on \(v\) vertices whose girth is at least \(g\)~\cite{erdos1959graph}.

\begin{tcolorbox}[promptstyle, title={\(v=10\), girth \(\ge 5\)}]
Generate a graph with 10 vertices where the girth (shortest cycle length) is at least 5.

Return the graph as a list of edges (u, v) with 0 <= u < v < 10, or state "UNSATISFIABLE" if no graph exists.
\end{tcolorbox}

\subsection{Maximal Triangle-Free Graph}
\label{app:pysms_mtf}
Generate a maximal triangle-free graph on \(v\) vertices: triangle-free, but adding any non-edge would create a triangle~\cite{kirchweger2021sat}.

\begin{tcolorbox}[promptstyle, title={\(v=10\)}]
Generate a maximal triangle-free graph with 10 vertices.

Return the graph as a list of edges (u, v) with 0 <= u < v < 10, or state "UNSATISFIABLE" if no graph exists.
\end{tcolorbox}

\subsection{Ramsey Graph}
\label{app:pysms_ramsey}
Generate a Ramsey-witness graph on \(v\) vertices that contains no \(r\)-clique and no \(s\)-independent-set; SAT iff \(v < R(r,s)\)~\cite{https://doi.org/10.1112/plms/s2-30.1.264,kirchweger2021sat}.

\begin{tcolorbox}[promptstyle, title={\(v=8\), avoid \(K_3\) and \(\bar{K}_3\)}]
Generate a graph with 8 vertices that contains no clique of size 3 and no independent set of size 3. (A Ramsey-witness graph: such graphs exist iff vertices < R(clique\_avoid, indset\_avoid).)

Return the graph as a list of edges (u, v) with 0 <= u < v < 8, or state "UNSATISFIABLE" if no graph exists.
\end{tcolorbox}

\subsection{Idempotent Quasigroup}
\label{app:quasigroup_idempotent}
Construct a QG3 quasigroup of order \(n\): a Latin square with idempotent diagonal satisfying \(x[x[i][j]][x[j][i]] = i\)~\cite{slaney1995automated}.

\begin{tcolorbox}[promptstyle, title={$n = 5$}]
Construct a quasigroup of order 5 satisfying QG3 idempotence and the involution property. Specifically: an n×n grid x with values in {0, ..., 4} such that:
  - Every row and every column is a permutation of 0..4 (Latin square)
  - x[i][i] = i for all i in [0, 5) (idempotent diagonal)
  - x[ x[i][j] ][ x[j][i] ] = i for all i, j in [0, 5) (QG3)
The third constraint involves nested indexing (the indices themselves are values from x). For n in [5, 12], few solutions exist and search is non-trivial.

Return the grid as a flat list of 25 integers in row-major order (cell (i,j) at index i*5+j), or state "UNSATISFIABLE".
\end{tcolorbox}

\subsection{$n$-Queens}
\label{app:queens}
Place \(n\) non-attacking queens on an \(n\times n\) chessboard~\cite{bezzel1848proposal}.

\begin{tcolorbox}[promptstyle, title={$n = 8$}]
Place 8 queens on a 8x8 chessboard such that no two queens attack each other. Queens attack along rows, columns, and diagonals.

Return a list of 8 integers where the i-th integer is the column position (0 to 7) of the queen in row i, or state "UNSATISFIABLE" if no solution exists.
\end{tcolorbox}

\subsection{Ramsey Edge Coloring}
\label{app:ramsey}
2-color the edges of \(K_n\) avoiding both a monochromatic red \(K_r\) and a monochromatic blue \(K_s\); SAT iff \(n < R(r,s)\)~\cite{https://doi.org/10.1112/plms/s2-30.1.264}.

\begin{tcolorbox}[promptstyle, title={$n=6,\ r=s=3$ (just below $R(3,3)=6$)}]
Find a 2-coloring of the edges of the complete graph $K_6$ such that there is no monochromatic red clique of size 3 and no monochromatic blue clique of size 3. Each edge is colored either 0 (red) or 1 (blue).

Return a list of 15 integers (0 or 1) representing the colors of edges listed in lexicographic order of (i,j) for i<j, or state "UNSATISFIABLE" if no such coloring exists.
\end{tcolorbox}

\subsection{Social Golfers}
\label{app:social_golfers}
Schedule \(g\cdot s\) golfers in \(g\) groups of \(s\) over \(w\) weeks so that no pair plays together more than once~\cite{csplib:prob010}.

\begin{tcolorbox}[promptstyle, title={4 groups of 3, 4 weeks}]
Social Golfers (CSPLib problem 10): schedule 12 = 4*3 golfers over 4 weeks. Each week, the 12 golfers are partitioned into 4 groups of 3 players each. Constraint: NO PAIR of golfers may be in the same group on more than one week. SAT iff a valid schedule exists; classic NP-hard scheduling problem.

Return a flat list of 48 integers: entry w*12 + p gives the group (in 0..3) of golfer p in week w. State "UNSATISFIABLE" if no schedule exists.
\end{tcolorbox}

\subsection{Sudoku}
\label{app:sudoku}
Fill an \(n^2\times n^2\) Sudoku grid (with \(n\times n\) blocks) so each row, column, and block contains every value in \(\{1,\dots,n^2\}\) exactly once~\cite{e86-a_5_1052}.

\begin{tcolorbox}[promptstyle, title={$n = 3$ (standard $9\times 9$)}]
Fill a Sudoku grid with block size 3 (so the full grid is 9x9, containing 9 rows, 9 columns, and 9 non-overlapping 3x3 blocks). Each row, column, and block must contain every integer from 1 to 9 exactly once.

Return a list of 81 integers (the grid in row-major order: cell at row i, column j is at index i*9+j), or state "UNSATISFIABLE" if no solution exists.
\end{tcolorbox}

\subsection{Van der Waerden Coloring}
\label{app:van_der_waerden}
Color \(\{1,\dots,n\}\) with \(k\) colors avoiding any monochromatic arithmetic progression of length \(L\); SAT iff \(n < W(k,L)\)~\cite{van1927beweis}.

\begin{tcolorbox}[promptstyle, title={$n=20,\ k=2,\ L=3$ (below $W(2,3)=9$? No: SAT past $W$ is impossible)}]
Color the integers 1, 2, ..., 20 with 2 colors (numbered 0 to 1) such that NO monochromatic arithmetic progression of length 3 exists: there is no a >= 1, d >= 1 with a + (2)*d <= 20 where the 3 positions a, a+d, ..., a+(2)*d all share the same color. Such a coloring exists iff 20 < W(2, 3). Known values: W(2,3)=9, W(2,4)=35, W(3,3)=27, W(2,5)=178.

Return a list of 20 integers, each 0 or 1, representing the color of integer i+1 (0-indexed), or state "UNSATISFIABLE" if no such coloring exists.
\end{tcolorbox}

\subsection{Vertex Cover}
\label{app:vertex_cover}
Decide whether a random graph admits a vertex cover of size at most \(k\); classical NP-complete decision~\cite{Karp1972}.

\begin{tcolorbox}[promptstyle, title={$n = 30,\ k = 12$}]
Decide whether the following random graph G with 30 vertices (numbered 0 to 29) has a vertex cover of size at most 12: a subset S of vertices such that every edge of G has at least one endpoint in S, and |S| <= 12. NP-complete.

Edges of G (<edge\_count> edges):
<edges>

Return the cover as a flat 0/1 vector of length 30 (entry i is 1 iff vertex i is in S). The number of 1s must be at most 12. State "UNSATISFIABLE" if no such cover exists.
\end{tcolorbox}



\newpage

\end{document}